\documentclass[10pt,twocolumn,letterpaper]{article}

\usepackage[final]{iccv}      

\usepackage[dvipsnames]{xcolor}
\definecolor{citecolor}{HTML}{0071bc}
\definecolor{color_ao}{gray}{0.5}
\definecolor{color_our}{HTML}{e6f2c2}
\definecolor{color_pre}{rgb}{0.52,0.59,0.69}
\definecolor{Gray}{gray}{0.9}
\definecolor{LighterGray}{gray}{0.93}
\definecolor{LightGrayForTableRule}{gray}{0.92}
\definecolor{DarkGray}{gray}{0.5}
\definecolor{Black}{rgb}{0.0, 0.0, 0.0}
\definecolor{NiceBlue}{rgb}{0.11764705882352941, 0.5647058823529412, 1.0}
\definecolor{NiceGreen}{HTML}{b6c783}
\usepackage{diagbox}
\definecolor{gray}{HTML}{b0aeae}
\definecolor{diffgray}{HTML}{878787}
\definecolor{NiceGray}{HTML}{696969}
\definecolor{applegreen}{rgb}{0.55, 0.71, 0.0}
\definecolor{ourbrown}{HTML}{5f0505}

\definecolor{demphcolor}{RGB}{144,144,144}
\newcommand{\demph}[1]{\textcolor{demphcolor}{#1}}

\newcount\Comments 
\usepackage{array,multirow,graphicx}
\usepackage{color}
\usepackage{wrapfig}
\definecolor{purple}{rgb}{1,0,1}
\newcolumntype{a}{>{\columncolor{lightblue}}c}
\newcommand{\kibitz}[2]{\ifnum\Comments=1\textcolor{#1}{#2}\fi}

\usepackage{tkz-kiviat}
\usepackage{xfp}
\usetikzlibrary{arrows}

\usepackage{epsfig}
\usepackage{graphicx}
\usepackage{tikz}
\usepackage{pgfplots}
    \pgfplotsset{compat=1.18}
\usepackage{caption}
\usepgfplotslibrary{groupplots}
\usetikzlibrary{patterns}

\usepackage{lipsum}

\definecolor{cvprblue}{rgb}{0.21,0.49,0.74}
\usepackage[pagebackref,breaklinks,colorlinks,citecolor=cvprblue]{hyperref}
\usepackage{multirow}
\usepackage{xcolor,colortbl}
\hypersetup{
    colorlinks,
    linkcolor={red},
    citecolor={green}
}
\usepackage{amssymb}
\usepackage{pifont}
\usepackage{marvosym}
\usepackage{fontawesome}

\definecolor{Light}{HTML}{effdfe}

\newcommand\blfootnote[1]{%
  \begingroup
  \renewcommand\thefootnote{}\footnote{#1}%
  \addtocounter{footnote}{-1}%
  \endgroup
}

\usepackage{listings}
\definecolor{lightblue}{HTML}{E9EEFF}
\definecolor{lightred}{HTML}{FFEFEF}
\definecolor{platinum}{HTML}{EEEEEE}

\definecolor{lightblue}{HTML}{E9EEFF}

\newcommand{\xpar}[1]{\noindent \textbf{#1}.}

\def\eg{\emph{e.g.}} 
\def\ie{\emph{i.e.}}

\newcommand{\x}{\mathbf{x}}
\newcommand{\y}{\mathbf{y}}

\newcommand{\VTG}{STG}

\definecolor{lavenderpurple}{rgb}{0.59, 0.48, 0.71}

\definecolor{lavenderpurple}{rgb}{0.59, 0.48, 0.71}

\newcommand{\SPcomment}[1]{}

\newcommand{\TAcomment}[1]{}

\newcommand{\EMNote}[1]{}

\newcommand{\EMcomment}[1]{}
\newcommand{\EM}[1]{{#1}}
\newcommand{\LTcomment}[1]{}

\newcommand{\ys}[1]{}

\newcommand{\todo}[1]{}
\newcommand{\TODO}[1]{}

\newcommand{\CC}[1]{\cellcolor{Light}}

\newcommand{\model}{\text{ED-VTG}}

\definecolor{customgray1}{gray}{0.5} 
\renewcommand{\demph}[1]{\textcolor{customgray1}{#1}}

\newcommand{\graycmidrule}[1]{%
  \arrayrulecolor{customgray1}%
  \cmidrule(lr){#1}%
  \arrayrulecolor{black}%
}

\def\paperID{9239} 
\def\confName{ICCV}
\def\confYear{2025}

\title{
Enrich and Detect: Video Temporal Grounding \\ with Multimodal LLMs \vspace{-7mm}
}

\author{Shraman Pramanick$^{\diamond1,2}$\textsuperscript{\Letter} \ \ \ \  Effrosyni Mavroudi$^{1}$  \ \ \ \ Yale Song$^{1}$ \ \ \ \  Rama Chellappa$^{2}$ \\  Lorenzo Torresani$^{3}$ \ \ \ \ Triantafyllos Afouras$^{1}$\textsuperscript{\Letter} \vspace{2mm}\\
$^{1}$FAIR, Meta, \ \ $^{2}$Johns Hopkins University \ \ $^{3}$Northeastern University \vspace{2mm}
\\
{\url{https://shramanpramanick.github.io/ED-VTG/} \vspace{-2mm}}
}

\begin{document}
\maketitle

\begin{abstract}
\vspace{-3mm}

We introduce \model, a method for fine-grained video temporal grounding utilizing multi-modal large language models. 
Our approach harnesses the capabilities of multimodal LLMs
to jointly process text and video, 
in order to effectively localize natural language queries in videos through a two-stage
process.
Rather than being directly grounded, language queries are initially transformed into enriched sentences that incorporate missing details and cues to aid in grounding. In the second stage, these enriched queries are grounded, using a lightweight decoder, which specializes at predicting accurate boundaries conditioned on contextualized representations of the enriched queries.
To mitigate noise and reduce the impact of hallucinations, our model is trained with a multiple-instance-learning objective that dynamically selects the optimal version of the query for each training sample.
We demonstrate state-of-the-art results across various benchmarks in temporal video grounding and paragraph grounding settings. Experiments reveal that our method significantly outperforms all previously proposed LLM-based temporal grounding approaches and is either superior or comparable to specialized models, while maintaining a clear advantage against them in zero-shot evaluation scenarios.

\vspace{-4mm}

\end{abstract}
\blfootnote{$^\star$Work done during an internship at FAIR.}
\blfootnote{$\textsuperscript{\Letter}$\url{spraman3@jhu.edu}, \url{afourast@meta.com}}
\begin{figure}[t]
\centering
\includegraphics[width=\columnwidth]{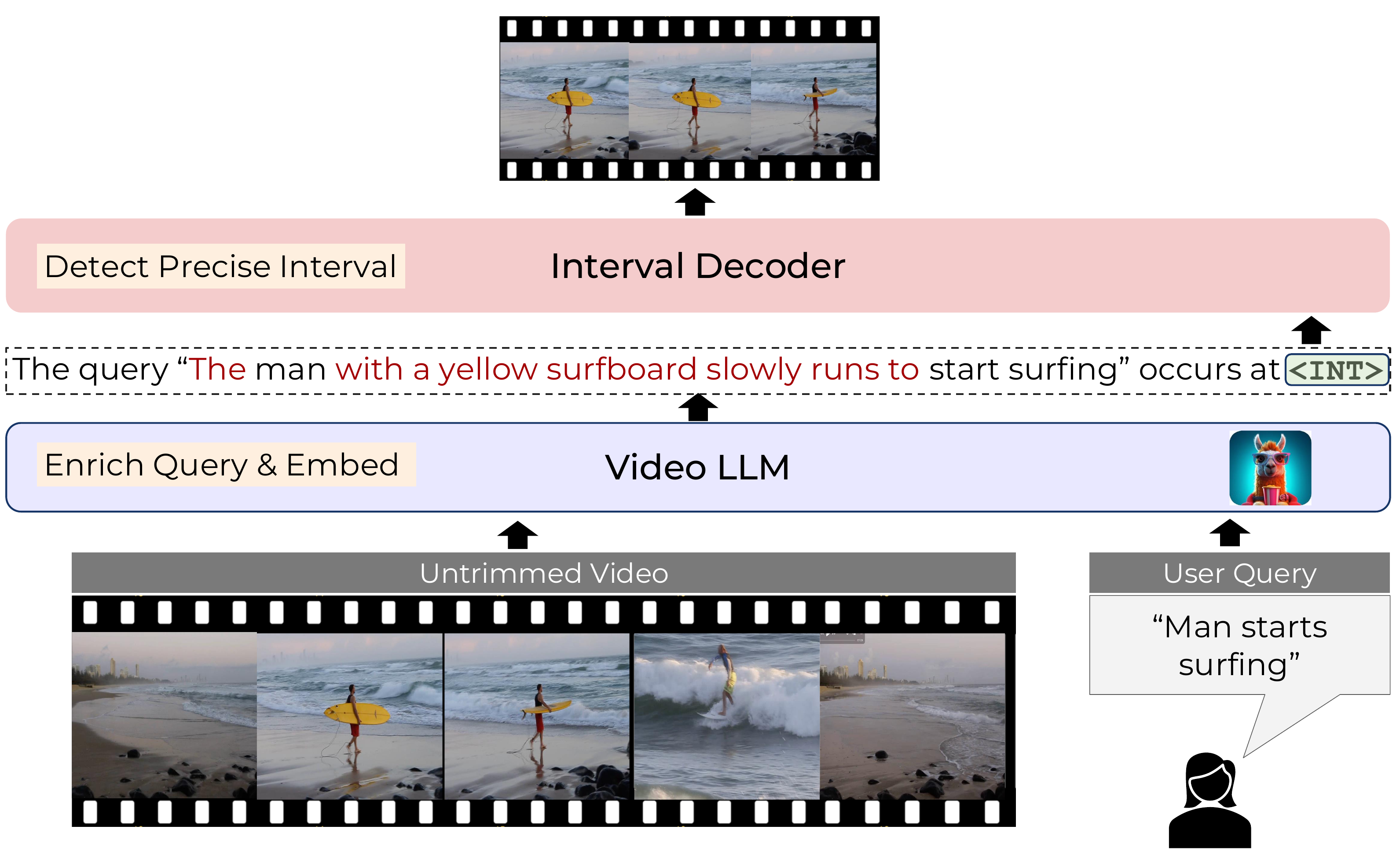}
\vspace{-6mm}
\caption{\textbf{Our proposed system.}
\model~performs video temporal grounding as a two-stage process: since user queries are often incomplete or coarse, the first stage involves producing an enriched query that adds additional details to the original, making it easier to ground. Meanwhile, a contextualized embedding is generated, containing all the information about the interval to be predicted. In the second stage, an interval decoder translates these embeddings into precise temporal boundaries.
}
\vspace{-5mm}
\label{fig:teaser}
\end{figure}

\vspace{-6mm}
\section{Introduction}
\vspace{-2mm}

Video temporal grounding \cite{zhang2020vslnet, lei2021momentdetr, moon2023qddetr, lin2023univtg, sun2024trdetr}
aims to identify temporal intervals in a video that correspond to a set of provided language queries.
The task is essential for applications such as video editing and content retrieval. Conversely, video captioning \cite{zhang2019object,pan2020spatio,sharma2020image,stefanini2022show,abdar2023review} entails generating a natural language description for a given video segment, effectively translating visual content into text.
The two tasks are in fact dual, as the outputs of one task are the inputs to the other, and vice versa.
Intuitively, there is significant potential in exploiting this synergy,
however it has largely remained unexplored, with previous works typically specializing in one of the two tasks \cite{chen2021end, zhang2019object, lei2021momentdetr, moon2023qddetr} or solving them in multi-task setting \cite{ren2024timechat, wang2024groundedvideollm} without investigating how each task can benefit the other.

In this work, we exploit this duality by leveraging captioning to enhance grounding.
Our key observation is that natural language queries often lack the completeness or detail necessary for effective temporal localization; indeed, existing grounding datasets frequently contain poorly worded, coarse, and potentially incomplete queries.
The quality and completeness of these queries however is crucial for the precision of the temporal grounding.
A natural hypothesis, therefore, is that more detailed queries, which could be obtained via conditional captioning,
can significantly enhance grounding accuracy.
For example, as illustrated in Figure~\ref{fig:teaser},
a vague query like \textit{`Man starts surfing'} can be refined into a more detailed description such as \textit{`The man with a yellow surfboard slowly runs to start surfing,'} resulting in grounding with more accurate temporal boundaries.
In other cases, refining an abstract concept may involve breaking down a complex query into simpler components that are more directly groundable. \ys{Add a short example?}

This key idea of query enrichment forms the basis of our proposed approach.
Concretely, we transform grounding into a two-stage reasoning process using a multi-modal LLM: first the model enriches the input query into a more detailed description by adding missing details based on the video content, and then temporally localizes the resulting enriched query in the video.

To effectively perform the temporal localization, we introduce a lightweight perception decoder that specializes in generating precise temporal boundaries, conditioned on a contextual representation, allowing the LLM to focus on language outputs, where it excels. The perception decoder allows us to leverage purposefully crafted training objectives for temporal grounding, building on prior knowledge and task-specific characteristics developed from years of research in object detection \cite{girshick2015rcnn, ren2016fasterrcnn, he2017maskrcnn, ross2017focal} and temporal localization \cite{zhang20202dtan, zhang2020vslnet, lei2021momentdetr, moon2023qddetr, lin2023univtg}.

Learning to jointly \textit{enrich and detect} requires high-quality enriched query labels.
We obtain those by using a strong external captioning model which we condition on the original queries and the video content of the target temporal boundaries.
However powerful, these models are prone to hallucinations and there is no guarantee that the enriched queries will always be easier to ground than the original ones. At the same time, annotating the ground truth to determine which query -- original or enriched -- is a better candidate, is extremely expensive.
To address this issue, we propose training in a multiple-instance learning (MIL) fashion, that enables the model to autonomously determine which query is better suited for the task during training.

Finally, we note that our proposed method is not equivalent to a data augmentation approach which simply pre-proecsses the training set to generate enhanced queries that are directly used for training. While this simpler alternative offers some of the same benefits, it suffers from the limitation that extracting enriched queries during training requires knowledge of the ground-truth temporal segments, \ie, the grounding targets. Since during inference these segments are unknown, the original queries must be used as input, which, as we will show experimentally, is suboptimal. Our method overcomes this limitation by learning to jointly enrich and detect, demonstrating superior performance.

To summarize, our contributions are as follows:
(i) we introduce a cascaded
approach to temporal grounding, where the model first enriches the provided language query based on the video context and then proceeds to localize it;
(ii) we enable multi-modal LLMs to accurately localize text queries using a lightweight decoder which allows for training with detection objectives tailored to the task;
(iii) we propose a multiple-instance learning paradigm that enables the model to dynamically select the query that leads to better temporal localization;
(iv) we achieve state-of-the-art results on several temporal grounding benchmarks, for both single query grounding and paragraph grounding, demonstrating, for the first time, an LLM-based model that surpasses or performs comparably to specialist models.


\section{Related Works}

\xpar{LLM-based temporal grounding}
Prior works have explored using LLMs for grounding natural language sentences in videos,
either using raw text tokens to represent timestamps \cite{huang2024vtimellm, ren2024timechat, li2024groundinggpt, ma2023llavilo, zeng2024timesuite, li2024videochat2} or by adding hundreds of special tokens to the LLM's vocabulary to represent video frames \cite{huang2025lita, qianmomentor, wang2024groundedvideollm}.
Our approach differs from these methods in that by utilizing a lightweight interval decoder we can apply detection losses such as L1 and gIoU with minimal added complexity; we additionally exploit the LLM's potential to describe video content in detail.

\xpar{Specialist models}
There is a rich variety of specialist models in the literature that are tailored to specific variants of temporal grounding, \eg~single query temporal grounding~\cite{zhang2020vslnet} and video paragraph grounding~\cite{tan2024siamgtr}, and as such, achieve strong performance~\cite{bao2021depnet, zeng2020drn, luo2022stlg, jiang2022svptr, tan2024siamgtr, liu2021cbln, zhang20202dtan, zhang20203dtpn, rodriguez2021dori}. Modern methods typically employ a multi-modal transformer~\cite{zhang2020vslnet} that fuses dense video features with text embeddings of the language query, followed by a specialized detector head for performing detections~\cite{rodriguez2021dori, rodriguez2023locformer, rodriguez2020tmlga, ghosh2019excl,lei2021momentdetr, moon2023qddetr}. However, because these models are often trained on limited datasets for such a narrowly defined task, they struggle with generalization. Indeed, the zero-shot performance of these methods is limited~\cite{xiao2022vgt, fu2023violetv2, xiao2024nextgqa, yang2022frozenbilm, yu2024sevila}.
In this work, we aim to address the shortcomings of previous methods by fully combining the generalization abilities of multi-modal LLMs with the advantages of specialist models.

\xpar{Dense captioning}
Our method is related to dense video captioning~\cite{krishna2017dense}, where the objective is to segment a given video into multiple parts and simultaneously provide descriptive captions for each segment.
Traditional approaches tackle this task either by first determining the segments and then providing descriptions~\cite{huang2024vtimellm,bao2021depnet,MDVC_Iashin_2020,ZhXuCoCVPR18,Li2018JointlyLA,zhou2018end},
or jointly learning both tasks~\cite{Deng_2021_CVPR,Li2018JointlyLA,yang2023vid2seq,Zhou2018EndtoEndDV,Zhu2022EndtoendDV}.
Recent advances have demonstrated video-conditioned LLMs to excel in this task \cite{zhu-etal-2022-end,HaoDenseCaptioning2024,zhou2024streaming}.
While there are similarities between our enrich-and-detect paradigm and dense captioning,
we solve a different task, namely video temporal grounding, where the input query is given and constrains the problem.

\begin{figure*}[!t]
\centering
\includegraphics[width=\textwidth]{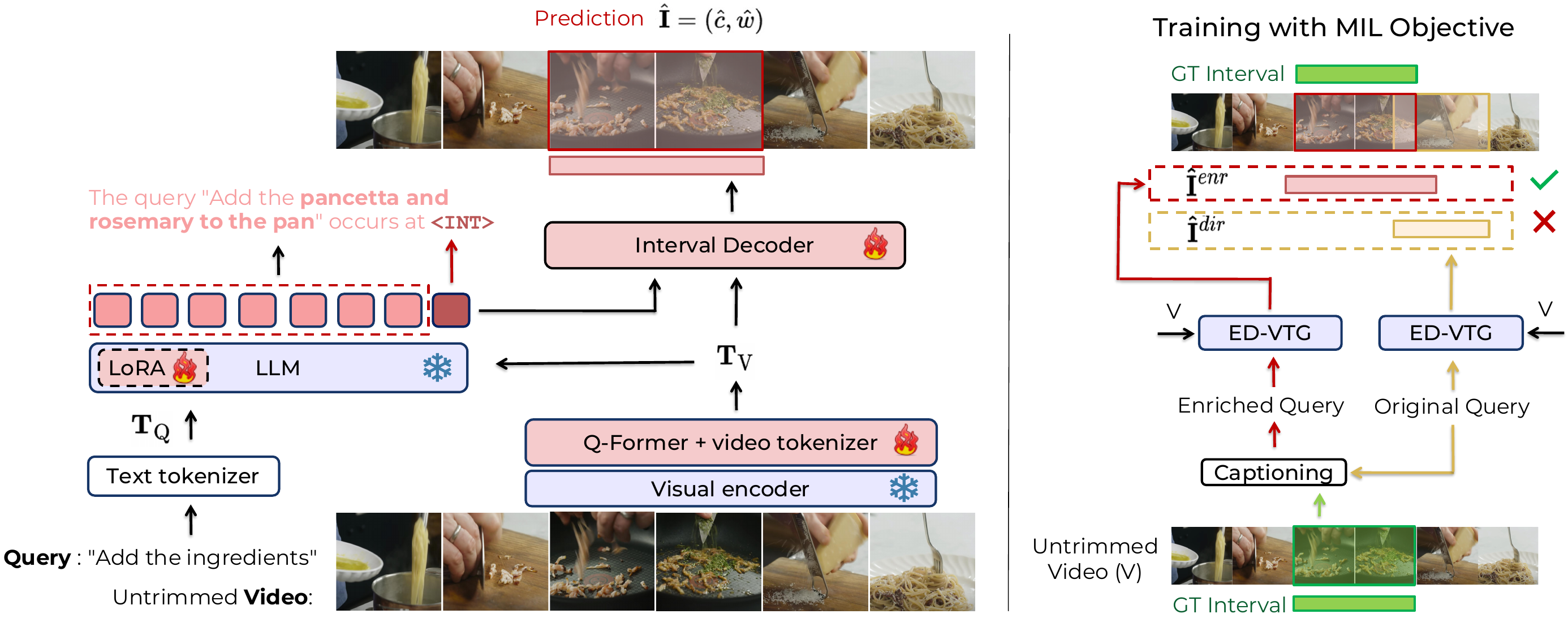}
\vspace{-3mm}
\vspace{-3mm}
\caption{\textbf{Overview of the proposed Enrich and Detect framework}: \emph{(left)} \model\ pipeline: Given an untrimmed video and a query $Q$ to be grounded, the inputs are first tokenized into video tokens $T_V$ and text tokens $T_Q$. The tokens are then fed into an LLM, which first generates an enriched query by, e.g., filling in any missing details and then emitting an interval token $\texttt{<INT>}$. The embedding of this special token is finally decoded into the predicted temporal interval via a lightweight interval decoder. In the example shown here, the vague input query is enriched into a more detailed one by our model which can be subsequently grounded more easily. \emph{(right)} Training: \model\ is trained using ground-truth temporal intervals and pseudo-labels of enriched queries, generated by an external off-the-shelf captioning model \cite{li2024llavaonevision}, which -- unlike our model at inference time -- has access to the ground-truth intervals. For every sample during training, the proposed multiple instance learning (MIL) framework allows \model\ to assess both the original or the enriched queries and generate two sets of predictions, $\hat{I}^{enr}$, the interval predicted using the enriched query, and $\hat{I}^{dir}$ using the original query. Next, the model backpropagates using the better prediction (i.e., lower grounding loss). Hence, during training, \model\ dynamically learns to decide for which sample enrichment is necessary and, based on that, performs detection.}
\vspace{-4mm}
\label{fig:method}
\end{figure*}

\xpar{Prompt augmentation with LLMs}
\EM{Off-the-shelf LLMs have recently been used to augment input prompts for tasks such as image retrieval and image classification by providing additional, clarifying descriptions. These augmented descriptions can aid generalization in various vision and NLP tasks such as visual question answering~\cite{Prasad2023RephraseAR,Teney21,Dong2017LearningTP}, dialog generation~\cite{gao-etal-2020-paraphrase}, and visual classification~\cite{MenonICLR23}.}
\EM{Our approach goes beyond using an off-the-shelf LLM for offline query enrichment, rather learning to dynamically enrich queries during inference time towards better grounding.}

Prior work has also investigated the use of captioners and language models for data augmentation by rephrasing existing descriptions or providing new improved ones. LaViLa\cite{zhao2022lavila} uses rephraser and narrator models to improve the quality of the training data for video-text alignment.
\EM{Augmentation during training is sufficient for that task, as the goal is to learn joint representations; for our grounding task however simply augmenting the training set, without enriching queries during inference is not as effective, as we empirically demonstrate.}

\xpar{Multiple instance learning}
Multiple instance learning (MIL)\cite{dietterich199731} is a technique commonly used in weakly-supervised vision problems, when a collection of potential solutions is available but exact annotations are not.
It has been successfully applied to a range of tasks, including classification~\cite{andrews2003support}, weakly supervised object detection\cite{bilen2016weakly,cinbis2017weakly} and temporal action localization\cite{paul2018w,nguyen2019weakly}.

\vspace{-1mm}
\section{Method}
\vspace{-1mm}

Given an untrimmed video $\mathcal{V}$ and a set of $N$ associated textual queries $\mathcal{Q}=\{Q_1, \ldots, Q_N\}$, temporal grounding aims to identify the corresponding temporal interval for each query. Formally, the output is a set of temporal intervals $\mathcal{I}=\{I_1, \ldots, I_N\}$. For $N=1$, the task takes the form of single-query temporal grounding, which we will simply refer to as STG. 
For clarity, we describe our approach in the STG setting, without loss of generality; the pipeline however readily extends to $N>1$.

Our approach aims to tackle temporal grounding by leveraging a multimodal LLM that $(i)$ transforms input queries into intermediate, enriched queries by adding missing details based on the video input, $(ii)$ generates contextualized embeddings of each latent segment to enable temporal interval prediction,  $(iii)$ decodes these embeddings into concrete temporal boundaries.

In the following, we formally introduce our proposed ED-VTG model in Section~\ref{sec:vtg}, and proceed to describe how to train it with enriched query pseudo-labels within a MIL framework (Section ~\ref{sec:training}).   

\vspace{-0.5mm}
\subsection{Model}
\label{sec:vtg}

Our model (Figure \ref{fig:method}) consists of three key modules: a vision encoder that extracts video representations,  a LLM  that jointly processes video and language, and a lightweight interval decoder that generates precise temporal boundaries.

\vspace{-5mm}
\paragraph{Enrich.} Given a single input query $Q$ about a video $V$ with $T$ frames, the vision encoder represents the video as a sequence of $R$ visual tokens \( \mathbf{T}_\mathrm{V} \in \mathbb{R}^{R\times D} \), where $D$ is the token dimension. The tokenized video features \( \mathbf{T}_\mathrm{V} \) are fed along with the tokenized query \( \mathbf{T}_\mathcal{Q} \) to the LLM, 
which generates an enriched query $\hat{Q}^{enr} = \{ \hat{y}_1, ..., \hat{y}_l, ... \hat{y}_{L^{enr}} \}$ one token at a time:
\begin{align}
\hat{y}_l = \mathcal{F}_\mathrm{LLM}(\hat{y}_{<l}, \mathbf{T}_\mathrm{V}, \mathbf{T}_\mathcal{Q}).
\label{eq:interval_ar}
\end{align}
When the model is ready to ground the query, the LLM emits a new, special token $\texttt{<INT>}$ to trigger interval prediction.
In other words, the text prediction takes the form:
\vspace{-6mm}
\begin{align}
\mathbf{\hat{y}} = \textit{``The query $\hat{Q}^{enr}$ occurs at  \(  \texttt{<INT>}\)''}
\end{align}
\vspace{-4mm}

\vspace{-6mm}
\paragraph{Detect.} To detect the temporal interval corresponding to the enriched query $\hat{Q}^{enr}$, we introduce an interval decoder $\mathcal{F}_\mathrm{dec}$ that predicts an interval $\hat{\mathbf{I}}$ parametrized by the center $\hat{c}$ and width $\hat{w}$ of the predicted interval. We selected this parameterization for its advantage in decoupling position from scale, as supported by literature in object detection~\cite{redmon2016you,Redmon2016YOLO9000BF,liu2016ssd,zhou2019objects}. The interval decoder takes the form
\begin{align}
\label{eq:interval_cw}
\hat{\mathbf{I}} = (\hat{c}, \hat{w}) = \mathcal{F}_\mathrm{dec}\left( \mathcal{G}(\mathbf{h}_\mathrm{int}), \mathbf{T}_\mathrm{V} \right),
\end{align}
where $\mathbf{h}_{int} \in \mathbb{R}^{D}$ is the hidden state of the LLM corresponding to the $\texttt{<INT>}$ token and $\mathcal{G}$ is a linear projection layer. The decoder, functioning as the regression component of a temporal detector, consists of two transformer layers followed by a multi-layer perceptron (MLP), which processes a concatenation of its two inputs, $(\mathcal{G}(\mathbf{h}_\mathrm{int}), \mathbf{T}_\mathrm{V})$. It finally outputs the predicted interval \(\hat{\mathbf{I}}\), grounding the input query \(\mathcal{Q}\).

Notice that our formulation allows the quality of the enriched query $\hat{Q}^{enr}$ to directly influence the accuracy of the predicted interval $\hat{\mathbf{I}}$. This is because the generation of the interval token \texttt{<INT>} (and consequently its hidden state $\mathbf{h}_\mathrm{int}$) is conditioned on the enriched query prediction. Establishing this \textit{cascaded} dependency chain, i.e., $(V, Q)$ $\rightarrow$  $\hat{Q}^{enr}$ and $(V, \hat{Q}^{enr}$) $\rightarrow$ $\hat{I}$, is the key idea of our approach.

\subsection{Training} 
\label{sec:training}

The model is trained end-to-end using two primary loss functions: a language modeling loss \(\mathcal{L}_\mathrm{LM}\) and a temporal grounding loss \(\mathcal{L}_\mathrm{grnd}\) that help supervise the ``enrich'' and ``detect'' aspects of our model, respectively.

\vspace{-3mm}
\paragraph{Enrich.} Given the target output text \(\mathbf{y} = \{y_1, y_2, \ldots, y_L\}\), \(\mathcal{L}_\mathrm{LM}\) is calculated as the cross-entropy loss that evaluates the likelihood of $\mathbf{y}$ under the predicted probability distribution generated by the model:
\vspace{-3mm}
\begin{align}
\mathcal{L}_\mathrm{LM} = -\sum_{t=1}^{T} \log P(y_l \mid \mathbf{y}_{<l}, \mathbf{T}_\mathrm{V}, \mathbf{T}_\mathcal{Q})
\end{align}
\vspace{-3mm}

To provide a proper supervisory signal for query enrichment, we need the ground-truth pair of $(Q, Q^{enr})$. However, such datasets do not exist, nor is it practical to annotate a dataset solely for this purpose. Here, we capitalize on the tremendous amount of progress made in the video captioning literature~\cite{sharma2020image,stefanini2022show,abdar2023review}, and use an off-the-shelf captioning model~\cite{li2024llavaonevision} to generate pseudo ground-truth $Q^{enr}$ by refining the original query $Q$ given its video V (see more details in Section~\ref{sec:pretraining_downstream_dataset}).

\vspace{-4mm}
\paragraph{Detect.} Given a target temporal interval \( I = (c, w)\), the grounding loss \(\mathcal{L}_\mathrm{grnd}\) is computed as a combination of the L1 loss and the generalized Intersection over Union (gIoU) loss~\cite{Rezatofighi2019GeneralizedIO}, applied on the predicted temporal interval \((\hat{c}, \hat{w})\):
\vspace{-2mm}
\begin{align}
\mathcal{L}_\mathrm{grnd} = & \ \lambda_\mathrm{L1} (\left| (\hat{c} - c \right| + \left| \hat{w} - w \right|) \nonumber \\
+ & \ \lambda_\mathrm{gIoU} \, \mathrm{gIoU}((\hat{c}, \hat{w}), (c, w))
\end{align}


\begin{table}[!t]
\centering

\small
\setlength{\tabcolsep}{4pt}
\resizebox{\columnwidth}{!}{\begin{tabular}{@{} l c c | c c | c c c @{}}

\toprule
\multirow{2}{*}{\textbf{Dataset}} & \multirow{2}{*}{\textbf{Domain}} & \multirow{2}{*}{\textbf{Tasks}} & \multirow{2}{*}{\textbf{Corpus}} & \multirow{2}{1.07 cm}{\centering \bf Eval. Protocol} & \multirow{2}{1.1 cm}{\centering \bf \# Train Samples} & \multirow{2}{1.4 cm}{\centering \bf Avg Vid Length \textit{(s)}} & \multirow{2}{1.4 cm}{\centering \bf Avg Span Length \textit{(s)}} \\ 
 
& & & & & & & \\

\midrule 

DiDeMo \cite{anne2017didemo} & Open & \VTG & PT & $-$ & 32.8K & 54.57 & 6.49 \scriptsize{\textcolor{ourbrown}{(11.9\%)}}  \\
QuerYD \cite{oncescu2021queryd} & Cooking & \VTG & PT & $-$ & 13.6K & 158.78 & 7.68 \scriptsize{\textcolor{ourbrown}{(4.8\%)}} \\
COIN \cite{tang2019coin} & Open & VPG & PT & $-$ & 7.5K & 143.71 & 15.06 \scriptsize{\textcolor{ourbrown}{(10.5\%)}} \\
HiREST \cite{zala2023hirest} & Open & \VTG, VPG & PT & $-$ & 0.8K & 208.03 & 44.50 \scriptsize{\textcolor{ourbrown}{(21.4\%)}} \\
VITT$^{\dagger}$ \cite{huang2020vitt} & Open & VPG & PT & $-$ & 4.9K & 287.17 & $-$ \\
YTTemporal \cite{zellers2022yttemporal} & Open & VPG & PT & $-$ & 28.8K & 327.36 & 4.0 \scriptsize{\textcolor{ourbrown}{(1.2\%)}} \\
CrossTask \cite{zhukov2019crosstask} & Procedural & AG & PT & $-$ & 2.7K & 297.0 & 9.61 \scriptsize{\textcolor{ourbrown}{(3.2\%)}} \\
VideoCC \cite{nagrani2022videocc} & Open & \VTG & PT & $-$ & 45.0K & 415.89 & 9.88 \scriptsize{\textcolor{ourbrown}{(2.3\%)}} \\
\midrule 
Charades-STA \cite{gao2017charadessta} & Indoor & \VTG & FT, Eval & ZS, FT & 12.4K & 31.17 & 8.29 \scriptsize{\textcolor{ourbrown}{(26.6\%)}}  \\
Charades-CD-OOD \cite{yuan2021charades-cd} & Indoor & VPG & FT, Eval & FT & 4.5K & 30.60 & 7.90 \scriptsize{\textcolor{ourbrown}{(25.8\%)}} \\
ANet-Captions \cite{krishna2017dense} & Open & \VTG, VPG & FT, Eval & ZS, FT & 9.5K & 117.63 & 35.61 \scriptsize{\textcolor{ourbrown}{(30.3\%)}} \\
TACoS \cite{regneri2013tacos} & Cooking & \VTG, VPG & FT, Eval & ZS, FT & 9.8K & 224.34 & 23.33 \scriptsize{\textcolor{ourbrown}{(10.4\%)}} \\
YouCook2 \cite{zhou2018youcook2} & Cooking & VPG & FT, Eval & FT & 1.2K & 311.41 & 20.07 \scriptsize{\textcolor{ourbrown}{(6.4\%)}} \\
NExT-GQA$^{\diamond}$ \cite{xiao2024nextgqa} & Open & QG & Eval & ZS & $-$ & 39.60 & 6.69 \scriptsize{\textcolor{ourbrown}{(16.9\%)}} \\
HT-Step \cite{afouras2024htstep} & Cooking & AG & FT, Eval & FT & 17.4K & 393.89 & 14.88 \scriptsize{\textcolor{ourbrown}{(3.7\%)}} \\
\bottomrule
\end{tabular}}
\vspace{-2mm}
\caption{\textbf{Dataset statistics, corresponding tasks, and evaluation protocol. The upper side} of the table represents datasets used for pre-training, resulting in a total of 136K samples. \textbf{The lower side} represents datasets used for fine-tuning and evaluation. We cover four different video grounding tasks: 
single-query temporal grounding (\VTG), video paragraph grounding (VPG), question grounding (QG), and article grounding (AG). QG is used only for evaluation to assess the model’s generalization capability. Average interval lengths compared to the corresponding video durations are shown in \textcolor{ourbrown}{brown}, denoting the annotation granularity. $^{\dagger}$VITT contains single timestamp annotation instead of intervals. $^{\diamond}$NExT-GQA contains only evaluation split.} 
\label{tab:dataset_details}
\vspace{-4mm}
\end{table}

\begin{table*}[!t]
\centering

\small
\setlength{\tabcolsep}{4pt}
\resizebox{0.98\textwidth}{!}{\begin{tabular}{l c c c | c c c c | c c c c | c c c c}

\toprule 

\multirow{2}{*}{\bf Method} & \multirow{2}{1.5 cm}{\bf \centering Generalist Model} & \multirow{2}{1.4cm}{\bf \centering \# Train Samples} & \multirow{2}{*}{\bf \centering Eval.} & \multicolumn{4}{c|}{\bf Charades-STA} & \multicolumn{4}{c|}{\bf ActivityNet-Captions} & \multicolumn{4}{c}{\bf TACoS} \\ 

& & & & R@0.3 & R@0.5 & R@0.7 & mIoU & R@0.3 & R@0.5 & R@0.7 & mIoU & R@0.3 & R@0.5 & R@0.7 & mIoU  \\

\midrule 


UniVTG \cite{lin2023univtg} & \ding{55} & 4.2M & ZS & 44.1 & 25.2 & 10.0 & 27.1 & $-$ & $-$ & $-$ & $-$ & 5.2 & 1.3 & 0.3 & 4.4 \\
SeViLA \cite{yu2024sevila} & \ding{55} & 129M & ZS & $-$ & $-$ & $-$ & $-$ & 31.6 & 19.0 & 10.1 & 23.0 & $-$ & $-$ & $-$ & $-$ \\
PSVL \cite{nam2021psvl} & \ding{55} & $-$ & ZS & 46.2 & 31.3 & 14.2 & 31.2 & 44.7 & 30.1 & 14.7 & 29.6 & $-$ & $-$ & $-$ & $-$ \\
LT-ZVG \cite{kim2023ltzvg} & \ding{55} & $-$ & ZS & 52.9 & 37.2 & 19.3 & 36.0 & 47.6 & 32.6 & 15.4 & 31.8 & $-$ & $-$ & $-$ & $-$ \\

\midrule

Video-LLaMA$^{\diamond}$ \cite{zhang2023videollama} & \ding{51} & 2.7M & ZS & 25.2 & 10.6 & 3.4 & 16.8 & 21.9 & 10.8 & 4.9 & 16.5 & 5.1 & 1.2 & 0.8 & 3.4 \\
Video-ChatGPT$^{\diamond}$ \cite{Maaz2023videochatgpt} & \ding{51} & 100K & ZS & 27.2 & 6.2 & 1.9 & 19.7 & 19.5 & 10.6 & 4.8 & 14.2 & 6.3 & 1.7 & 1.0 & 4.3 \\
Valley \cite{luo2023valley} & \ding{51} & 100K & ZS & 28.4 & 1.8 & 0.3 & 21.4 & 30.6 & 13.7 & 8.1 & 21.9 & $-$ & $-$ & $-$ & $-$ \\
VideoChat2 \cite{li2024videochat2} & \ding{51} & 2M & ZS & 38.0 & 14.3 & 3.8 & 24.6 & 40.8 & 27.8 & 9.3 & 27.9 & $-$ & $-$ & $-$ & $-$ \\
Momenter \cite{qianmomentor} & \ding{51} & 10M & ZS & 42.6 & 26.6 & 11.6 & 28.5 & 42.9 & 23.0 & 12.4 & 29.3 & $-$ & $-$ & $-$ & $-$ \\
VTimeLLM$^{\diamond}$ \cite{huang2024vtimellm} & \ding{51} & 170K & ZS & 51.0 & 27.5 & 11.4 & 31.2 & 44.0 & 27.8 & \underline{14.3} & 30.4 & 7.0 & 1.8 & 0.8 & 4.5 \\
TimeChat$^{\diamond}$ \cite{ren2024timechat} & \ding{51} & 125K & ZS & $-$ & 32.2 & 13.4 & $-$ & $-$ & $-$ & $-$ & $-$ & 6.8 & 2.1 & 0.8 & 4.7 \\
HawkEye \cite{wang2024hawkeye} & \ding{51} & 715K & ZS & 50.6 & 31.4 & 14.5 & 33.7 & \underline{49.1} & \underline{29.3} & 10.7 & \underline{32.7} & $-$ & $-$ & $-$ & $-$ \\
ChatVTG \cite{qu2024chatvtg} & \ding{51} & 100K & ZS & \underline{52.7} & \underline{33.0} & \underline{15.9} & \underline{34.9} & 40.7 & 22.5 & 9.4 & 27.2 & \underline{8.1} & \underline{3.7} & \underline{1.3} & \underline{5.5} \\

\rowcolor{Light}
\model & \ding{51} & 136K & ZS & \bf 59.5 & \bf 39.3 & \bf 19.8 & \bf 40.2 & \bf 52.1 & \bf 33.1 & \bf 16.0 & \bf 35.2 & \bf 14.5 & \bf 6.0 & \bf 2.3 & \bf 12.7 \\

\midrule

\bf \textcolor{blue}{$\Delta_{\text{Ours - HawkEye}}$} & $-$ & $-$ & ZS & \textcolor{blue}{8.9} \textcolor{blue}{$\uparrow$} & \textcolor{blue}{7.9} \textcolor{blue}{$\uparrow$} & \textcolor{blue}{5.3} \textcolor{blue}{$\uparrow$} & \textcolor{blue}{6.5} \textcolor{blue}{$\uparrow$} & \textcolor{blue}{3.0} \textcolor{blue}{$\uparrow$} & \textcolor{blue}{3.8} \textcolor{blue}{$\uparrow$} & \textcolor{blue}{5.3} \textcolor{blue}{$\uparrow$} & \textcolor{blue}{2.5} \textcolor{blue}{$\uparrow$} &\textcolor{blue}{$-$} & \textcolor{blue}{$-$} & \textcolor{blue}{$-$}& \textcolor{blue}{$-$} \\
\bf \textcolor{blue}{$\Delta_{\text{Ours - ChatVTG}}$} & $-$ & $-$ & ZS & \textcolor{blue}{$-$} & \textcolor{blue}{$-$} & \textcolor{blue}{$-$} & \textcolor{blue}{$-$} & \textcolor{blue}{$-$} & \textcolor{blue}{$-$} & \textcolor{blue}{$-$}& \textcolor{blue}{$-$} & \textcolor{blue}{6.4} \textcolor{blue}{$\uparrow$} & \textcolor{blue}{2.3} \textcolor{blue}{$\uparrow$} & \textcolor{blue}{1.0} \textcolor{blue}{$\uparrow$} & \textcolor{blue}{7.2} \textcolor{blue}{$\uparrow$} \\

\bottomrule

\end{tabular}}
\vspace{-2mm}
\caption{\textbf{Zero-shot STG results on Charades, ActivityNet, and TACoS test splits.} For all three datasets, \model\ gains significant improvement over \emph{all} existing methods, including task-specific, non-generalist models.
We use \textbf{boldface} for the best and \underline{underline} the second-best result for each metric, among the generalist models. 
$^{\diamond}$Official checkpoints are used for TACoS evaluation.}
\label{tab:stg_zero_shot}
\vspace{-5mm}
\end{table*}

\vspace{-4mm}
\subsubsection{MIL framework}
\label{sec:mil_training}
\EM{A caveat with the pseudo-labeled enriched queries is that they can be noisy and include hallucinations; as a result, some of them may lead to a temporal interval prediction inferior to the original query. To address this, we propose considering \textit{multiple} options for the target query (2 in the case of a single query scenario, $Q$ and $Q^{enr}$), by adopting a} multiple instance learning (MIL) framework~\cite{dietterich199731}. This approach allows the model to choose between the enriched and the original input queries during training, depending on which version \EM{leads to a better temporal interval prediction. During inference, this means that our ED-VTG model can choose to either enrich the query if it misses important details, or decide to ``carry over'' the original query into $\hat{Q}^{enr}$ when it is concrete enough (in which case $Q = \hat{Q}^{enr}$). 
}

Formally, to perform MIL, we collect two temporal interval predictions by running two forward passes with different LLM inputs (in a teacher-forcing fashion), \ie~$\mathbf{y}^\mathrm{dir}$ that is formed using the original query $\mathcal{Q}$, and $\mathbf{y}^\mathrm{enr}$ using the pseudo-labeled enriched query $\mathcal{Q}^{enr}$. 
We illustrate this schematically in Figure~ \ref{fig:method} (right).
Recall that through the contextualized interval representation \(\mathbf{h}_\mathrm{int}\), the predicted interval depends on the LLM teacher-forced input. Hence, these inputs produce respective interval predictions, \(\hat{\mathbf{I}}^\mathrm{dir}\) and \(\hat{\mathbf{I}}^\mathrm{enr}\). We select the query version that results in the smallest grounding loss and use it to compute our overall training objective:
\vspace{-2mm}
\begin{align}
\label{eq:overall_objective}
\mathcal{L} =
  \begin{cases}
    \lambda_\mathrm{LM} \mathcal{L}_\mathrm{LM}^\mathrm{dir} + \lambda_\mathrm{grnd} \mathcal{L}_\mathrm{grnd}^\mathrm{dir} & \; \text{if } \mathcal{L}_\mathrm{grnd}^\mathrm{dir} < \mathcal{L}_\mathrm{grnd}^\mathrm{enr} \\
    \lambda_\mathrm{LM} \mathcal{L}_\mathrm{LM}^\mathrm{enr} + \lambda_\mathrm{grnd} \mathcal{L}_\mathrm{grnd}^\mathrm{enr} & \; \text{otherwise}
  \end{cases}
\end{align}
where $\mathcal{L}_\mathrm{LM}^\mathrm{dir}$ and $\mathcal{L}_\mathrm{LM}^\mathrm{enr}$ are the language losses for targets $\mathbf{y}^\mathrm{dir}$ and $\mathbf{y}^\mathrm{enr}$ respectively, and 
$\mathcal{L}_\mathrm{grnd}^\mathrm{dir}$ and
$\mathcal{L}_\mathrm{grnd}^\mathrm{enr}$ the grounding losses for predictions \(\hat{\mathbf{I}}^\mathrm{dir}\) and \(\hat{\mathbf{I}}^\mathrm{enr}\) respectively.
The hyperparameters \(\lambda_\mathrm{LM}\) and \(\lambda_\mathrm{grnd}\) are the relative weights of the language modeling and grounding losses. \vspace{-2mm} 

\section{Experiments}

\vspace{-2mm}

We design experiments to study three key questions related to our architecture and training framework: \textbf{Q1)\;}How does \model\ perform on various video grounding tasks in comparison to the current state-of-the-art? \textbf{Q2)\;}How beneficial is our query enrichment approach, within the MIL paradigm, as opposed to directly grounding the original queries? \textbf{Q3)\;}Does utilizing an interval decoder with specifically tailored grounding objectives offer advantages over predicting timestamps as raw text or special tokens?

\begin{table*}[!t]
\centering

\small
\setlength{\tabcolsep}{4pt}
\resizebox{0.98\textwidth}{!}{\begin{tabular}{l c c c | c c c c | c c c c | c c c c}

\toprule

\multirow{2}{*}{\bf Method} & \multirow{2}{1.5 cm}{\bf \centering Generalist Model} & \multirow{2}{1.4cm}{\bf \centering \# Train Samples} & \multirow{2}{*}{\bf \centering Eval.} & \multicolumn{4}{c|}{\bf Charades-STA} & \multicolumn{4}{c|}{\bf ActivityNet-Captions} & \multicolumn{4}{c}{\bf TACoS} \\

& & & & R@0.3 & R@0.5 & R@0.7 & mIoU & R@0.3 & R@0.5 & R@0.7 & mIoU & R@0.3 & R@0.5 & R@0.7 & mIoU  \\

\midrule

\demph{VSLNet (C3D) \cite{zhang2020vslnet}} & \demph{\ding{55}} & \demph{$-$} & \demph{FT} & \demph{64.3} & \demph{47.3} & \demph{30.2} & \demph{45.2} & \demph{63.2} & \demph{43.2} & \demph{26.2} & \demph{43.2} & \demph{29.6} & \demph{24.3} & \demph{20.0} & \demph{24.1} \\
\demph{MS-2D-TAN (I3D) \cite{zhang2021ms2dtan}} & \demph{\ding{55}} & \demph{$-$} & \demph{FT} & \demph{$-$} & \demph{56.6} & \demph{36.2} & \demph{$-$} & \demph{62.1} & \demph{45.5} & \demph{28.3} & \demph{$-$} & \demph{42.0} & \demph{33.6} & \demph{22.1} & \demph{$-$} \\
\demph{Moment-DETR \cite{lei2021momentdetr}} & \demph{\ding{55}} & \demph{236K} & \demph{FT} & \demph{65.8} & \demph{52.1} & \demph{30.6} & \demph{45.5} & \demph{$-$} & \demph{$-$} & \demph{$-$} & \demph{$-$} & \demph{38.0} & \demph{24.7} & \demph{12.0} & \demph{25.5} \\
\demph{UnLoc-B \cite{yan2023unloc}} & \demph{\ding{55}} & \demph{650K} & \demph{FT} & \demph{$-$} & \demph{58.1} & \demph{35.4} & \demph{$-$} & \demph{$-$} & \demph{48.0} & \demph{29.7} & \demph{$-$} & \demph{$-$} & \demph{$-$} & \demph{$-$} & \demph{$-$} \\
\demph{MomentDiff \cite{li2024momentdiff}} & \demph{\ding{55}} & \demph{$-$} & \demph{FT} & \demph{$-$} & \demph{55.6} & \demph{32.4} & \demph{$-$} & \demph{$-$} & \demph{$-$} & \demph{$-$} & \demph{$-$} & \demph{46.6} & \demph{28.9} & \demph{12.4} & \demph{30.4} \\
\demph{LGI \cite{mun2020lgi}} & \demph{\ding{55}} & \demph{$-$} & \demph{FT} & \demph{73.0} & \demph{59.5} & \demph{35.5} & \demph{51.4} & \demph{58.5} & \demph{41.5} & \demph{23.1} & \demph{41.1} & \demph{$-$} & \demph{$-$} & \demph{$-$} & \demph{$-$} \\
\demph{BAM-DETR \cite{lee2024bamdetr}} & \demph{\ding{55}} & \demph{$-$} & \demph{FT} & \demph{72.9} & \demph{60.0} & \demph{39.4} & \demph{52.3} & \demph{$-$} & \demph{$-$} & \demph{$-$} & \demph{$-$} & \demph{56.7} & \demph{41.5} & \demph{26.8} & \demph{39.3} \\
\demph{InternVideo2$^{\star}$ + CG-DETR \cite{wang2024internvideo2}} & \demph{\ding{55}} & \demph{2.1M} & \demph{FT} & \demph{\bf 79.7} & \demph{70.0} & \demph{48.9} & \demph{58.8} & \demph{$-$} & \demph{$-$} & \demph{$-$} & \demph{$-$} & \demph{$-$} & \demph{$-$} & \demph{$-$} & \demph{$-$} \\


\demph{SG-DETR \cite{gordeev2024sgdetr}} & \demph{\ding{55}} & \demph{$-$} & \demph{FT} & \demph{$-$} & \demph{\bf 71.1} & \demph{\bf 52.8} & \demph{60.7} & \demph{$-$} & \demph{$-$} & \demph{$-$} & \demph{$-$} & \demph{$-$} & \demph{46.4} & \demph{33.9} & \demph{42.4} \\



\demph{EMB (ELA) \cite{huang2022emb}} & \demph{\ding{55}} & \demph{$-$} & \demph{FT} & \demph{\bf 79.7} & \demph{69.2} & \demph{51.4} & \demph{\bf 62.2} & \demph{\bf 73.7} & \demph{\bf 58.7} & \demph{\bf 40.7} & \demph{\bf 56.2} & \demph{\bf 63.3} & \demph{\bf 52.5} & \demph{\bf 37.0} & \demph{\bf 48.4} \\

\midrule

BLIP-2 (frames only) \cite{li2023blip2} & \ding{51} & 129M & FT & $-$ &  43.3 & \underline{32.6} & $-$ & $-$ & 25.8 & 9.7 & $-$ & $-$ & $-$ & $-$ & $-$ \\
VideoChat2 \cite{li2024videochat2} & \ding{51} & 2M & FT & $-$ & $-$ & $-$ & $-$ & 55.5 & \underline{34.7} & 17.7 & 38.9 & $-$ & $-$ & $-$ & $-$ \\
TimeChat \cite{ren2024timechat} & \ding{51} & 125K & FT & $-$ & 46.7 & 23.7 & $-$ & $-$ & $-$ & $-$ & $-$ & \underline{27.7} & \underline{15.1} & \underline{6.4} & \underline{18.4} \\
HawkEye \cite{wang2024hawkeye} & \ding{51} & 715K & FT & \underline{72.5} & \underline{58.3} & 28.8 & \underline{49.3} & \underline{55.9} & \underline{34.7} & \underline{17.9} & \underline{39.1} & $-$ & $-$ & $-$ & $-$ \\
VtimeLLM \cite{huang2024vtimellm} & \ding{51} & 170K & FT & $-$ & $-$ & $-$ & $-$ & $-$ & $-$ & $-$ & $-$ & 26.8 & 14.4 & 6.1 & 18.0 \\

\rowcolor{Light}
\model & \ding{51} & 136K & FT & \bf 78.2 & \bf 62.1 & \bf 35.0 & \bf 52.6 & \bf 67.6 & \bf 45.1 & \bf 22.7 & \bf 44.9 & \bf 46.0 & \bf 31.5 & \bf 15.8 & \bf 32.4 \\

\midrule

\bf \textcolor{blue}{$\Delta_{\text{Ours - HawkEye}}$} & $-$ & $-$ & FT & \textcolor{blue}{5.7} \textcolor{blue}{$\uparrow$} & \textcolor{blue}{3.8} \textcolor{blue}{$\uparrow$} & \textcolor{blue}{6.2} \textcolor{blue}{$\uparrow$} & \textcolor{blue}{3.3} \textcolor{blue}{$\uparrow$} & \textcolor{blue}{11.7} \textcolor{blue}{$\uparrow$} & \textcolor{blue}{10.4} \textcolor{blue}{$\uparrow$} & \textcolor{blue}{4.8} \textcolor{blue}{$\uparrow$} & \textcolor{blue}{5.8} \textcolor{blue}{$\uparrow$} &\textcolor{blue}{$-$} & \textcolor{blue}{$-$} & \textcolor{blue}{$-$}& \textcolor{blue}{$-$} \\
\bf \textcolor{blue}{$\Delta_{\text{Ours - VTimeLLM}}$} & $-$ & $-$ & FT & \textcolor{blue}{$-$} & \textcolor{blue}{$-$} & \textcolor{blue}{$-$} & \textcolor{blue}{$-$} & \textcolor{blue}{$-$} & \textcolor{blue}{$-$} & \textcolor{blue}{$-$}& \textcolor{blue}{$-$} & \textcolor{blue}{19.2} \textcolor{blue}{$\uparrow$} & \textcolor{blue}{17.1} \textcolor{blue}{$\uparrow$} & \textcolor{blue}{9.7} \textcolor{blue}{$\uparrow$} & \textcolor{blue}{14.4} \textcolor{blue}{$\uparrow$} \\

\bottomrule

\end{tabular}}
\vspace{-3mm}
\caption{\textbf{Fine-tuned STG results on Charades, ActivityNet, and TACoS test splits.} For all datasets, \model\ achieves strong improvements over previous generalist models, and performs comparably to task-specific expert models. $^{\star}$Though InterVideo2 is a generalist model, it fine-tunes CG-DETR \cite{moon2023cgdetr} head for grounding tasks, using the LLM as a feature extractor. For completeness, we report all competitive existing works, including task-specific SOTA specialist models (shown in gray),
but directly compare \model \ only with generalist frameworks - using \textbf{boldface} for the best and \underline{underlining} the second best results among generalist models (see discussion in Section \ref{sec:evaluation_protocol}).}
\label{tab:stg_fine_tune}
\vspace{-5.5mm}
\end{table*}


\ys{Perhaps describe PT and FT here upfront to avoid confusion? We can mention why we do multi-task PT along with potential benefits.}

\vspace{-1mm}
\subsection{Datasets} \label{sec:pretraining_downstream_dataset}

Table \ref{tab:dataset_details} summarizes the datasets that we used during the pre-training and fine-tuning stages. During pre-training, we use a total of 136K medium-to-long duration videos from 8 public datasets annotated with text queries and the corresponding intervals. As a preprocessing step, we \EM{collect pseudo-labels for enriched queries}
using an external captioning model~\cite{li2024llavaonevision}. In short, we take each video segment defined by the ground-truth intervals and prompt the captioning model to enrich the original query while preserving its meaning given the video segment. We provide full details of this process with the exact prompt used in the supplementary material.

\vspace{-1mm}

\subsection{Tasks} \label{sec:pretraining_downstream_tasks}
\vspace{-1mm}

Table \ref{tab:dataset_details} also summarizes the tasks
for which we use each dataset. Here we briefly 
describe these tasks, and highlight how our approach is applied to solve them.

\textbf{Single-Query Temporal Grounding (STG)} involves identifying a single time window in response to a \EM{single input} language query \EM{(N=1)}.

\textbf{Video Paragraph Grounding (VPG)} involves grounding $N>1$ sentences to $N$ corresponding time windows.
\EM{Our ED-VTG model can be trained to predict multiple enriched queries (one per original input query) interleaved with $\texttt{<INT>}$ tokens that get separately decoded into intervals.}
Since there are multiple queries in the input, we run multiple forward passes through the LLM to perform MIL, wherein each pass selects a random number of queries to be enriched.

\textbf{Question Grounding (QG)} involves retrieving \textit{evidence intervals} to answer questions, facilitating explainable video QA. \EM{As in STG, the input is a single query, and the output a single interval.}

\textbf{Article grounding (AG)} is an extension of VPG where the model is given multiple queries as input, some of which may not be groundable in the video. Hence, the model must 1) identify which queries are groundable and 2) predict intervals for the groundable queries.

Note that, while we train our model jointly on multiple tasks, they are unified as a single task in the form of (Video, Text) $\rightarrow$ (Time intervals). Besides the differences in training data sources, we use the identical training objective (Eq.~\ref{eq:overall_objective}) during pre-training and fine-tuning stages.

\subsection{Evaluation Protocol}
\label{sec:evaluation_protocol}
Following the common practices in the literature \cite{ren2024timechat, huang2024vtimellm, wang2024hawkeye}, we evaluate \model\ in two primary evaluation protocols: $(i)$ \textbf{Zero-shot (ZS)}, where the pre-trained model is assessed directly without any \EM{fine-tuning on downstream datasets},
and $(ii)$ \textbf{Fine-tuned (FT)}, where the model undergoes additional training on specific tasks and datasets.
We also present results without pre-training to assess its importance.
For the \VTG\ and VPG tasks, following existing works \cite{ren2024timechat, huang2024vtimellm, wang2024hawkeye, bao2021depnet, rodriguez2023locformer}, we report the mean
intersection over union (mIoU) and Recall@1 \EM{for} IoU$\geq m$ (R@$m$), with $m \in\{$0.3, 0.5, 0.7$\}$. For the QG task, we report intersection over prediction (IoP) \EM{in addition to}
IoU, following the NeXT-GQA \cite{xiao2024nextgqa} evaluation protocol. For AG, we adhere to the HT-Step \cite{afouras2024htstep} protocol and report article-grounding mAP scores across various IoU thresholds.

\noindent \textbf{Comparison to non-generalist models.}
In the results section, we report all existing SOTA and competitive methods for a complete comparison, however we divide them into generalist and non-generalist (specialist) models.
We note that specialist methods are heavily tailored to the task and in practice often overfit specific datasets which limits their transferability (as is evidenced by the zero-shot results in Table~\ref{tab:stg_zero_shot}).
We therefore focus the discussion around our method's comparison to other generalist models.

\begin{table*}[!t]
\centering
\begin{subtable}[c]{0.24\textwidth}
\centering
\small
\setlength{\tabcolsep}{4pt}
\resizebox{\textwidth}{!}{\begin{tabular}{l c | c c c}

\toprule

\multirow{2}{*}{\bf Method} & \multirow{2}{1.4 cm}{\bf \centering Generalist Model} & \multicolumn{3}{c}{\bf Charades-CD-OOD} \\

& & R@0.3 & R@0.5 & mIoU  \\

\midrule

\demph{DepNet \cite{bao2021depnet}} & \demph{\ding{55}} & \demph{45.6} & \demph{27.6} & \demph{29.3} \\
\demph{DRN \cite{zeng2020drn}} & \demph{\ding{55}} & \demph{40.5} & \demph{30.4} & \demph{$-$} \\
\demph{STLG \cite{luo2022stlg}} & \demph{\ding{55}} & \demph{48.3} & \demph{30.4} & \demph{$-$} \\
\demph{SVPTR \cite{jiang2022svptr}} & \demph{\ding{55}} & \demph{50.3} & \demph{28.5} & \demph{32.1} \\
\demph{SiamGTR \cite{tan2024siamgtr}} & \demph{\ding{55}} & \demph{59.1} & \demph{35.5} & \demph{38.9} \\
\graycmidrule{1-5}
VTimeLLM$^{\dagger}$ \cite{huang2024vtimellm} & \ding{51} & 53.2 & 34.0 & 35.1 \\
TimeChat$^{\dagger}$ \cite{ren2024timechat} & \ding{51} & \underline{60.5} & \underline{36.1} & \underline{38.3} \\

\rowcolor{Light}
\model & \ding{51} & \bf 70.7 & \bf 47.3 & \bf 45.0 \\

\midrule
\bf \textcolor{blue}{$\Delta_{\text{Ours - TimeChat}}$} & $-$ & \textcolor{blue}{10.2} \textcolor{blue}{$\uparrow$} & \textcolor{blue}{11.2} \textcolor{blue}{$\uparrow$} & \textcolor{blue}{6.7} \textcolor{blue}{$\uparrow$} \\

\bottomrule
\end{tabular}}
\subcaption{Results on \textbf{Charades-CD-OOD}.}
\label{tab:vpg_charades_cd_ood}
\end{subtable}
\hspace{0.1em}
\begin{subtable}[c]{0.24\textwidth}
\centering
\small
\setlength{\tabcolsep}{4pt}
\resizebox{\textwidth}{!}{\begin{tabular}{l c | c c c}

\toprule

\multirow{2}{*}{\bf Method} & \multirow{2}{1.4 cm}{\bf \centering Generalist Model} & \multicolumn{3}{c}{\bf ANet-Captions} \\

& & R@0.3 & R@0.5 & mIoU  \\

\midrule

\demph{CBLN \cite{liu2021cbln}} & \demph{\ding{55}} & \demph{66.3} & \demph{48.1} & \demph{27.6} \\
\demph{2D-TAN \cite{zhang20202dtan}} & \demph{\ding{55}} & \demph{59.5} & \demph{44.5} & \demph{$-$}\\
\demph{3D-TPN \cite{zhang20203dtpn}} & \demph{\ding{55}} & \demph{67.6} & \demph{51.5} & \demph{$-$}\\
\demph{DepNet \cite{bao2021depnet}} & \demph{\ding{55}} & \demph{72.8} & \demph{55.9} & \demph{$-$}\\
\demph{SVPTR \cite{jiang2022svptr}} & \demph{\ding{55}} & \demph{\bf 78.1} & \demph{\bf 61.7} & \demph{\bf 55.9}\\

\graycmidrule{1-5}

VTimeLLM$^{\dagger}$ \cite{huang2024vtimellm} & \ding{51} & 66.1 & 50.3 & 45.6 \\
TimeChat$^{\dagger}$ \cite{ren2024timechat} & \ding{51} & \underline{67.9} & \underline{51.5} & \underline{47.0} \\

\rowcolor{Light}
\model & \ding{51} & \bf 74.1 & \bf 58.0 & \bf 53.7 \\

\midrule

\bf \textcolor{blue}{$\Delta_{\text{Ours - TimeChat}}$} & $-$ & \textcolor{blue}{6.2} \textcolor{blue}{$\uparrow$} & \textcolor{blue}{6.5} \textcolor{blue}{$\uparrow$} & \textcolor{blue}{6.7} \textcolor{blue}{$\uparrow$} \\

\bottomrule

\end{tabular}}
\subcaption{Results on \textbf{ActivityNet}.}
\label{tab:vpg_anet_captions}
\end{subtable}
\hspace{0.1em}
\begin{subtable}[c]{0.24\textwidth}
\centering
\small
\setlength{\tabcolsep}{4pt}
\resizebox{\textwidth}{!}{\begin{tabular}{l c | c c c}

\toprule

\multirow{2}{*}{\bf Method} & \multirow{2}{1.4 cm}{\bf \centering Generalist Model} & \multicolumn{3}{c}{\bf TACoS} \\

& & R@0.3 & R@0.5 & mIoU  \\

\midrule

\demph{CMIN \cite{zhang2019cmin}} & \demph{\ding{55}} & \demph{24.6} & \demph{18.1} & \demph{$-$} \\
\demph{2D-TAN \cite{zhang20202dtan}} & \demph{\ding{55}} & \demph{37.3} & \demph{25.3} & \demph{$-$}\\
\demph{3D-TPN \cite{zhang20203dtpn}} & \demph{\ding{55}} & \demph{40.3} & \demph{26.5} & \demph{$-$}\\
\demph{DepNet \cite{bao2021depnet}} & \demph{\ding{55}} & \demph{41.3} & \demph{27.2} & \demph{$-$}\\
\demph{SVPTR \cite{jiang2022svptr}} & \demph{\ding{55}} & \demph{\bf 47.9} & \demph{\bf 28.2} & \demph{\bf 31.4}\\

\graycmidrule{1-5}
VTimeLLM$^{\dagger}$ \cite{huang2024vtimellm} & \ding{51} & \underline{40.2} & \underline{25.6} & \underline{27.9} \\
TimeChat$^{\dagger}$ \cite{ren2024timechat} & \ding{51} & 39.5 & \underline{25.6} & 27.8 \\

\rowcolor{Light}
\model & \ding{51} & \bf 46.2 & \bf 27.8 & \bf 30.7 \\

\midrule

\bf \textcolor{blue}{$\Delta_{\text{Ours - TimeChat}}$} & $-$ & \textcolor{blue}{6.7} \textcolor{blue}{$\uparrow$} & \textcolor{blue}{2.2} \textcolor{blue}{$\uparrow$} & \textcolor{blue}{2.9} \textcolor{blue}{$\uparrow$} \\

\bottomrule

\end{tabular}}
\subcaption{Results on \textbf{TACoS}.}
\label{tab:vpg_tacos}
\end{subtable}
\hspace{0.1em}
\begin{subtable}[c]{0.24\textwidth}
\centering
\small
\setlength{\tabcolsep}{4pt}
\resizebox{\textwidth}{!}{\begin{tabular}{l c | c c c}

\toprule

\multirow{2}{*}{\bf Method} & \multirow{2}{1.4 cm}{\bf \centering Generalist Model} & \multicolumn{3}{c}{\bf YouCook2} \\

& & R@0.3 & R@0.5 & mIoU  \\

\midrule

\demph{DORi \cite{rodriguez2021dori}} & \demph{\ding{55}} & \demph{43.4} & \demph{30.5} & \demph{30.5} \\
\demph{DORi$^{\star}$ \cite{rodriguez2021dori}} & \demph{\ding{55}} & \demph{42.3} & \demph{29.9} & \demph{29.9} \\
\demph{LocFormer \cite{rodriguez2023locformer}} & \demph{\ding{55}} & \demph{46.8} & \demph{\bf 31.3} & \demph{30.9} \\
\demph{ExCL \cite{ghosh2019excl}} & \demph{\ding{55}} & \demph{26.6} & \demph{16.2} & \demph{18.9} \\
\demph{TMLGA \cite{rodriguez2020tmlga}} & \demph{\ding{55}} & \demph{34.8} & \demph{23.1} & \demph{24.4} \\

\graycmidrule{1-5}
VTimeLLM$^{\dagger}$ \cite{huang2024vtimellm} & \ding{51} & \underline{41.3} & 18.5 & 24.3 \\
TimeChat$^{\dagger}$ \cite{ren2024timechat} & \ding{51} & 40.9 & \underline{19.0} & \underline{26.6} \\

\rowcolor{Light}
\model & \ding{51} & \bf 48.1 & \bf 28.0 & \bf 31.5 \\

\midrule

\bf \textcolor{blue}{$\Delta_{\text{Ours - TimeChat}}$} & $-$ & \textcolor{blue}{7.2} \textcolor{blue}{$\uparrow$} & \textcolor{blue}{9.0} \textcolor{blue}{$\uparrow$} & \textcolor{blue}{4.9} \textcolor{blue}{$\uparrow$} \\

\bottomrule

\end{tabular}}
\subcaption{Results on \textbf{YouCook2}.}
\label{tab:vpg_youcook2}
\end{subtable}
\vspace{-3mm}
\caption{\textbf{Performance on VPG task on four different benchmarks: Charades-CD-OOD, ActivityNet-Captions, TACoS, YouCook2.} \model\ significantly improves over previous LLM-based models, and performs comparably to state-of-the-art specialist models. $^{\dagger}$We fine-tune VTimeLLM and TimeChat checkpoints for \EM{the VPG task}. Dori$^\star$ represents frozen text (BERT) encoder during fine-tuning.
}
\label{tab:main_vpg}
\vspace{-5mm}
\end{table*}

\begin{table}[!t]
\centering

\small
\setlength{\tabcolsep}{4pt}
\resizebox{\columnwidth}{!}{\begin{tabular}{l c | c c c c c c}

\toprule

\multirow{2}{*}{\bf Method} & \multirow{2}{1.75 cm}{\bf \centering Generalist Model} & \multicolumn{6}{c}{\bf NExT-GQA} \\

& & mIoP & IoP@0.3 & IoP@0.5 & mIoU & IoU@0.3 & IoU@0.5  \\

\midrule

\demph{VGT \cite{xiao2022vgt}} & \demph{\ding{55}} & \demph{24.7} & \demph{26.0} & \demph{24.6} & \demph{3.0} & \demph{4.2} & \demph{1.4} \\
\demph{VIOLETv2 \cite{fu2023violetv2}} & \demph{\ding{55}} & \demph{23.6} & \demph{25.1} & \demph{23.3} & \demph{3.1} & \demph{4.3} & \demph{1.3} \\
\demph{Temp[CLIP] NG+ \cite{xiao2024nextgqa}} & \demph{\ding{55}} & \demph{25.7} & \demph{31.4} & \demph{25.5} & \demph{12.1} & \demph{17.5} & \demph{8.9} \\
\demph{FrozenBiLM NG+ \cite{yang2022frozenbilm}} & \demph{\ding{55}} & \demph{24.2} & \demph{28.5} & \demph{23.7} & \demph{9.6} & \demph{13.5} & \demph{6.1} \\
\demph{SeViLA \cite{yu2024sevila}} & \demph{\ding{55}} & \demph{29.5} & \demph{34.7} & \demph{22.9} & \demph{21.7} & \demph{29.2} & \demph{13.8} \\

\graycmidrule{1-8}
LLoVi 7B$^{\diamond}$ \cite{zhang2023llovi} & \ding{51} & 20.7 & $-$ & 20.5 & 8.7 & $-$ & 6.0 \\
VideoStreaming$^{\diamond}$ \cite{qian2024videostreaming} & \ding{51} & 32.2 & $-$ & 31.0 & 19.3 & $-$ & 13.3 \\
LongRepo 7B$^{\diamond}$ \cite{kahatapitiya2024longrepo} & \ding{51} & 20.3 & $-$ & 20.0 & 8.7 & $-$ & 6.0\\
DeVi \cite{qin2024devi} & \ding{51} & \underline{33.8} & $-$ & \underline{32.2} & 20.7 & 17.4 & $-$ \\
HawkEye \cite{wang2024hawkeye} & \ding{51} & $-$ & $-$ & $-$ & \underline{25.7} & \underline{37.0} & \underline{19.5} \\
\rowcolor{Light}
\model & \ding{51} & \bf 34.7 & \bf 45.1 & \bf 33.5 & \bf 26.6 & \bf 39.5 & \bf 19.8 \\

\midrule

\bf \textcolor{blue}{$\Delta_{\text{Ours - SeViLA}}$} & $-$ & \textcolor{blue}{5.2} \textcolor{blue}{$\uparrow$} & \textcolor{blue}{10.4} \textcolor{blue}{$\uparrow$} & \textcolor{blue}{10.6} \textcolor{blue}{$\uparrow$} & \textcolor{blue}{4.9} \textcolor{blue}{$\uparrow$} & \textcolor{blue}{10.3} \textcolor{blue}{$\uparrow$} & \textcolor{blue}{6.0} \textcolor{blue}{$\uparrow$} \\

\bf \textcolor{blue}{$\Delta_{\text{Ours - HawkEye}}$} & $-$ & \textcolor{black}{$-$} & \textcolor{blue}{$-$} & \textcolor{black}{$-$} & \textcolor{blue}{0.9} \textcolor{blue}{$\uparrow$} & \textcolor{blue}{2.5} \textcolor{blue}{$\uparrow$} & \textcolor{blue}{0.3} \textcolor{blue}{$\uparrow$} \\

\bottomrule

\end{tabular}}
\vspace{-2mm}
\caption{\textbf{Performance on QG task on NeXT-GQA test split.} \model\ consistently achieves consistent improvements over the existing models across all metrics. $^{\diamond}$Results of LLoVi, LongRepo and VideoStreaming are from \cite{qian2024videostreaming}.} \label{tab:qg_nextgqa}
\vspace{-2mm}
\end{table}

\begin{table}[!t]
\centering

\small
\setlength{\tabcolsep}{4pt}
\resizebox{\columnwidth}{!}{\begin{tabular}{l | c c c c | c c c c}

\toprule

\multirow{3}{*}{\bf Method} & \multicolumn{4}{c|}{\bf Seen} & \multicolumn{4}{c}{\bf Unseen} \\
& \multicolumn{4}{c|}{$\uparrow$ mAP@IoU} & \multicolumn{4}{c}{$\uparrow$ mAP@IoU} \\
& @0.3 & @0.5 & @0.7 & @[0.3-0.7] & @0.3 & @0.5 & @0.7 & @[0.3-0.7] \\

\midrule

UMT \cite{liu2022umt} & 15.7 & 8.7 & 3.2 & 9.1 & 9.4 & 4.9 & 1.7 & 5.3 \\
MT+BCE \cite{mavroudi2023vina, afouras2024htstep} & \underline{46.2} & 29.9 & 12.9 & 29.8 & \underline{31.6} & 18.7 & 7.7 & 19.3 \\
ActionFormer-T \cite{zhang2022actionformer} & 41.2 & \underline{30.8} & \bf 18.3 & \underline{30.2} & 29.7 & \underline{20.3} & \underline{10.7} & \underline{20.4} \\

Timechat$^{\dagger}$ \cite{ren2024timechat} & 45.3 & 29.0 & 14.4 & 29.0 & 30.7 & 17.8 & 7.5 & 18.7 \\
\rowcolor{Light}
\model & \bf 48.9 & \bf 31.5 & \underline{18.0} & \bf 32.5 & \bf 33.0 & \bf 21.2 & \bf 11.1 & \bf 21.6 \\

\midrule

\bf \textcolor{blue}{$\Delta_{\text{Ours - TimeChat}}$} & \textcolor{blue}{3.6} \textcolor{blue}{$\uparrow$} & \textcolor{blue}{2.5} \textcolor{blue}{$\uparrow$} & \textcolor{blue}{3.6} \textcolor{blue}{$\uparrow$} & \textcolor{blue}{3.5} \textcolor{blue}{$\uparrow$} & \textcolor{blue}{2.3} \textcolor{blue}{$\uparrow$}& \textcolor{blue}{3.4} \textcolor{blue}{$\uparrow$}& \textcolor{blue}{3.6} \textcolor{blue}{$\uparrow$} & \textcolor{blue}{2.9} \textcolor{blue}{$\uparrow$} \\

\bottomrule

\end{tabular}}
\vspace{-2mm}
\caption{\textbf{Performance on AG task on HT-Step seen and unseen val split.} \model\ is the first LLM-based model to report results on 
\EM{for video grounding in the presence of negative, non-groundable queries}
and sets a new \EM{state of the art} in terms of average mAP score across various IoU thresholds. $^{\dagger}$We fine-tune the official TimeChat checkpoint for AG task.} \label{tab:gvpg_htstep}
\vspace{-6mm}
\end{table}

\vspace{-2mm}
\subsection{Implementation Details}
\label{sec:implementation_details}


We initialize the video encoder and LLM with the Video-LLaMA-7B \cite{zhang2023videollama} checkpoint, which is a similarly sized backbone used in existing LLM-based video grounding models \cite{huang2024vtimellm, ren2024timechat, wang2024hawkeye, qianmomentor}. Video-LLaMA is trained on video captioning tasks with WebVid \cite{bain2021frozen} and VideoChat \cite{li2023videochat}. The video encoder includes a ViT-G/14 from EVA-CLIP \cite{sun2023evaclip} as the frame feature extractor, followed by image and video QFormers. We initialize the decoder with random weights.
We keep the ViT frozen, apply LoRA \cite{hu2022lora} with rank 32 to the LLM and fully tune the Q-Formers, decoder, and linear layers.
We use LLaVA OneVision (OV) 72B \cite{li2024llavaonevision} as the external captioner for obtaining enriched queries in the training sets.
For the VPG task, we run four forward passes to perform MIL.
We pre-train our model for 40 epochs with a batch size 256, using AdamW \cite{loshchilov2018adamw} with a peak learning rate of 5e-5 and a cosine scheduler \cite{loshchilov2016sgdr} with a linear warmup for the first 20\% steps.
Pre-training takes 2 days on 16 V100 nodes (8 cards with 32G GPU memory each).
Additional details on pre-training, fine-tuning, and task-specific instructions are provided in the supplementary material.



\vspace{-2mm}
\subsection{Results}
\vspace{-2mm}



\vspace{1mm}

\noindent \textbf{Single-Query Temporal Grounding (\VTG).} We start by comparing \model \ against the state-of-the-art across three different \VTG\ benchmarks, namely Charades-STA, ActivityNet-Captions, and TACoS, in a zero-shot evaluation setting. We show the results in Table \ref{tab:stg_zero_shot}.  On Charades, \model\ achieves ZS scores of 59.5, 39.3, and 19.8 for R@0.3, R@0.5, and R@0.7, respectively, significantly outperforming all baseline models. Notably, \model\ surpasses Momenter~\cite{qianmomentor} and HawkEye~\cite{wang2024hawkeye} by $\textbf{11.4}$ and $\textbf{6.2}$ absolute mIoU points despite these models being pre-trained with 100x and 6x more segment-level data, respectively. A similar trend is observed on ActivityNet and TACoS where our model achieves improvements of 2.5 and 7.2 mIoU points over the nearest LLM-based models in zero-shot setting. Notably, the TACoS dataset contains short, under-specified queries, longer input videos, and fine-grained interval annotations, which pose challenges for LLMs with a fixed number of input frames. The enriched query descriptions that \model\ generates enable it to more accurately align them to video frames and precisely retrieve the correct intervals.

On the fine-tuned STG evaluation, as shown in Table \ref{tab:stg_fine_tune}, \model\ demonstrates an impressive gain of 5.7 and 11.7 points in R@0.3 over HawkEye on Charades and ActivityNet, respectively. Similarly on TACoS, \model\ surpasses all existing MLLMs by a considerable margin, e.g. 14.0 and 14.4 mIoU points over TimeChat and VTimeLLM. Furthermore, \model\ also beats many existing task-specific specialist models in all three benchmarks, significantly reducing the gap between specialist models and MLLMs for STG.



\begin{figure*}[!t]
\centering
\includegraphics[width=1.0\textwidth]{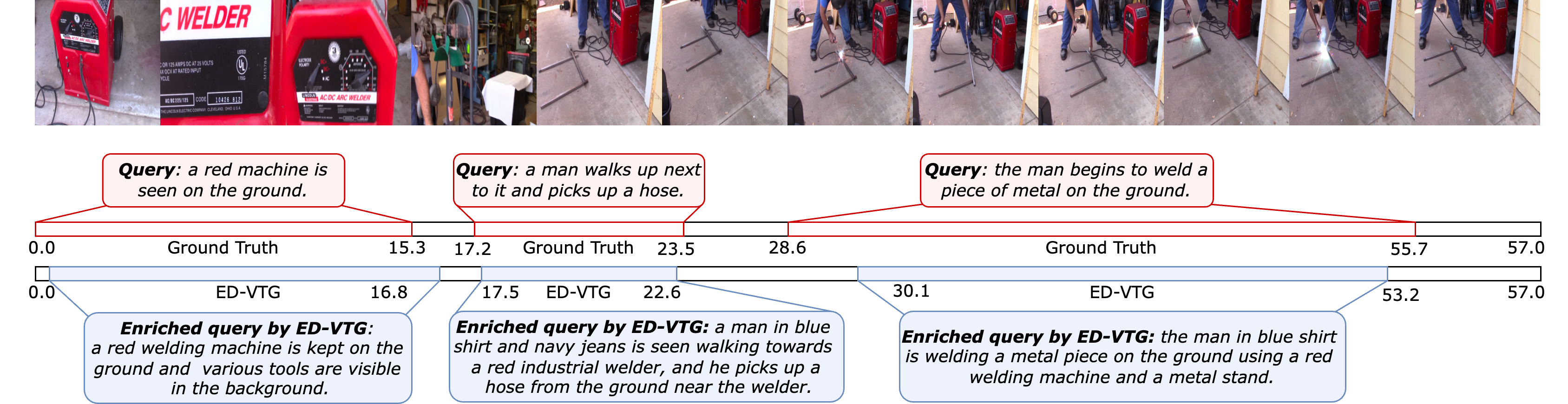}
\vspace{-1mm}
\caption{\textbf{Example of query enrichment and detection made by \model\ on video paragraph grounding (VPG) task from the ActivityNet-Captions \cite{krishna2017dense} dataset.} In this specific sample, three different queries are enriched and localized together.}
\vspace{-2mm}
\label{fig:visualization_vpg}
\end{figure*}

\begin{table*}[!t]
\centering

\small
\setlength{\tabcolsep}{4pt}
\resizebox{\textwidth}{!}{\begin{tabular}{l | c c c | c c c | c c c || c c c | c c c | c c c}

\toprule

\multirow{3}{1.5cm}{\bf Training Paradigm} & \multicolumn{9}{c||}{\bf Charades-STA \VTG} & \multicolumn{9}{c}{\bf ActivityNet-Captions \VTG} \\

& \multicolumn{3}{c}{ZS} & \multicolumn{3}{c}{FT w/o PT} & \multicolumn{3}{c||}{FT} & \multicolumn{3}{c}{ZS} & \multicolumn{3}{c}{FT w/o PT} & \multicolumn{3}{c}{FT} \\

& R@0.3 & R@0.5 & mIoU & R@0.3 & R@0.5 & mIoU & R@0.3 & R@0.5 & mIoU & R@0.3 & R@0.5 & mIoU & R@0.3 & R@0.5 & mIoU & R@0.3 & R@0.5 & mIoU \\

\midrule

Detect & 48.1 & 30.6 & 31.0 & 51.4 & 31.5 & 33.2 & 68.9 & 49.0 & 45.8 & 46.3 & 26.0 & 29.6 & 50.3 & 30.1 & 34.0 & 61.1 & 38.0 & 39.2  \\
Enrich \& Detect & 58.1 & 37.3 & 37.7 & 60.1 & 37.0 & 38.4 & 75.1 & 56.6 & 49.7 & 50.7 & 29.5 & 33.4 & 56.3 & 35.5 & 37.8 & 65.5 & 43.4 & 43.8 \\
\rowcolor{Light}
Enrich \& Detect w/ MIL & 59.5 & 39.3 & 40.2 & 62.8 & 38.4 & 40.3 & \bf 78.2 & \bf 62.1 & \bf 52.6 & 52.1 & 33.1 & 35.2 & 57.5 & 36.2 & 38.6 & \bf 67.6 & \bf 45.1 & \bf 44.9 \\

\bottomrule

\end{tabular}}
\vspace{-2mm}
\caption{\textbf{Ablation on the effect of enriched queries.} Our proposed enrich \& detect framework significantly gains over directly grounding the input queries across different evaluation settings. Introducing the MIL framework further improve the performance. FT w/o PT refers directly fine-tuning on respective datasets, without performing pre-training.}

\label{tab:ablation_direct_cot_mil}
\vspace{-5mm}
\end{table*}

\begin{table}[!t]
\centering

\small
\setlength{\tabcolsep}{4pt}
\resizebox{\columnwidth}{!}{\begin{tabular}{l | c c c | c c c}

\toprule

\multirow{2}{*}{\bf Training Paradigm} & \multicolumn{3}{c |}{\bf \centering Charades-STA \VTG} & \multicolumn{3}{c}{\bf \centering ANet-Captions \VTG} \\

& R@0.3 & R@0.5 & mIoU & R@0.3 & R@0.5 & mIoU  \\

\midrule

Detect & 51.4 & 31.5 & 33.2 & 50.3 & 30.1 & 34.0 \\
Offline Enrich + Detect & 51.7 & 31.5 & 33.4 & 49.8 & 29.9 & 33.7 \\
\rowcolor{Light}
Enrich \& Detect & \bf 60.1 & \bf 37.0 & \bf 38.4 & \bf 56.3 & \bf 35.5 & \bf 37.8  \\

\bottomrule

\end{tabular}}
\vspace{-3mm}
\caption{\textbf{Ablation on enrichment as a training pre-processing step.} The two step enrich \& detect framework is more helpful since the trained model learn to perform autonomous enrichment during evaluation. Reported results are in FT w/o PT setting.} \label{tab:ablation_offline_enrich}
\vspace{-3mm}
\end{table}

\vspace{1mm}

\begin{table}[!t]
\centering

\small
\setlength{\tabcolsep}{4pt}
\resizebox{\columnwidth}{!}{\begin{tabular}{c | c c c | c c c | c c c}

\toprule

\multirow{2}{*}{\bf Decoder} & \multicolumn{3}{c |}{\bf \centering Objectives} & \multicolumn{3}{c |}{\bf \centering Charades-STA \VTG} & \multicolumn{3}{c}{\bf \centering ANet-Captions \VTG} \\

& LM & L1 & gIoU & R@0.3 & R@0.5 & mIoU & R@0.3 & R@0.5 & mIoU  \\

\midrule

$-$ & \ding{51} & $-$ & $-$ & 54.2 & 33.2 & 34.1 & 51.0 & 31.6 & 35.5 \\

\ding{51} & \ding{51} & \ding{51} & $-$ & 58.5 & 36.0 & 37.0 & 55.8 & 34.4 & 37.1 \\
\ding{51} & \ding{51} & $-$ & \ding{51} & 58.9 & 36.2 & 37.1 & 55.8 & 34.7 & 37.3 \\
\rowcolor{Light}
\ding{51} & \ding{51} & \ding{51} & \ding{51} & \bf 60.1 & \bf 37.0 & \bf 38.4 & \bf 56.3 & \bf 35.5 & \bf 37.8 \\

\bottomrule

\end{tabular}}
\vspace{-2mm}
\caption{\textbf{Ablation study on different training objectives.} Training the model only using the LM loss (without the decoder) leads to significant performance drop. Results are in FT w/o PT setting.}
\label{tab:ablation_training_objectives}
\vspace{-6mm}
\end{table}

\vspace{1mm}
\noindent \textbf{Video Paragraph Grounding (VPG).} Next, we fine-tune \model\ for the VPG task, where the model processes multiple input queries in a temporal sequence. While specialist models have reported results in the past, no LLM-based models have previously addressed this challenging task. To establish a baseline for comparison, we fine-tune the officially released VTimeLLM and TimeChat pre-trained checkpoints. As shown in Table \ref{tab:vpg_charades_cd_ood}, \model\ significantly outperforms both LLMs on the Charades-CD-OOD dataset, achieving an absolute gain of 9.9 and 6.7 mIoU points over VTimeLLM and TimeChat, respectively. Additionally, \model\ surpasses all specialist models on this dataset by a substantial margin, setting a new state-of-the-art. 

We also evaluate VPG performance on three other benchmarks: ActivityNet-Captions, TACoS, and YouCook2, as presented in Tables \ref{tab:vpg_anet_captions}, \ref{tab:vpg_tacos}, and \ref{tab:vpg_youcook2}. Consistent with the results on Charades, \model\ outperforms LLM-based models across all metrics on these datasets, with mIoU gains of 6.7, 2.9, and 4.9 over TimeChat on ActivityNet-Captions, TACoS, and YouCook2, respectively. \model\ also exceeds many existing specialist VPG models, demonstrating the effectiveness of our enriched queries and cascaded interval decoder.


\vspace{1mm}
\noindent \textbf{Question Grounding (QG).} To further assess the model's generalization capabilities, we conduct zero-shot evaluation on the held-out QG task.
Our results on the NeXT-GQA test set are presented in Table \ref{tab:qg_nextgqa}. Notably, \model\ achieves state-of-the-art performance across all metrics, outperforming both specialist and LLM-based models by a significant margin. Specifically, \model\ surpasses the existing best baseline, HawkEye, by 2.5 points in terms of IoU@0.3 score, demonstrating its strong generalizability.

\vspace{1mm}
\noindent \textbf{Article Grounding (AG).} We assess \model\ on the AG on the HT-Step benchmark.
Table \ref{tab:gvpg_htstep} shows the fine-tuned performance on the AG task, where \model\ shows significant improvements over existing baselines. Notably, on the challenging \textit{unseen} split, our model achieves the best results across all metrics, surpassing both the LLM and specialist models by a decent margin. The ability
\EM{to handle non-groundable queries} underscores \model's real-world applicability, as it does not always assume that the query is occurring in the input video.

\vspace{-2mm}
\subsection{Ablation Study}

\noindent \textbf{Effect of Enriched Queries.} We examine the impact of query enrichment through a step-by-step ablation, as shown in Tables \ref{tab:ablation_direct_cot_mil} and \ref{tab:ablation_offline_enrich}. Initially, we compare the effect of enrichment without MIL paradigm against direct grounding. As indicated in the first two rows of Table \ref{tab:ablation_direct_cot_mil}, enriched queries lead to significant improvements on the Charades-STA and ActivityNet-Captions \VTG\ benchmarks. In the zero-shot (ZS) setting, enrichment results in gains of 6.7 and 3.8 mIoU points on these datasets, respectively. Introducing the MIL paradigm further enhances performance, adding 2.5 and 1.8 mIoU points. Similar improvements are observed in other evaluation settings, highlighting the substantial effectiveness of query enrichment within the MIL framework.

Additionally, we investigate the effect of offline enrichment in Table \ref{tab:ablation_offline_enrich}. In this scenario, instead of using a two-step grounding process, we enrich the queries as a training enrichment step, and the model is then directly provided with these enriched queries as input and asked to directly perform grounding. However, we find that offline enrichment is not advantageous, primarily due to the lack of enrichment during evaluation. In contrast, our two-step grounding approach allows the trained model to learn how to enrich queries and improve them autonomously when necessary, resulting in significant performance gains.


\noindent \textbf{Training Objectives.} We ablate different training objectives on the Charades and ActivityNet \VTG\ benchmarks, as shown in Table \ref{tab:ablation_training_objectives}.
The decoder achieves optimal performance when both L1 and gIoU objectives are used together; omitting either one slightly reduces the scores.

\vspace{-1mm}
\subsection{Qualitative Results and Error Analysis}
\vspace{-1mm}


Figure \ref{fig:visualization_vpg} visualizes a VPG sample from the ActivityNet-Captions dataset where \model\ meaningfully enriches the three input queries, then proceeds to precisely ground them. Please refer to supplementary for more qualative results,
including intuitive demonstrations of the flexibility in choosing between enrichment or direct detection, and comparison with TimeChat basline.
We also present there some interesting failure cases, such as the ones involving small and obscured objects in long input videos.
\vspace{-2mm}

\section{Conclusion}
\vspace{-2mm}
In this paper, we presented \model, a novel method for fine-grained video temporal grounding using multi-modal LLMs. By enhancing queries with additional details, utilizing a lightweight decoder and trained in a multiple-instance framework, \model~accurately locates temporal boundaries in videos. Our experiments show that our method outperforms existing LLM-based works and is competitive with specialized models, especially in zero-shot settings, setting a new, strong benchmark for video grounding tasks.

\section{Acknowledgement}
\vspace{-2mm}
This codebase for this project is built on the TimeChat \cite{ren2024timechat} and VTimeLLM \cite{huang2024vtimellm} repository. We would like to thank the respective authors for their contributions. We gratefully acknowledge the following colleagues at FAIR for valuable discussions and support of our project: Tushar Nagarajan, Huiyu Wang, Yujie Lu, and Arjun Somayazulu.

{
\small
\bibliographystyle{ieeenat_fullname}
\bibliography{main}

\begin{thebibliography}{135}
\providecommand{\natexlab}[1]{#1}
\providecommand{\url}[1]{\texttt{#1}}
\expandafter\ifx\csname urlstyle\endcsname\relax
  \providecommand{\doi}[1]{doi: #1}\else
  \providecommand{\doi}{doi: \begingroup \urlstyle{rm}\Url}\fi

\bibitem[Abdar et~al.(2023)Abdar, Kollati, Kuraparthi, Pourpanah, McDuff, Ghavamzadeh, Yan, Mohamed, Khosravi, Cambria, et~al.]{abdar2023review}
Moloud Abdar, Meenakshi Kollati, Swaraja Kuraparthi, Farhad Pourpanah, Daniel McDuff, Mohammad Ghavamzadeh, Shuicheng Yan, Abduallah Mohamed, Abbas Khosravi, Erik Cambria, et~al.
\newblock A review of deep learning for video captioning.
\newblock \emph{arXiv preprint arXiv:2304.11431}, 2023.

\bibitem[Achiam et~al.(2023)Achiam, Adler, Agarwal, Ahmad, Akkaya, Aleman, Almeida, Altenschmidt, Altman, Anadkat, et~al.]{achiam2023gpt4}
Josh Achiam, Steven Adler, Sandhini Agarwal, Lama Ahmad, Ilge Akkaya, Florencia~Leoni Aleman, Diogo Almeida, Janko Altenschmidt, Sam Altman, Shyamal Anadkat, et~al.
\newblock Gpt-4 technical report.
\newblock \emph{arXiv preprint arXiv:2303.08774}, 2023.

\bibitem[Afouras et~al.(2024)Afouras, Mavroudi, Nagarajan, Wang, and Torresani]{afouras2024htstep}
Triantafyllos Afouras, Effrosyni Mavroudi, Tushar Nagarajan, Huiyu Wang, and Lorenzo Torresani.
\newblock Ht-step: Aligning instructional articles with how-to videos.
\newblock In \emph{NeurIPS}, 2024.

\bibitem[Andrews et~al.(2003)Andrews, Tsochantaridis, and Hofmann]{andrews2003support}
Stuart Andrews, Ioannis Tsochantaridis, and Thomas Hofmann.
\newblock Support vector machines for multiple-instance learning.
\newblock In \emph{NeurIPS}, pages 561--568, 2003.

\bibitem[Anne~Hendricks et~al.(2017)Anne~Hendricks, Wang, Shechtman, Sivic, Darrell, and Russell]{anne2017didemo}
Lisa Anne~Hendricks, Oliver Wang, Eli Shechtman, Josef Sivic, Trevor Darrell, and Bryan Russell.
\newblock Localizing moments in video with natural language.
\newblock In \emph{ICCV}, pages 5803--5812, 2017.

\bibitem[Bain et~al.(2021)Bain, Nagrani, Varol, and Zisserman]{bain2021frozen}
Max Bain, Arsha Nagrani, G{\"u}l Varol, and Andrew Zisserman.
\newblock Frozen in time: A joint video and image encoder for end-to-end retrieval.
\newblock In \emph{ICCV}, pages 1728--1738, 2021.

\bibitem[Bao et~al.(2021)Bao, Zheng, and Mu]{bao2021depnet}
Peijun Bao, Qian Zheng, and Yadong Mu.
\newblock Dense events grounding in video.
\newblock In \emph{AAAI}, pages 920--928, 2021.

\bibitem[Bilen and Vedaldi(2016)]{bilen2016weakly}
Hakan Bilen and Andrea Vedaldi.
\newblock Weakly supervised deep detection networks.
\newblock In \emph{CVPR}, pages 2846--2854, 2016.

\bibitem[Cao et~al.(2021)Cao, Chen, Shou, Zhang, and Zou]{cao2021gtr}
Meng Cao, Long Chen, Mike~Zheng Shou, Can Zhang, and Yuexian Zou.
\newblock On pursuit of designing multi-modal transformer for video grounding.
\newblock In \emph{EMNLP}, pages 9810--9823, 2021.

\bibitem[Cao et~al.(2024)Cao, Zhang, Du, Yu, Li, and Wang]{cao2024flashvtg}
Zhuo Cao, Bingqing Zhang, Heming Du, Xin Yu, Xue Li, and Sen Wang.
\newblock Flashvtg: Feature layering and adaptive score handling network for video temporal grounding.
\newblock In \emph{WACV}, 2024.

\bibitem[Chen et~al.(2021)Chen, Tsai, and Yang]{chen2021end}
Yi-Wen Chen, Yi-Hsuan Tsai, and Ming-Hsuan Yang.
\newblock End-to-end multi-modal video temporal grounding.
\newblock \emph{NeurIPS}, 34:\penalty0 28442--28453, 2021.

\bibitem[Cinbis et~al.(2015)Cinbis, Verbeek, and Schmid]{cinbis2017weakly}
Ramazan Cinbis, Jakob Verbeek, and Cordelia Schmid.
\newblock Weakly supervised object localization with multi-fold multiple instance learning.
\newblock \emph{IEEE TPAMI}, 39, 2015.

\bibitem[Deng et~al.(2021)Deng, Chen, Chen, He, and Wu]{Deng_2021_CVPR}
Chaorui Deng, Shizhe Chen, Da Chen, Yuan He, and Qi Wu.
\newblock Sketch, ground, and refine: Top-down dense video captioning.
\newblock In \emph{CVPR}, pages 234--243, 2021.

\bibitem[Dietterich et~al.(1997)Dietterich, Lathrop, and Lozano-P{\'e}rez]{dietterich199731}
Thomas~G. Dietterich, Richard~H. Lathrop, and Tom{\'a}s Lozano-P{\'e}rez.
\newblock Solving the multiple instance problem with axis-parallel rectangles.
\newblock \emph{Artificial Intelligence}, 89\penalty0 (1):\penalty0 31--71, 1997.

\bibitem[Dong et~al.(2017)Dong, Mallinson, Reddy, and Lapata]{Dong2017LearningTP}
Li Dong, Jonathan Mallinson, Siva Reddy, and Mirella Lapata.
\newblock Learning to paraphrase for question answering.
\newblock In \emph{EMNLP}, 2017.

\bibitem[Dubey et~al.(2024)Dubey, Jauhri, Pandey, Kadian, Al-Dahle, Letman, Mathur, Schelten, Yang, Fan, et~al.]{dubey2024llama3}
Abhimanyu Dubey, Abhinav Jauhri, Abhinav Pandey, Abhishek Kadian, Ahmad Al-Dahle, Aiesha Letman, Akhil Mathur, Alan Schelten, Amy Yang, Angela Fan, et~al.
\newblock The llama 3 herd of models.
\newblock \emph{arXiv preprint arXiv:2407.21783}, 2024.

\bibitem[Fu et~al.(2023)Fu, Li, Gan, Lin, Wang, Wang, and Liu]{fu2023violetv2}
Tsu-Jui Fu, Linjie Li, Zhe Gan, Kevin Lin, William~Yang Wang, Lijuan Wang, and Zicheng Liu.
\newblock An empirical study of end-to-end video-language transformers with masked visual modeling.
\newblock In \emph{CVPR}, pages 22898--22909, 2023.

\bibitem[Gao et~al.(2017{\natexlab{a}})Gao, Sun, Yang, and Nevatia]{gao2017charadessta}
Jiyang Gao, Chen Sun, Zhenheng Yang, and Ram Nevatia.
\newblock Tall: Temporal activity localization via language query.
\newblock In \emph{ICCV}, pages 5267--5275, 2017{\natexlab{a}}.

\bibitem[Gao et~al.(2017{\natexlab{b}})Gao, Sun, Yang, and Nevatia]{gao2017ctrl}
Jiyang Gao, Chen Sun, Zhenheng Yang, and Ram Nevatia.
\newblock Tall: Temporal activity localization via language query.
\newblock In \emph{ICCV}, pages 5267--5275, 2017{\natexlab{b}}.

\bibitem[Gao et~al.(2020)Gao, Zhang, Ou, and Yu]{gao-etal-2020-paraphrase}
Silin Gao, Yichi Zhang, Zhijian Ou, and Zhou Yu.
\newblock Paraphrase augmented task-oriented dialog generation.
\newblock In \emph{ACL}, pages 639--649, 2020.

\bibitem[Ghosh et~al.(2019)Ghosh, Agarwal, Parekh, and Hauptmann]{ghosh2019excl}
Soham Ghosh, Anuva Agarwal, Zarana Parekh, and Alexander~G Hauptmann.
\newblock Excl: Extractive clip localization using natural language descriptions.
\newblock In \emph{NAACL}, pages 1984--1990, 2019.

\bibitem[Girshick et~al.(2015)Girshick, Donahue, Darrell, and Malik]{girshick2015rcnn}
Ross Girshick, Jeff Donahue, Trevor Darrell, and Jitendra Malik.
\newblock Region-based convolutional networks for accurate object detection and segmentation.
\newblock \emph{IEEE TPAMI}, 38\penalty0 (1):\penalty0 142--158, 2015.

\bibitem[Gordeev et~al.(2024)Gordeev, Dokholyan, Tolstykh, and Kuprashevich]{gordeev2024sgdetr}
Aleksandr Gordeev, Vladimir Dokholyan, Irina Tolstykh, and Maksim Kuprashevich.
\newblock Saliency-guided detr for moment retrieval and highlight detection.
\newblock \emph{arXiv preprint arXiv:2410.01615}, 2024.

\bibitem[He et~al.(2017)He, Gkioxari, Doll{\'a}r, and Girshick]{he2017maskrcnn}
Kaiming He, Georgia Gkioxari, Piotr Doll{\'a}r, and Ross Girshick.
\newblock Mask r-cnn.
\newblock In \emph{ICCV}, pages 2961--2969, 2017.

\bibitem[Hu et~al.(2022)Hu, Shen, Wallis, Allen-Zhu, Li, Wang, Wang, and Chen]{hu2022lora}
Edward~J Hu, Yelong Shen, Phillip Wallis, Zeyuan Allen-Zhu, Yuanzhi Li, Shean Wang, Lu Wang, and Weizhu Chen.
\newblock Lo{RA}: Low-rank adaptation of large language models.
\newblock In \emph{ICLR}, 2022.

\bibitem[Huang et~al.(2024)Huang, Wang, Chen, Song, and Zhu]{huang2024vtimellm}
Bin Huang, Xin Wang, Hong Chen, Zihan Song, and Wenwu Zhu.
\newblock Vtimellm: Empower llm to grasp video moments.
\newblock In \emph{CVPR}, pages 14271--14280, 2024.

\bibitem[Huang et~al.(2025)Huang, Liao, Radhakrishnan, Yin, Molchanov, Yu, and Kautz]{huang2025lita}
De-An Huang, Shijia Liao, Subhashree Radhakrishnan, Hongxu Yin, Pavlo Molchanov, Zhiding Yu, and Jan Kautz.
\newblock Lita: Language instructed temporal-localization assistant.
\newblock In \emph{ECCV}, pages 202--218. Springer, 2025.

\bibitem[Huang et~al.(2020)Huang, Pang, Zhu, Rivera, and Soricut]{huang2020vitt}
Gabriel Huang, Bo Pang, Zhenhai Zhu, Clara Rivera, and Radu Soricut.
\newblock Multimodal pretraining for dense video captioning.
\newblock In \emph{AACL}, pages 470--490, 2020.

\bibitem[Huang et~al.(2022)Huang, Jin, Gong, and Liu]{huang2022emb}
Jiabo Huang, Hailin Jin, Shaogang Gong, and Yang Liu.
\newblock Video activity localisation with uncertainties in temporal boundary.
\newblock In \emph{ECCV}, pages 724--740. Springer, 2022.

\bibitem[Iashin and Rahtu(2020)]{MDVC_Iashin_2020}
Vladimir Iashin and Esa Rahtu.
\newblock Multi-modal dense video captioning.
\newblock In \emph{CVPR Workshops}, 2020.

\bibitem[Jang et~al.(2023)Jang, Park, Kim, Kwon, and Sohn]{jang2023eatr}
Jinhyun Jang, Jungin Park, Jin Kim, Hyeongjun Kwon, and Kwanghoon Sohn.
\newblock Knowing where to focus: Event-aware transformer for video grounding.
\newblock In \emph{CVPR}, pages 13846--13856, 2023.

\bibitem[Jiang et~al.(2022)Jiang, Xu, Zhang, Shen, Cao, and Shen]{jiang2022svptr}
Xun Jiang, Xing Xu, Jingran Zhang, Fumin Shen, Zuo Cao, and Heng~Tao Shen.
\newblock Semi-supervised video paragraph grounding with contrastive encoder.
\newblock In \emph{CVPR}, pages 2466--2475, 2022.

\bibitem[Kahatapitiya et~al.(2024)Kahatapitiya, Ranasinghe, Park, and Ryoo]{kahatapitiya2024longrepo}
Kumara Kahatapitiya, Kanchana Ranasinghe, Jongwoo Park, and Michael~S Ryoo.
\newblock Language repository for long video understanding.
\newblock \emph{arXiv preprint arXiv:2403.14622}, 2024.

\bibitem[Kim et~al.(2023)Kim, Park, Lee, Park, and Sohn]{kim2023ltzvg}
Dahye Kim, Jungin Park, Jiyoung Lee, Seongheon Park, and Kwanghoon Sohn.
\newblock Language-free training for zero-shot video grounding.
\newblock In \emph{WACV}, pages 2539--2548, 2023.

\bibitem[Koupaee and Wang(2018)]{koupaee2018wikihow}
Mahnaz Koupaee and William~Yang Wang.
\newblock Wikihow: A large scale text summarization dataset.
\newblock \emph{arXiv preprint arXiv:1810.09305}, 2018.

\bibitem[Krishna et~al.(2017)Krishna, Hata, Ren, Fei-Fei, and Carlos~Niebles]{krishna2017dense}
Ranjay Krishna, Kenji Hata, Frederic Ren, Li Fei-Fei, and Juan Carlos~Niebles.
\newblock Dense-captioning events in videos.
\newblock In \emph{ICCV}, pages 706--715, 2017.

\bibitem[Lee and Byun(2024)]{lee2024bamdetr}
Pilhyeon Lee and Hyeran Byun.
\newblock Bam-detr: Boundary-aligned moment detection transformer for temporal sentence grounding in videos.
\newblock In \emph{ECCV}, pages 220--238. Springer, 2024.

\bibitem[Lei et~al.(2021)Lei, Berg, and Bansal]{lei2021momentdetr}
Jie Lei, Tamara~L Berg, and Mohit Bansal.
\newblock Detecting moments and highlights in videos via natural language queries.
\newblock \emph{NeurIPS}, 34:\penalty0 11846--11858, 2021.

\bibitem[Li et~al.(2025)Li, Zhang, Guo, Zhang, Li, Zhang, Zhang, Li, Liu, and Li]{li2024llavaonevision}
Bo Li, Yuanhan Zhang, Dong Guo, Renrui Zhang, Feng Li, Hao Zhang, Kaichen Zhang, Yanwei Li, Ziwei Liu, and Chunyuan Li.
\newblock Llava-onevision: Easy visual task transfer.
\newblock \emph{TMLR}, 2025.

\bibitem[Li et~al.(2023{\natexlab{a}})Li, Li, Savarese, and Hoi]{li2023blip2}
Junnan Li, Dongxu Li, Silvio Savarese, and Steven Hoi.
\newblock Blip-2: Bootstrapping language-image pre-training with frozen image encoders and large language models.
\newblock In \emph{ICML}, pages 19730--19742. PMLR, 2023{\natexlab{a}}.

\bibitem[Li et~al.(2023{\natexlab{b}})Li, He, Wang, Li, Wang, Luo, Wang, Wang, and Qiao]{li2023videochat}
KunChang Li, Yinan He, Yi Wang, Yizhuo Li, Wenhai Wang, Ping Luo, Yali Wang, Limin Wang, and Yu Qiao.
\newblock Videochat: Chat-centric video understanding.
\newblock \emph{arXiv preprint arXiv:2305.06355}, 2023{\natexlab{b}}.

\bibitem[Li et~al.(2024{\natexlab{a}})Li, Wang, He, Li, Wang, Liu, Wang, Xu, Chen, Luo, et~al.]{li2024videochat2}
Kunchang Li, Yali Wang, Yinan He, Yizhuo Li, Yi Wang, Yi Liu, Zun Wang, Jilan Xu, Guo Chen, Ping Luo, et~al.
\newblock Mvbench: A comprehensive multi-modal video understanding benchmark.
\newblock In \emph{CVPR}, pages 22195--22206, 2024{\natexlab{a}}.

\bibitem[Li et~al.(2023{\natexlab{c}})Li, Yin, Li, Chen, Wang, Ren, Li, Yang, Xu, Sun, et~al.]{li2023m3it}
Lei Li, Yuwei Yin, Shicheng Li, Liang Chen, Peiyi Wang, Shuhuai Ren, Mukai Li, Yazheng Yang, Jingjing Xu, Xu Sun, et~al.
\newblock {M$^3$IT: A Large-Scale Dataset towards Multi-Modal Multilingual Instruction Tuning}.
\newblock \emph{arXiv preprint arXiv:2306.04387}, 2023{\natexlab{c}}.

\bibitem[Li et~al.(2024{\natexlab{b}})Li, Xie, Xie, Zhao, Zhang, Zheng, Zhao, and Zhang]{li2024momentdiff}
Pandeng Li, Chen-Wei Xie, Hongtao Xie, Liming Zhao, Lei Zhang, Yun Zheng, Deli Zhao, and Yongdong Zhang.
\newblock Momentdiff: Generative video moment retrieval from random to real.
\newblock \emph{NeurIPS}, 36, 2024{\natexlab{b}}.

\bibitem[Li et~al.(2018)Li, Yao, Pan, Chao, and Mei]{Li2018JointlyLA}
Yehao Li, Ting Yao, Yingwei Pan, Hongyang Chao, and Tao Mei.
\newblock Jointly localizing and describing events for dense video captioning.
\newblock \emph{CVPR}, pages 7492--7500, 2018.

\bibitem[Li et~al.(2024{\natexlab{c}})Li, Xu, Zhang, Song, Cai, Qi, Zhou, Pan, Li, Tu, et~al.]{li2024groundinggpt}
Zhaowei Li, Qi Xu, Dong Zhang, Hang Song, Yiqing Cai, Qi Qi, Ran Zhou, Junting Pan, Zefeng Li, Vu Tu, et~al.
\newblock Groundinggpt: Language enhanced multi-modal grounding model.
\newblock In \emph{ACL}, pages 6657--6678, 2024{\natexlab{c}}.

\bibitem[Lin et~al.(2023)Lin, Zhang, Chen, Pramanick, Gao, Wang, Yan, and Shou]{lin2023univtg}
Kevin~Qinghong Lin, Pengchuan Zhang, Joya Chen, Shraman Pramanick, Difei Gao, Alex~Jinpeng Wang, Rui Yan, and Mike~Zheng Shou.
\newblock Univtg: Towards unified video-language temporal grounding.
\newblock In \emph{ICCV}, pages 2794--2804, 2023.

\bibitem[Liu et~al.(2021)Liu, Qu, Dong, Zhou, Cheng, Wei, Xu, and Xie]{liu2021cbln}
Daizong Liu, Xiaoye Qu, Jianfeng Dong, Pan Zhou, Yu Cheng, Wei Wei, Zichuan Xu, and Yulai Xie.
\newblock Context-aware biaffine localizing network for temporal sentence grounding.
\newblock In \emph{CVPR}, pages 11235--11244, 2021.

\bibitem[Liu et~al.(2022{\natexlab{a}})Liu, Qu, Di, Cheng, Xu, and Zhou]{liu2022mgslnet}
Daizong Liu, Xiaoye Qu, Xing Di, Yu Cheng, Zichuan Xu, and Pan Zhou.
\newblock Memory-guided semantic learning network for temporal sentence grounding.
\newblock In \emph{AAAI}, pages 1665--1673, 2022{\natexlab{a}}.

\bibitem[Liu et~al.(2016)Liu, Anguelov, Erhan, Szegedy, Reed, Fu, and Berg]{liu2016ssd}
Wei Liu, Dragomir Anguelov, Dumitru Erhan, Christian Szegedy, Scott Reed, Cheng-Yang Fu, and Alexander~C Berg.
\newblock Ssd: Single shot multibox detector.
\newblock In \emph{ECCV}, pages 21--37, 2016.

\bibitem[Liu et~al.(2022{\natexlab{b}})Liu, Li, Wu, Chen, Shan, and Qie]{liu2022umt}
Ye Liu, Siyuan Li, Yang Wu, Chang-Wen Chen, Ying Shan, and Xiaohu Qie.
\newblock Umt: Unified multi-modal transformers for joint video moment retrieval and highlight detection.
\newblock In \emph{CVPR}, pages 3042--3051, 2022{\natexlab{b}}.

\bibitem[Loshchilov and Hutter(2017)]{loshchilov2016sgdr}
Ilya Loshchilov and Frank Hutter.
\newblock Sgdr: Stochastic gradient descent with warm restarts.
\newblock In \emph{ICLR}, 2017.

\bibitem[Loshchilov and Hutter(2019)]{loshchilov2018adamw}
Ilya Loshchilov and Frank Hutter.
\newblock Decoupled weight decay regularization.
\newblock In \emph{ICLR}, 2019.

\bibitem[Louradour(2023)]{whisper}
Jérôme Louradour.
\newblock whisper-timestamped., 2023.

\bibitem[Luo et~al.(2022)Luo, Chen, Chen, Wu, and Jiang]{luo2022stlg}
Fan Luo, Shaoxiang Chen, Jingjing Chen, Zuxuan Wu, and Yu-Gang Jiang.
\newblock Self-supervised learning for semi-supervised temporal language grounding.
\newblock \emph{IEEE Transactions on Multimedia}, 25:\penalty0 7747--7757, 2022.

\bibitem[Luo et~al.(2023)Luo, Zhao, Yang, Dong, Li, Lu, Wang, Hu, Qiu, and Wei]{luo2023valley}
Ruipu Luo, Ziwang Zhao, Min Yang, Junwei Dong, Da Li, Pengcheng Lu, Tao Wang, Linmei Hu, Minghui Qiu, and Zhongyu Wei.
\newblock Valley: Video assistant with large language model enhanced ability.
\newblock \emph{arXiv preprint arXiv:2306.07207}, 2023.

\bibitem[Ma et~al.(2023)Ma, Zang, Feng, Fang, Ban, Wei, He, Li, and Sun]{ma2023llavilo}
Kaijing Ma, Xianghao Zang, Zerun Feng, Han Fang, Chao Ban, Yuhan Wei, Zhongjiang He, Yongxiang Li, and Hao Sun.
\newblock Llavilo: Boosting video moment retrieval via adapter-based multimodal modeling.
\newblock In \emph{ICCV Workshops}, pages 2790--2795. IEEE, 2023.

\bibitem[Maaz et~al.(2024)Maaz, Rasheed, Khan, and Khan]{Maaz2023videochatgpt}
Muhammad Maaz, Hanoona Rasheed, Salman Khan, and Fahad~Shahbaz Khan.
\newblock Video-chatgpt: Towards detailed video understanding via large vision and language models.
\newblock In \emph{ACL}, 2024.

\bibitem[Mavroudi et~al.(2023)Mavroudi, Afouras, and Torresani]{mavroudi2023vina}
Effrosyni Mavroudi, Triantafyllos Afouras, and Lorenzo Torresani.
\newblock Learning to ground instructional articles in videos through narrations.
\newblock In \emph{ICCV}, pages 15201--15213, 2023.

\bibitem[Menon and Vondrick(2023)]{MenonICLR23}
Sachit Menon and Carl Vondrick.
\newblock Visual classification via description from large language models.
\newblock \emph{ICLR}, 2023.

\bibitem[Miech et~al.(2019)Miech, Zhukov, Alayrac, Tapaswi, Laptev, and Sivic]{miech2019howto100m}
Antoine Miech, Dimitri Zhukov, Jean-Baptiste Alayrac, Makarand Tapaswi, Ivan Laptev, and Josef Sivic.
\newblock Howto100m: Learning a text-video embedding by watching hundred million narrated video clips.
\newblock In \emph{ICCV}, pages 2630--2640, 2019.

\bibitem[Moon et~al.(2023{\natexlab{a}})Moon, Hyun, Lee, and Heo]{moon2023cgdetr}
WonJun Moon, Sangeek Hyun, SuBeen Lee, and Jae-Pil Heo.
\newblock Correlation-guided query-dependency calibration for video temporal grounding.
\newblock \emph{arXiv preprint arXiv:2311.08835}, 2023{\natexlab{a}}.

\bibitem[Moon et~al.(2023{\natexlab{b}})Moon, Hyun, Park, Park, and Heo]{moon2023qddetr}
WonJun Moon, Sangeek Hyun, SangUk Park, Dongchan Park, and Jae-Pil Heo.
\newblock Query-dependent video representation for moment retrieval and highlight detection.
\newblock In \emph{CVPR}, pages 23023--23033, 2023{\natexlab{b}}.

\bibitem[Mun et~al.(2020)Mun, Cho, and Han]{mun2020lgi}
Jonghwan Mun, Minsu Cho, and Bohyung Han.
\newblock Local-global video-text interactions for temporal grounding.
\newblock In \emph{CVPR}, pages 10810--10819, 2020.

\bibitem[Nagrani et~al.(2022)Nagrani, Seo, Seybold, Hauth, Manen, Sun, and Schmid]{nagrani2022videocc}
Arsha Nagrani, Paul~Hongsuck Seo, Bryan Seybold, Anja Hauth, Santiago Manen, Chen Sun, and Cordelia Schmid.
\newblock Learning audio-video modalities from image captions.
\newblock In \emph{ECCV}, pages 407--426. Springer, 2022.

\bibitem[Nam et~al.(2021)Nam, Ahn, Kang, Ha, and Choi]{nam2021psvl}
Jinwoo Nam, Daechul Ahn, Dongyeop Kang, Seong~Jong Ha, and Jonghyun Choi.
\newblock Zero-shot natural language video localization.
\newblock In \emph{ICCV}, pages 1470--1479, 2021.

\bibitem[Nguyen et~al.(2019)Nguyen, Liu, Prasad, Bui, Pham, and Venkatesh]{nguyen2019weakly}
Phuc~Xuan Nguyen, Tianzhu Liu, Gaurav Prasad, Hung~Hai Bui, Binh Pham, and Svetha Venkatesh.
\newblock Weakly supervised action localization by sparse temporal pooling network.
\newblock In \emph{CVPR}, pages 6752--6761, 2019.

\bibitem[Oncescu et~al.(2021)Oncescu, Henriques, Liu, Zisserman, and Albanie]{oncescu2021queryd}
Andreea-Maria Oncescu, Joao~F Henriques, Yang Liu, Andrew Zisserman, and Samuel Albanie.
\newblock Queryd: A video dataset with high-quality text and audio narrations.
\newblock In \emph{ICASSP}, pages 2265--2269. IEEE, 2021.

\bibitem[Pan et~al.(2020)Pan, Cai, Huang, Lee, Gaidon, Adeli, and Niebles]{pan2020spatio}
Boxiao Pan, Haoye Cai, De-An Huang, Kuan-Hui Lee, Adrien Gaidon, Ehsan Adeli, and Juan~Carlos Niebles.
\newblock Spatio-temporal graph for video captioning with knowledge distillation.
\newblock In \emph{CVPR}, pages 10870--10879, 2020.

\bibitem[Paul et~al.(2018)Paul, Roy, and Roy-Chowdhury]{paul2018w}
Sujoy Paul, Sourya Roy, and Amit~K Roy-Chowdhury.
\newblock W-talc: Weakly-supervised temporal activity localization and classification.
\newblock In \emph{ECCV}, pages 563--579, 2018.

\bibitem[Pramanick et~al.(2024)Pramanick, Han, Hou, Nag, Lim, Ballas, Wang, Chellappa, and Almahairi]{pramanick2024jack}
Shraman Pramanick, Guangxing Han, Rui Hou, Sayan Nag, Ser-Nam Lim, Nicolas Ballas, Qifan Wang, Rama Chellappa, and Amjad Almahairi.
\newblock Jack of all tasks master of many: Designing general-purpose coarse-to-fine vision-language model.
\newblock In \emph{CVPR}, pages 14076--14088, 2024.

\bibitem[Prasad et~al.(2024)Prasad, Stengel-Eskin, and Bansal]{Prasad2023RephraseAR}
Archiki Prasad, Elias Stengel-Eskin, and Mohit Bansal.
\newblock Rephrase, augment, reason: Visual grounding of questions for vision-language models.
\newblock In \emph{ICLR}, 2024.

\bibitem[Qian et~al.(2024{\natexlab{a}})Qian, Li, Wu, Ye, Fei, Chua, Zhuang, and Tang]{qianmomentor}
Long Qian, Juncheng Li, Yu Wu, Yaobo Ye, Hao Fei, Tat-Seng Chua, Yueting Zhuang, and Siliang Tang.
\newblock Momentor: Advancing video large language model with fine-grained temporal reasoning.
\newblock In \emph{ICML}, 2024{\natexlab{a}}.

\bibitem[Qian et~al.(2024{\natexlab{b}})Qian, Dong, Zhang, Zang, Ding, Lin, and Wang]{qian2024videostreaming}
Rui Qian, Xiaoyi Dong, Pan Zhang, Yuhang Zang, Shuangrui Ding, Dahua Lin, and Jiaqi Wang.
\newblock Streaming long video understanding with large language models.
\newblock \emph{NeurIPS}, 37:\penalty0 119336--119360, 2024{\natexlab{b}}.

\bibitem[Qin et~al.(2024)Qin, Xiao, and Yao]{qin2024devi}
Hangyu Qin, Junbin Xiao, and Angela Yao.
\newblock Question-answering dense video events.
\newblock \emph{arXiv preprint arXiv:2409.04388}, 2024.

\bibitem[Qu et~al.(2024)Qu, Chen, Liu, Li, and Zhao]{qu2024chatvtg}
Mengxue Qu, Xiaodong Chen, Wu Liu, Alicia Li, and Yao Zhao.
\newblock Chatvtg: Video temporal grounding via chat with video dialogue large language models.
\newblock In \emph{CVPR}, pages 1847--1856, 2024.

\bibitem[Radford et~al.(2023)Radford, Kim, Xu, Brockman, McLeavey, and Sutskever]{radford2023robust}
Alec Radford, Jong~Wook Kim, Tao Xu, Greg Brockman, Christine McLeavey, and Ilya Sutskever.
\newblock Robust speech recognition via large-scale weak supervision.
\newblock In \emph{ICML}, pages 28492--28518. PMLR, 2023.

\bibitem[Redmon and Farhadi(2017)]{Redmon2016YOLO9000BF}
Joseph Redmon and Ali Farhadi.
\newblock Yolo9000: Better, faster, stronger.
\newblock In \emph{CVPR}, pages 6517--6525, 2017.

\bibitem[Redmon et~al.(2016)Redmon, Divvala, Girshick, and Farhadi]{redmon2016you}
Joseph Redmon, Santosh Divvala, Ross Girshick, and Ali Farhadi.
\newblock You only look once: Unified, real-time object detection.
\newblock In \emph{CVPR}, pages 779--788, 2016.

\bibitem[Regneri et~al.(2013)Regneri, Rohrbach, Wetzel, Thater, Schiele, and Pinkal]{regneri2013tacos}
Michaela Regneri, Marcus Rohrbach, Dominikus Wetzel, Stefan Thater, Bernt Schiele, and Manfred Pinkal.
\newblock Grounding action descriptions in videos.
\newblock \emph{TACL}, 1:\penalty0 25--36, 2013.

\bibitem[Ren et~al.(2016)Ren, He, Girshick, and Sun]{ren2016fasterrcnn}
Shaoqing Ren, Kaiming He, Ross Girshick, and Jian Sun.
\newblock Faster r-cnn: Towards real-time object detection with region proposal networks.
\newblock \emph{IEEE TPAMI}, 39\penalty0 (6):\penalty0 1137--1149, 2016.

\bibitem[Ren et~al.(2024)Ren, Yao, Li, Sun, and Hou]{ren2024timechat}
Shuhuai Ren, Linli Yao, Shicheng Li, Xu Sun, and Lu Hou.
\newblock Timechat: A time-sensitive multimodal large language model for long video understanding.
\newblock In \emph{CVPR}, pages 14313--14323, 2024.

\bibitem[Research and Center(2013)]{youdescribe}
Video~Description Research and Development Center.
\newblock Youdescribe, 2013.

\bibitem[Rezatofighi et~al.(2019)Rezatofighi, Tsoi, Gwak, Sadeghian, Reid, and Savarese]{Rezatofighi2019GeneralizedIO}
Seyed~Hamid Rezatofighi, Nathan Tsoi, JunYoung Gwak, Amir Sadeghian, Ian~D. Reid, and Silvio Savarese.
\newblock Generalized intersection over union: A metric and a loss for bounding box regression.
\newblock \emph{CVPR}, pages 658--666, 2019.

\bibitem[Rodriguez et~al.(2020)Rodriguez, Marrese-Taylor, Saleh, Li, and Gould]{rodriguez2020tmlga}
Cristian Rodriguez, Edison Marrese-Taylor, Fatemeh~Sadat Saleh, Hongdong Li, and Stephen Gould.
\newblock Proposal-free temporal moment localization of a natural-language query in video using guided attention.
\newblock In \emph{WACV}, pages 2464--2473, 2020.

\bibitem[Rodriguez-Opazo et~al.(2021)Rodriguez-Opazo, Marrese-Taylor, Fernando, Li, and Gould]{rodriguez2021dori}
Cristian Rodriguez-Opazo, Edison Marrese-Taylor, Basura Fernando, Hongdong Li, and Stephen Gould.
\newblock Dori: Discovering object relationships for moment localization of a natural language query in a video.
\newblock In \emph{WACV}, pages 1079--1088, 2021.

\bibitem[Rodriguez-Opazo et~al.(2023)Rodriguez-Opazo, Marrese-Taylor, Fernando, Takamura, and Wu]{rodriguez2023locformer}
Cristian Rodriguez-Opazo, Edison Marrese-Taylor, Basura Fernando, Hiroya Takamura, and Qi Wu.
\newblock Memory-efficient temporal moment localization in long videos.
\newblock In \emph{EACL}, pages 1909--1924, 2023.

\bibitem[Rohrbach et~al.(2012)Rohrbach, Regneri, Andriluka, Amin, Pinkal, and Schiele]{rohrbach2012script}
Marcus Rohrbach, Michaela Regneri, Mykhaylo Andriluka, Sikandar Amin, Manfred Pinkal, and Bernt Schiele.
\newblock Script data for attribute-based recognition of composite activities.
\newblock In \emph{ECCV}, pages 144--157. Springer, 2012.

\bibitem[Ross and Doll{\'a}r(2017)]{ross2017focal}
T-YLPG Ross and GKHP Doll{\'a}r.
\newblock Focal loss for dense object detection.
\newblock In \emph{CVPR}, pages 2980--2988, 2017.

\bibitem[Sharma et~al.(2020)Sharma, Agrahari, Singh, Firoj, and Mishra]{sharma2020image}
Himanshu Sharma, Manmohan Agrahari, Sujeet~Kumar Singh, Mohd Firoj, and Ravi~Kumar Mishra.
\newblock Image captioning: a comprehensive survey.
\newblock In \emph{2020 International Conference on Power Electronics \& IoT Applications in Renewable Energy and its Control (PARC)}, pages 325--328. IEEE, 2020.

\bibitem[Stefanini et~al.(2022)Stefanini, Cornia, Baraldi, Cascianelli, Fiameni, and Cucchiara]{stefanini2022show}
Matteo Stefanini, Marcella Cornia, Lorenzo Baraldi, Silvia Cascianelli, Giuseppe Fiameni, and Rita Cucchiara.
\newblock From show to tell: A survey on deep learning-based image captioning.
\newblock \emph{IEEE TPAMI}, 45\penalty0 (1):\penalty0 539--559, 2022.

\bibitem[Sun et~al.(2024)Sun, Zhou, Chen, and Xie]{sun2024trdetr}
Hao Sun, Mingyao Zhou, Wenjing Chen, and Wei Xie.
\newblock Tr-detr: Task-reciprocal transformer for joint moment retrieval and highlight detection.
\newblock In \emph{AAAI}, pages 4998--5007, 2024.

\bibitem[Sun et~al.(2023)Sun, Fang, Wu, Wang, and Cao]{sun2023evaclip}
Quan Sun, Yuxin Fang, Ledell Wu, Xinlong Wang, and Yue Cao.
\newblock Eva-clip: Improved training techniques for clip at scale.
\newblock \emph{arXiv preprint arXiv:2303.15389}, 2023.

\bibitem[Tan et~al.(2024)Tan, Lai, Zheng, and Hu]{tan2024siamgtr}
Chaolei Tan, Jianhuang Lai, Wei-Shi Zheng, and Jian-Fang Hu.
\newblock Siamese learning with joint alignment and regression for weakly-supervised video paragraph grounding.
\newblock In \emph{CVPR}, pages 13569--13580, 2024.

\bibitem[Tang et~al.(2019)Tang, Ding, Rao, Zheng, Zhang, Zhao, Lu, and Zhou]{tang2019coin}
Yansong Tang, Dajun Ding, Yongming Rao, Yu Zheng, Danyang Zhang, Lili Zhao, Jiwen Lu, and Jie Zhou.
\newblock Coin: A large-scale dataset for comprehensive instructional video analysis.
\newblock In \emph{CVPR}, pages 1207--1216, 2019.

\bibitem[Teney et~al.(2021)Teney, Abbasnejad, and van~den Hengel]{Teney21}
Damien Teney, Ehsan Abbasnejad, and Anton van~den Hengel.
\newblock Unshuffling data for improved generalization in visual question answering.
\newblock In \emph{ICCV}, pages 1397--1407, 2021.

\bibitem[Thomee et~al.(2016)Thomee, Shamma, Friedland, Elizalde, Ni, Poland, Borth, and Li]{thomee2016yfcc100m}
Bart Thomee, David~A Shamma, Gerald Friedland, Benjamin Elizalde, Karl Ni, Douglas Poland, Damian Borth, and Li-Jia Li.
\newblock Yfcc100m: The new data in multimedia research.
\newblock \emph{Communications of the ACM}, 59\penalty0 (2):\penalty0 64--73, 2016.

\bibitem[Wang et~al.(2024{\natexlab{a}})Wang, Xu, Cheng, Diao, Zhou, Cao, Wang, Ge, and Huang]{wang2024groundedvideollm}
Haibo Wang, Zhiyang Xu, Yu Cheng, Shizhe Diao, Yufan Zhou, Yixin Cao, Qifan Wang, Weifeng Ge, and Lifu Huang.
\newblock Grounded-videollm: Sharpening fine-grained temporal grounding in video large language models.
\newblock \emph{arXiv preprint arXiv:2410.03290}, 2024{\natexlab{a}}.

\bibitem[Wang et~al.(2023)Wang, Zhang, Zheng, Jiang, Cheng, and Luo]{wang2023gvl}
Teng Wang, Jinrui Zhang, Feng Zheng, Wenhao Jiang, Ran Cheng, and Ping Luo.
\newblock Learning grounded vision-language representation for versatile understanding in untrimmed videos.
\newblock \emph{arXiv preprint arXiv:2303.06378}, 2023.

\bibitem[Wang et~al.(2024{\natexlab{b}})Wang, Chen, Chen, Wu, Zhu, Zeng, Luo, Lu, Zhou, Qiao, et~al.]{wang2024visionllm}
Wenhai Wang, Zhe Chen, Xiaokang Chen, Jiannan Wu, Xizhou Zhu, Gang Zeng, Ping Luo, Tong Lu, Jie Zhou, Yu Qiao, et~al.
\newblock Visionllm: Large language model is also an open-ended decoder for vision-centric tasks.
\newblock \emph{NeurIPS}, 36, 2024{\natexlab{b}}.

\bibitem[Wang et~al.(2024{\natexlab{c}})Wang, Li, Li, Yu, He, Wang, Chen, Pei, Zheng, Xu, Wang, et~al.]{wang2024internvideo2}
Yi Wang, Kunchang Li, Xinhao Li, Jiashuo Yu, Yinan He, Chenting Wang, Guo Chen, Baoqi Pei, Rongkun Zheng, Jilan Xu, Zun Wang, et~al.
\newblock Internvideo2: Scaling video foundation models for multimodal video understanding.
\newblock \emph{arXiv preprint arXiv:2403.15377}, 2024{\natexlab{c}}.

\bibitem[Wang et~al.(2024{\natexlab{d}})Wang, Meng, Liang, Wang, Liu, and Zhao]{wang2024hawkeye}
Yueqian Wang, Xiaojun Meng, Jianxin Liang, Yuxuan Wang, Qun Liu, and Dongyan Zhao.
\newblock Hawkeye: Training video-text llms for grounding text in videos.
\newblock \emph{arXiv preprint arXiv:2403.10228}, 2024{\natexlab{d}}.

\bibitem[Wang et~al.(2022)Wang, Wang, Wu, Li, and Wu]{wang2022mmn}
Zhenzhi Wang, Limin Wang, Tao Wu, Tianhao Li, and Gangshan Wu.
\newblock Negative sample matters: A renaissance of metric learning for temporal grounding.
\newblock In \emph{AAAI}, pages 2613--2623, 2022.

\bibitem[Wu et~al.(2024)Wu, Liu, Qiao, and Sun]{HaoDenseCaptioning2024}
Hao Wu, Huabin Liu, Yu Qiao, and Xiao Sun.
\newblock Dibs: Enhancing dense video captioning with unlabeled videos via pseudo boundary enrichment and online refinement.
\newblock In \emph{CVPR}, pages 18699--18708, 2024.

\bibitem[Xiao et~al.(2021)Xiao, Shang, Yao, and Chua]{xiao2021nextqa}
Junbin Xiao, Xindi Shang, Angela Yao, and Tat-Seng Chua.
\newblock Next-qa: Next phase of question-answering to explaining temporal actions.
\newblock In \emph{CVPR}, pages 9777--9786, 2021.

\bibitem[Xiao et~al.(2022)Xiao, Zhou, Chua, and Yan]{xiao2022vgt}
Junbin Xiao, Pan Zhou, Tat-Seng Chua, and Shuicheng Yan.
\newblock Video graph transformer for video question answering.
\newblock In \emph{ECCV}, pages 39--58. Springer, 2022.

\bibitem[Xiao et~al.(2024)Xiao, Yao, Li, and Chua]{xiao2024nextgqa}
Junbin Xiao, Angela Yao, Yicong Li, and Tat-Seng Chua.
\newblock Can i trust your answer? visually grounded video question answering.
\newblock In \emph{CVPR}, pages 13204--13214, 2024.

\bibitem[Yan et~al.(2023)Yan, Xiong, Nagrani, Arnab, Wang, Ge, Ross, and Schmid]{yan2023unloc}
Shen Yan, Xuehan Xiong, Arsha Nagrani, Anurag Arnab, Zhonghao Wang, Weina Ge, David Ross, and Cordelia Schmid.
\newblock Unloc: A unified framework for video localization tasks.
\newblock In \emph{CVPR}, pages 13623--13633, 2023.

\bibitem[Yang et~al.(2022)Yang, Miech, Sivic, Laptev, and Schmid]{yang2022frozenbilm}
Antoine Yang, Antoine Miech, Josef Sivic, Ivan Laptev, and Cordelia Schmid.
\newblock Zero-shot video question answering via frozen bidirectional language models.
\newblock In \emph{NeurIPS}, pages 124--141, 2022.

\bibitem[Yang et~al.(2023)Yang, Nagrani, Seo, Miech, Pont-Tuset, Laptev, Sivic, and Schmid]{yang2023vid2seq}
Antoine Yang, Arsha Nagrani, Paul~Hongsuck Seo, Antoine Miech, Jordi Pont-Tuset, Ivan Laptev, Josef Sivic, and Cordelia Schmid.
\newblock Vid2seq: Large-scale pretraining of a visual language model for dense video captioning.
\newblock In \emph{CVPR}, pages 10714--10726, 2023.

\bibitem[Yu et~al.(2024)Yu, Cho, Yadav, and Bansal]{yu2024sevila}
Shoubin Yu, Jaemin Cho, Prateek Yadav, and Mohit Bansal.
\newblock Self-chained image-language model for video localization and question answering.
\newblock \emph{NeurIPS}, 36, 2024.

\bibitem[Yuan et~al.(2021)Yuan, Lan, Wang, Chen, Wang, and Zhu]{yuan2021charades-cd}
Yitian Yuan, Xiaohan Lan, Xin Wang, Long Chen, Zhi Wang, and Wenwu Zhu.
\newblock A closer look at temporal sentence grounding in videos: Dataset and metric.
\newblock In \emph{Proceedings of the 2nd international workshop on human-centric multimedia analysis}, pages 13--21, 2021.

\bibitem[Zala et~al.(2023)Zala, Cho, Kottur, Chen, Oguz, Mehdad, and Bansal]{zala2023hirest}
Abhay Zala, Jaemin Cho, Satwik Kottur, Xilun Chen, Barlas Oguz, Yashar Mehdad, and Mohit Bansal.
\newblock Hierarchical video-moment retrieval and step-captioning.
\newblock In \emph{CVPR}, pages 23056--23065, 2023.

\bibitem[Zellers et~al.(2022)Zellers, Lu, Lu, Yu, Zhao, Salehi, Kusupati, Hessel, Farhadi, and Choi]{zellers2022yttemporal}
Rowan Zellers, Jiasen Lu, Ximing Lu, Youngjae Yu, Yanpeng Zhao, Mohammadreza Salehi, Aditya Kusupati, Jack Hessel, Ali Farhadi, and Yejin Choi.
\newblock Merlot reserve: Neural script knowledge through vision and language and sound.
\newblock In \emph{CVPR}, pages 16375--16387, 2022.

\bibitem[Zeng et~al.(2020)Zeng, Xu, Huang, Chen, Tan, and Gan]{zeng2020drn}
Runhao Zeng, Haoming Xu, Wenbing Huang, Peihao Chen, Mingkui Tan, and Chuang Gan.
\newblock Dense regression network for video grounding.
\newblock In \emph{CVPR}, pages 10287--10296, 2020.

\bibitem[Zeng et~al.(2025)Zeng, Li, Wang, Li, Jiang, Yan, Li, Shi, Yue, Wang, et~al.]{zeng2024timesuite}
Xiangyu Zeng, Kunchang Li, Chenting Wang, Xinhao Li, Tianxiang Jiang, Ziang Yan, Songze Li, Yansong Shi, Zhengrong Yue, Yi Wang, et~al.
\newblock Timesuite: Improving mllms for long video understanding via grounded tuning.
\newblock In \emph{ICLR}, 2025.

\bibitem[Zhang et~al.(2024)Zhang, Lu, Islam, Wang, Yu, Bansal, and Bertasius]{zhang2023llovi}
Ce Zhang, Taixi Lu, Md~Mohaiminul Islam, Ziyang Wang, Shoubin Yu, Mohit Bansal, and Gedas Bertasius.
\newblock A simple llm framework for long-range video question-answering.
\newblock In \emph{EMNLP}, pages 21715--21737, 2024.

\bibitem[Zhang et~al.(2022)Zhang, Wu, and Li]{zhang2022actionformer}
Chen-Lin Zhang, Jianxin Wu, and Yin Li.
\newblock Actionformer: Localizing moments of actions with transformers.
\newblock In \emph{ECCV}, pages 492--510. Springer, 2022.

\bibitem[Zhang et~al.(2020{\natexlab{a}})Zhang, Sun, Jing, and Zhou]{zhang2020vslnet}
Hao Zhang, Aixin Sun, Wei Jing, and Joey~Tianyi Zhou.
\newblock Span-based localizing network for natural language video localization.
\newblock In \emph{ACL}, pages 6543--6554, 2020{\natexlab{a}}.

\bibitem[Zhang et~al.(2023)Zhang, Li, and Bing]{zhang2023videollama}
Hang Zhang, Xin Li, and Lidong Bing.
\newblock Video-llama: An instruction-tuned audio-visual language model for video understanding.
\newblock In \emph{EMNLP}, pages 543--553, 2023.

\bibitem[Zhang and Peng(2019)]{zhang2019object}
Junchao Zhang and Yuxin Peng.
\newblock Object-aware aggregation with bidirectional temporal graph for video captioning.
\newblock In \emph{CVPR}, pages 8327--8336, 2019.

\bibitem[Zhang et~al.(2020{\natexlab{b}})Zhang, Peng, Fu, and Luo]{zhang20202dtan}
Songyang Zhang, Houwen Peng, Jianlong Fu, and Jiebo Luo.
\newblock Learning 2d temporal adjacent networks for moment localization with natural language.
\newblock In \emph{AAAI}, pages 12870--12877, 2020{\natexlab{b}}.

\bibitem[Zhang et~al.(2020{\natexlab{c}})Zhang, Peng, Fu, and Luo]{zhang20203dtpn}
Songyang Zhang, Houwen Peng, Jianlong Fu, and Jiebo Luo.
\newblock Learning 2d temporal adjacent networks for moment localization with natural language.
\newblock In \emph{AAAI}, pages 12870--12877, 2020{\natexlab{c}}.

\bibitem[Zhang et~al.(2021)Zhang, Peng, Fu, Lu, and Luo]{zhang2021ms2dtan}
Songyang Zhang, Houwen Peng, Jianlong Fu, Yijuan Lu, and Jiebo Luo.
\newblock Multi-scale 2d temporal adjacency networks for moment localization with natural language.
\newblock \emph{IEEE TPAMI}, 44\penalty0 (12):\penalty0 9073--9087, 2021.

\bibitem[Zhang et~al.(2019)Zhang, Lin, Zhao, and Xiao]{zhang2019cmin}
Zhu Zhang, Zhijie Lin, Zhou Zhao, and Zhenxin Xiao.
\newblock Cross-modal interaction networks for query-based moment retrieval in videos.
\newblock In \emph{ACM SIGIR}, pages 655--664, 2019.

\bibitem[Zhao et~al.(2023)Zhao, Misra, Kr{\"a}henb{\"u}hl, and Girdhar]{zhao2022lavila}
Yue Zhao, Ishan Misra, Philipp Kr{\"a}henb{\"u}hl, and Rohit Girdhar.
\newblock Learning video representations from large language models.
\newblock In \emph{CVPR}, pages 6586--6597, 2023.

\bibitem[Zhou et~al.(2018{\natexlab{a}})Zhou, Xu, and Corso]{zhou2018youcook2}
Luowei Zhou, Chenliang Xu, and Jason Corso.
\newblock Towards automatic learning of procedures from web instructional videos.
\newblock In \emph{AAAI}, 2018{\natexlab{a}}.

\bibitem[Zhou et~al.(2018{\natexlab{b}})Zhou, Xu, and Corso]{ZhXuCoCVPR18}
Luowei Zhou, Chenliang Xu, and Jason~J Corso.
\newblock Towards automatic learning of procedures from web instructional videos.
\newblock In \emph{AAAI}, 2018{\natexlab{b}}.

\bibitem[Zhou et~al.(2018{\natexlab{c}})Zhou, Zhou, Corso, Socher, and Xiong]{Zhou2018EndtoEndDV}
Luowei Zhou, Yingbo Zhou, Jason~J. Corso, Richard Socher, and Caiming Xiong.
\newblock End-to-end dense video captioning with masked transformer.
\newblock \emph{CVPR}, pages 8739--8748, 2018{\natexlab{c}}.

\bibitem[Zhou et~al.(2018{\natexlab{d}})Zhou, Zhou, Corso, Socher, and Xiong]{zhou2018end}
Luowei Zhou, Yingbo Zhou, Jason~J Corso, Richard Socher, and Caiming Xiong.
\newblock End-to-end dense video captioning with masked transformer.
\newblock In \emph{CVPR}, pages 8739--8748, 2018{\natexlab{d}}.

\bibitem[Zhou et~al.(2019)Zhou, Wang, and Kr{\"a}henb{\"u}hl]{zhou2019objects}
Xingyi Zhou, Dequan Wang, and Philipp Kr{\"a}henb{\"u}hl.
\newblock Objects as points.
\newblock In \emph{arXiv preprint arXiv:1904.07850}, 2019.

\bibitem[Zhou et~al.(2024)Zhou, Arnab, Buch, Yan, Myers, Xiong, Nagrani, and Schmid]{zhou2024streaming}
Xingyi Zhou, Anurag Arnab, Shyamal Buch, Shen Yan, Austin Myers, Xuehan Xiong, Arsha Nagrani, and Cordelia Schmid.
\newblock Streaming dense video captioning.
\newblock In \emph{CVPR}, pages 18243--18252, 2024.

\bibitem[Zhu et~al.(2022{\natexlab{a}})Zhu, Pang, Thapliyal, Wang, and Soricut]{Zhu2022EndtoendDV}
Wanrong Zhu, Bo Pang, Ashish~V. Thapliyal, William~Yang Wang, and Radu Soricut.
\newblock End-to-end dense video captioning as sequence generation.
\newblock \emph{ArXiv}, abs/2204.08121, 2022{\natexlab{a}}.

\bibitem[Zhu et~al.(2022{\natexlab{b}})Zhu, Pang, Thapliyal, Wang, and Soricut]{zhu-etal-2022-end}
Wanrong Zhu, Bo Pang, Ashish~V. Thapliyal, William~Yang Wang, and Radu Soricut.
\newblock End-to-end dense video captioning as sequence generation.
\newblock In \emph{Proceedings of the 29th International Conference on Computational Linguistics}, pages 5651--5665, Gyeongju, Republic of Korea, 2022{\natexlab{b}}. International Committee on Computational Linguistics.

\bibitem[Zhukov et~al.(2019)Zhukov, Alayrac, Cinbis, Fouhey, Laptev, and Sivic]{zhukov2019crosstask}
Dimitri Zhukov, Jean-Baptiste Alayrac, Ramazan~Gokberk Cinbis, David Fouhey, Ivan Laptev, and Josef Sivic.
\newblock Cross-task weakly supervised learning from instructional videos.
\newblock In \emph{CVPR}, pages 3537--3545, 2019.

\end{thebibliography}
}

\newpage
\clearpage
\appendix
\counterwithin{figure}{section}
\numberwithin{table}{section}


\twocolumn[
\begin{center}
    \LARGE \bf Appendix
\end{center}
\vspace{2mm}
]


\xpar{Supplementary material contents}
This supplementary document is structured as follows:
Section \ref{sec:qualitative} visualizes additional qualitative results which aim to provide further insight into \model's function and performance;
Section \ref{sec:ablations} provides additional ablations;
Section \ref{sec:comparison_specialist_baselines} provides additional comparison with task-specific specialist baselines for fine-tuned STG task;
Section \ref{sec:pseudolabel_gen} presents more details on the pseudo-label generation process;
Section \ref{sec:example_instructions} discusses the instructions used for different tasks;
Section \ref{sec:error_analysis} presents some failure cases;
Section \ref{sec:impl_details} explains our hyper-parameter selection;
Section \ref{sec:dataset_details} provides more details on the processing of all datasets used for training
(Section \ref{sec:dataset_details_pretraining}) and fine-tuning and evaluation (Section \ref{sec:dataset_details_ft_and_eval}).

\section{Additional Qualitative Results}
\begin{figure*}[!t]
\centering
\includegraphics[width=\textwidth]{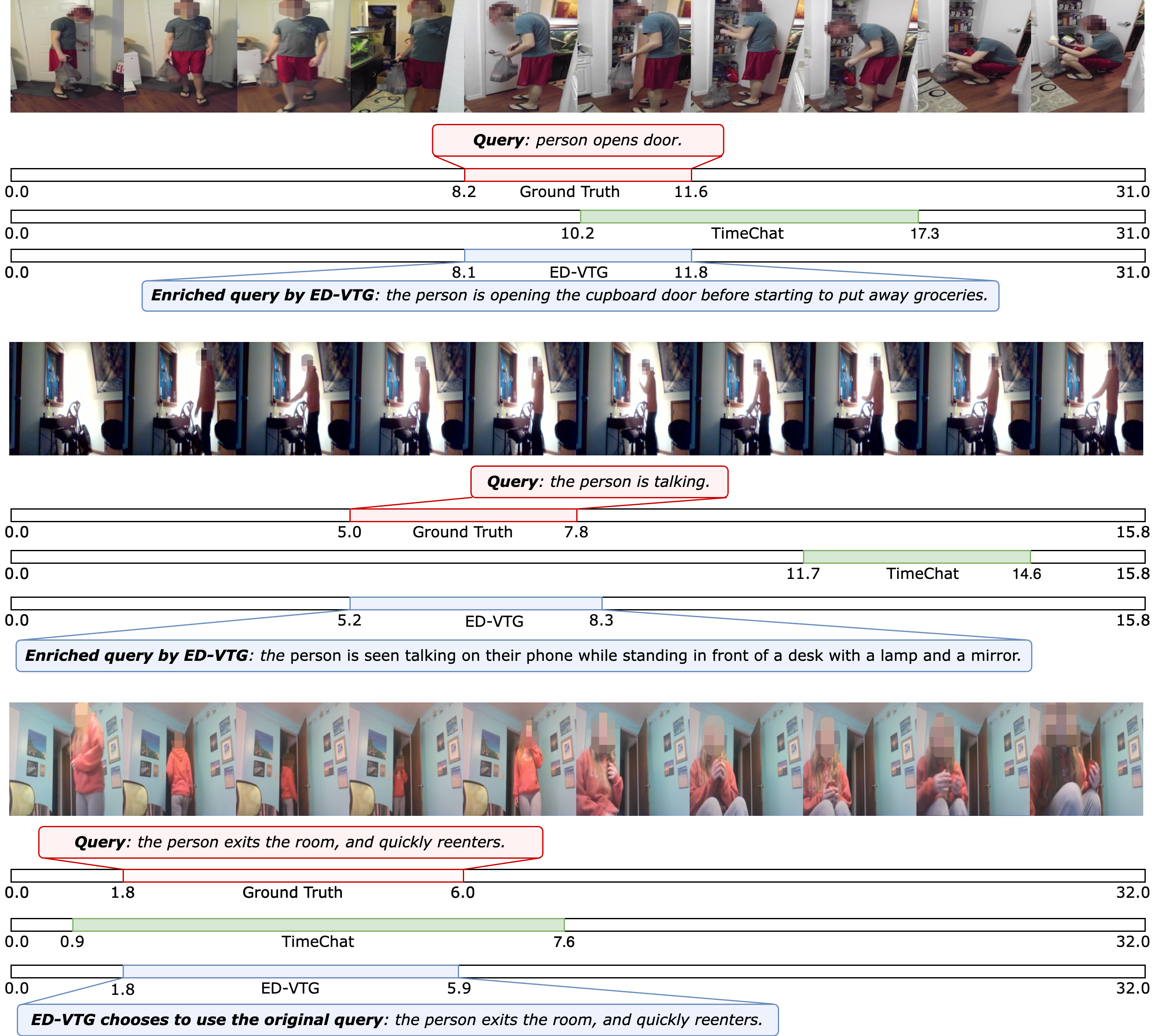}
\vspace{-3mm}
\caption{\textbf{Examples of query enrichment and localization made by \model\ on single-query temporal grounding (STG) task from the Charades-STA \cite{gao2017charadessta} dataset.} We also show the prediction made by one baseline model, TimeChat \cite{ren2024timechat}, which directly ground the input queries using raw-text timestamp representation. Since we train \model\ using the MIL paradigm, the model can choose to use the input query directly or enrich it during evaluation. In the last example, since the input query is clear and explicit, the model directly localizes it.}
\vspace{-3mm}
\label{fig:comparison_timechat_stg}
\end{figure*}

\begin{figure*}[!t]
\centering
\includegraphics[width=\textwidth]{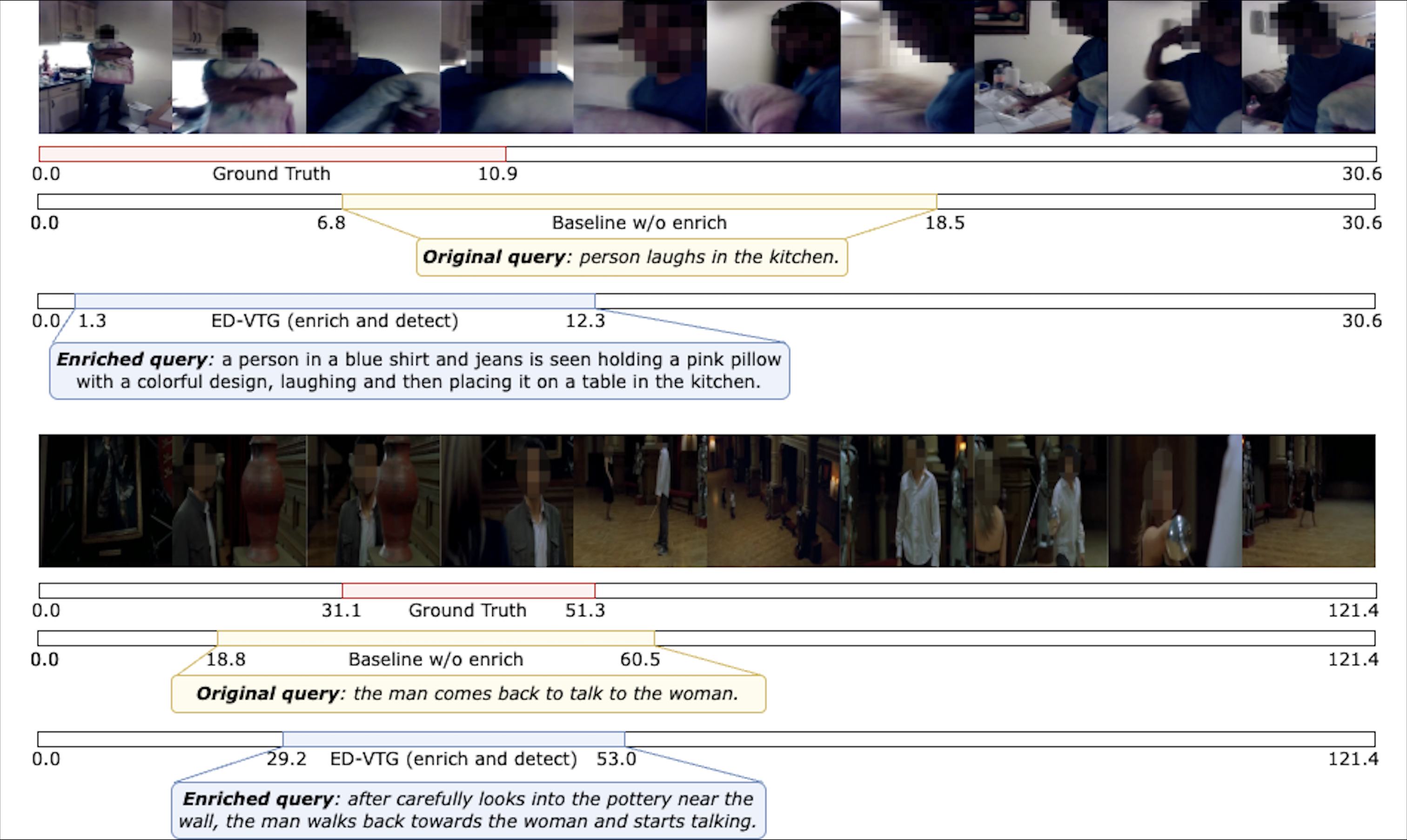}
\vspace{-3mm}
\caption{
\textbf{Comparison of detections of \model~using its predicted enriched queries against a baseline version trained to always use the original queries.} The enriched queries contain additional relevant details and context that enable \model~to perform more accurate temporal localization. In the first example, which is taken from Charades-STA~\cite{gao2017charadessta}, the additional details in the enriched query provide a more complete description of objects and actions that is more easily groundable. In the second example, sourced from the ActivityNet-Captions~\cite{krishna2017dense} dataset, the enriched query provides additional temporal context which leads to more precise temporal boundary prediction.
}
\vspace{-3mm}
\label{fig:comparison_without_enrich_stg}
\end{figure*}

We show qualitative examples in Figure~\ref{fig:comparison_timechat_stg}, where we compare \model's predictions to the TimeChat \cite{ren2024timechat} baseline and the ground truth annotations.
\model~is trained with MIL, therefore during inference it can choose to enrich the original query if it is incomplete, or use ts as is when it is sufficient.
In Figure~\ref{fig:comparison_without_enrich_stg} we show example detections of \model\ using the predicted enriched queries and a baseline version where a model with the same architecture is trained to always use the original queries.
These examples clearly demonstrate how the enriched queries often contain relevant details that enable ED-VTG to perform more accurate temporal localization than the baseline.
\label{sec:qualitative}

\section{Additional Ablation Study}
\label{sec:ablations}

We conduct additional ablation experiments on two different training augmentations for query transformations compared to our cascaded enrich and detect setup, and report zero-shot numbers with increasing amount of pre-training data, showing the scalability of \model.

\begin{table}[!t]
\centering

\small
\setlength{\tabcolsep}{4pt}
\resizebox{\columnwidth}{!}{\begin{tabular}{l | c c c | c c c}

\toprule

\multirow{2}{*}{\bf Training Paradigm} & \multicolumn{3}{c |}{\bf \centering Charades-STA \VTG} & \multicolumn{3}{c}{\bf \centering ANet-Captions \VTG} \\

& R@0.3 & R@0.5 & mIoU & R@0.3 & R@0.5 & mIoU  \\

\midrule

Detect & 51.4 & 31.5 & 33.2 & 50.3 & 30.1 & 34.0 \\
Offline Paraphrasing + Detect & 51.4 & 31.6 & 32.7 & 50.5 & 30.8 & 33.9  \\
Offline Enrich w/o Interval Anno. + Detect & 51.7 & 31.1 & 31.9 & 49.5 & 29.1 & 32.9 \\
Offline Enrich + Detect & 51.7 & 31.5 & 33.4 & 49.8 & 29.9 & 33.7 \\
\rowcolor{Light}
Enrich \& Detect & \bf 60.1 & \bf 37.0 & \bf 38.4 & \bf 56.3 & \bf 35.5 & \bf 37.8  \\

\bottomrule

\end{tabular}}
\vspace{-3mm}
\caption{\textbf{Ablation on enrichment as a training pre-processing
step.} We compare the proposed enrich \& detect framework with two additional augmentations using LLMs. In the ``Offline Paraphrasing + Detect" setup, we use a blind LLaMA 3.1 8B \cite{dubey2024llama3} to paraphrase and grammatically correct the input queries. In the ``Offline Enrich w/o Interval Annotation + Detect" setup, we augment the queries with LLaVA OneVision 72B \cite{li2024llavaonevision} as pre-processing, where the model sees the video, but does not have access to the ground truth labels. We observe that the proposed enrich \& detect is superior since the trained model learns to perform autonomous enrichment during evaluation, which proves that the cascaded detection paradigm is significantly different than training augmentation. Reported results are in FT w/o PT setting. \vspace{-2mm}}
\label{tab:ablation_offline_paraphrase}
\end{table}

\vspace{1mm}
\noindent \textbf{Offline Query Paraphrasing.} In this setup, we use a blind LLaMA 3.1 8B \cite{dubey2024llama3} to paraphrase and grammatically correct the input queries in the training set. Notably, the LLaMA model is text-only, and does not have access to the video, and hence can not \textit{enrich} the queries, but just paraphrases them for better grammatical construction. During evaluation, we also augment the queries in the same fashion. As shown in Table \ref{tab:ablation_offline_paraphrase}, such an augmentation techniques does not bring any notable improvement on Charades and ActivityNet datasets for STG task.

\vspace{1mm}
\noindent \textbf{Offline Query Enrichment w/o Annotated Intervals.} In this second setup, we employ a multimodal LLaVA OneVision 72B model \cite{li2024llavaonevision} for query enrichment as a form of training augmentation. Unlike the approach in Table 8 of the main paper, we do not crop the input video to the ground-truth interval in this setup. As a result, the model often incorporates irrelevant contextual information into the query, which is not helpful for localizing the desired interval. Consequently, as shown in Table \ref{tab:ablation_offline_paraphrase}, this type of augmentation negatively impacts model performance. Overall, these ablation experiments demonstrate that our proposed enrich \& detect approach is fundamentally different from training augmentations using LLMs. The trained model can independently enrich queries with necessary details or choose to directly ground the input query.

\noindent \textbf{Pre-training Dataset Size.} Table \ref{tab:pretraining_data} shows the effect of increasing training data on zero-shot Charades-STA STG and NExT-GQA QG datasets. We perform best when incorporating all tasks and datasets, denoting the usefulness of unified pre-training.

\begin{table}[!t]
\vspace{-2mm}
\centering

\small
\setlength{\tabcolsep}{4pt}
\resizebox{\columnwidth}{!}{\begin{tabular}{l c | c c c | c c}

\toprule

\multirow{2}{*}{\bf Pre-training Tasks} & \bf \# Samples & \multicolumn{3}{c |}{\bf \centering Charades-STA \VTG} & \multicolumn{2}{c}{\bf \centering NExT-GQA QG} \\

& & R@0.3 & R@0.5 & mIoU & mIoP & mIoU  \\

\midrule

STG & 91.8K & 55.3 & 35.9 & 37.0 & 32.5 & 24.8 \\
STG + VPG & 133.4K & 59.0 & 38.7 & 39.8 & 34.1 & 26.1 \\
STG + VPG + AG & 136K & 59.5 & 39.1 & 39.9 & 34.2 & 26.6 \\

\bottomrule

\end{tabular}}
\vspace{-1mm}
\caption{\textbf{Ablation on the number of pre-training tasks and samples.} We receive the best scores when using all tasks together, showing the benefit of unified pre-training and model's scalability. Reported results are in zero-shot setting.}
\label{tab:pretraining_data}
\vspace{-7mm}
\end{table}

\begin{figure}[!t]
\centering
 \begin{subfigure}[b]{0.47\columnwidth}
\centering
\includegraphics[width=\textwidth]{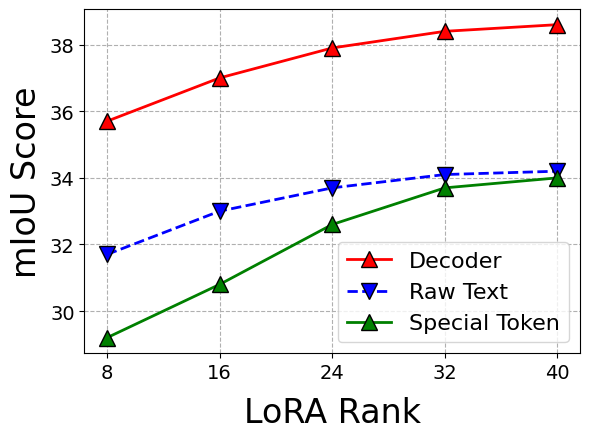}
\vspace{-4mm}
\caption{Results on \textbf{Charades \VTG}.}
\label{fig:caption_charades}
 \end{subfigure}
 \hspace{0.003\textwidth}
 \begin{subfigure}[b]{0.47\columnwidth}
\centering
\includegraphics[width=\textwidth]{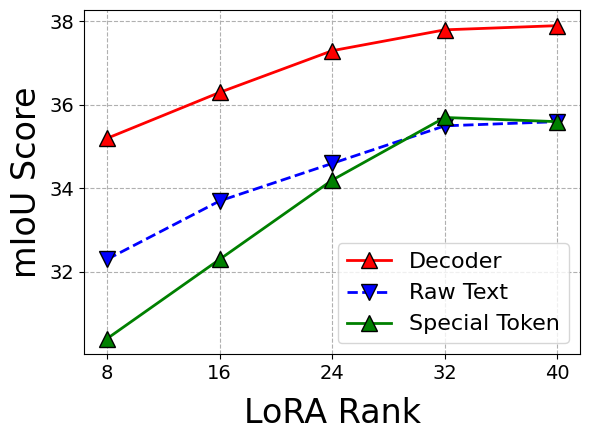}
\hspace{6mm}
\vspace{-4mm}
\caption{Results on \textbf{ActivityNet \VTG}.}
\label{fig:caption_activitynet}
 \end{subfigure}

\vspace{-1mm}
\caption{\textbf{Ablation study on timestamp representation by the interval decoder.} We compare performance of our proposed lightweight decoder vs timestamp as raw text \cite{huang2024vtimellm, ren2024timechat, li2024groundinggpt, ma2023llavilo} vs timestamp representation by special tokens \cite{huang2025lita, qianmomentor, wang2024groundedvideollm}, and find the decoder to be significantly better than both other techniques. Reported results are in FT w/o PT setting.}

\label{fig:decoder_rawtext_specialtoken_ablation}
\vspace{-5mm}
\end{figure}

\noindent \textbf{Comparison of Latency.} We compare the inference speed of \model\ with and without the interval decoder on the Charades \VTG\ benchmark in ZS setting. Using the same compute infrastructure and averaging over 3 evaluation runs, the model without decoder requires 2.10 seconds for every sample, while with decoder, it spends 2.15 seconds. Moreover, the training speeds of both models are similar, with the decoder adding only a negligible 0.2\% to the total trainable parameters. This suggests that incorporating the decoder has a minimal impact on the model's latency.

\vspace{1mm}
\noindent \textbf{Effect of interval decoder.} We examine the impact of different timestamp representations in Figure \ref{fig:decoder_rawtext_specialtoken_ablation}, comparing our lightweight decoder to using raw text or special tokens for generating time intervals. For this analysis, we fine-tune the Video-LLaMA checkpoint on the Charades and ActivityNet \VTG\ benchmarks, as shown in Figures \ref{fig:caption_charades} and \ref{fig:caption_activitynet}. Both datasets exhibit noticeable performance degradation when the decoder is omitted. Additionally, using hundreds of special tokens increases training complexity, leading to significantly poorer results at lower LoRA ranks. Since numeric digits or tokens representing frame indices lack a causal relationship in autoregressive generation, the decoder facilitates a more efficient training process. Furthermore, introducing tailored grounding objectives enables the model to produce precise timestamps.

\vspace{-3mm}
\section{Comparison with Specialist Baselines}
\label{sec:comparison_specialist_baselines}

Table \ref{tab:stg_fine_tune_supplementary} extensively compares the \model with various task-specific specialist models for the fine-tuned STG task on Charades-STA, ActivityNet-Captions, and TACoS dataset. On Charades, \model\ beats strong specialist baselines like UnLoc \cite{yan2023unloc}, UniVTG \cite{lin2023univtg}, MomentDiff \cite{li2024momentdiff}, QD-DETR \cite{moon2023qddetr}, CG-DETR \cite{moon2023cgdetr}, etc., while models like EMB \cite{huang2022emb}, EaTR \cite{jang2023eatr}, and SG-DETR \cite{gordeev2024sgdetr} perform better than ours. We observe a similar trend on the other two benchmarks. However, since the specialist models are often tailored to a particular task and dataset, they usually show poor transferability, whereas \model\ demonstrates state-of-the-art zero-shot performance, as shown in Table 2 of our main paper. Nevertheless, the strong performance by \model\ on fine-tuning setting significantly closes the gap between MLLMs and specialist baselines.

\begin{table*}[!t]
\centering

\small
\setlength{\tabcolsep}{4pt}
\resizebox{\textwidth}{!}{\begin{tabular}{l c c c | c c c c | c c c c | c c c c}

\toprule

\multirow{2}{*}{\bf Method} & \multirow{2}{1.5 cm}{\bf \centering Generalist Model} & \multirow{2}{1.4cm}{\bf \centering \# Train Samples} & \multirow{2}{*}{\bf \centering Eval.} & \multicolumn{4}{c|}{\bf Charades-STA} & \multicolumn{4}{c|}{\bf ActivityNet-Captions} & \multicolumn{4}{c}{\bf TACoS} \\

& & & & R@0.3 & R@0.5 & R@0.7 & mIoU & R@0.3 & R@0.5 & R@0.7 & mIoU & R@0.3 & R@0.5 & R@0.7 & mIoU  \\

\midrule

\demph{VSLNet (C3D) \cite{zhang2020vslnet}} & \demph{\ding{55}} & \demph{$-$} & \demph{FT} & \demph{64.3} & \demph{47.3} & \demph{30.2} & \demph{45.2} & \demph{63.2} & \demph{43.2} & \demph{26.2} & \demph{43.2} & \demph{29.6} & \demph{24.3} & \demph{20.0} & \demph{24.1} \\

\demph{CTRL \cite{gao2017ctrl}} & \demph{\ding{55}} & \demph{$-$} & \demph{FT} & \demph{$-$} & \demph{23.6} & \demph{8.9} & \demph{$-$} & \demph{$-$} & \demph{$-$} & \demph{$-$} & \demph{$-$} & \demph{18.3} & \demph{13.3} & \demph{$-$} & \demph{$-$} \\

\demph{GTR-H \cite{cao2021gtr}} & \demph{\ding{55}} & \demph{$-$} & \demph{FT} & \demph{$-$} & \demph{62.6} & \demph{39.7} & \demph{$-$} & \demph{$-$} & \demph{50.6} & \demph{29.1} & \demph{$-$} & \demph{$-$} & \demph{40.4} & \demph{30.2} & \demph{$-$} \\
\demph{2D-TAN \cite{zhang20202dtan}} & \demph{\ding{55}} & \demph{$-$} & \demph{FT} & \demph{57.3} & \demph{45.8} & \demph{27.9} & \demph{41.1} & \demph{60.3} & \demph{43.4} & \demph{25.0} & \demph{42.5} & \demph{40.0} & \demph{28.0} & \demph{12.9} & \demph{27.2} \\
\demph{MS-2D-TAN (I3D) \cite{zhang2021ms2dtan}} & \demph{\ding{55}} & \demph{$-$} & \demph{FT} & \demph{$-$} & \demph{56.6} & \demph{36.2} & \demph{$-$} & \demph{62.1} & \demph{45.5} & \demph{28.3} & \demph{$-$} & \demph{42.0} & \demph{33.6} & \demph{22.1} & \demph{$-$} \\
\demph{Moment-DETR \cite{lei2021momentdetr}} & \demph{\ding{55}} & \demph{236K} & \demph{FT} & \demph{65.8} & \demph{52.1} & \demph{30.6} & \demph{45.5} & \demph{$-$} & \demph{$-$} & \demph{$-$} & \demph{$-$} & \demph{38.0} & \demph{24.7} & \demph{12.0} & \demph{25.5} \\

\demph{UMT$^{\dagger}$ \cite{liu2022umt}} & \demph{\ding{55}} & \demph{236K} & \demph{FT} & \demph{$-$} & \demph{48.3} & \demph{29.3} & \demph{$-$} & \demph{$-$} & \demph{$-$} & \demph{$-$} & \demph{$-$} & \demph{$-$} & \demph{$-$} & \demph{$-$} & \demph{$-$} \\

\demph{UnLoc-B \cite{yan2023unloc}} & \demph{\ding{55}} & \demph{650K} & \demph{FT} & \demph{$-$} & \demph{58.1} & \demph{35.4} & \demph{$-$} & \demph{$-$} & \demph{48.0} & \demph{29.7} & \demph{$-$} & \demph{$-$} & \demph{$-$} & \demph{$-$} & \demph{$-$} \\
\demph{MomentDiff \cite{li2024momentdiff}} & \demph{\ding{55}} & \demph{$-$} & \demph{FT} & \demph{$-$} & \demph{55.6} & \demph{32.4} & \demph{$-$} & \demph{$-$} & \demph{$-$} & \demph{$-$} & \demph{$-$} & \demph{46.6} & \demph{28.9} & \demph{12.4} & \demph{30.4} \\
\demph{LGI \cite{mun2020lgi}} & \demph{\ding{55}} & \demph{$-$} & \demph{FT} & \demph{73.0} & \demph{59.5} & \demph{35.5} & \demph{51.4} & \demph{58.5} & \demph{41.5} & \demph{23.1} & \demph{41.1} & \demph{$-$} & \demph{$-$} & \demph{$-$} & \demph{$-$} \\
\demph{FlashVTG (SF+C) \cite{cao2024flashvtg}} & \demph{\ding{55}} & \demph{$-$} & \demph{FT} & \demph{$-$} & \demph{60.1} & \demph{38.0} & \demph{$-$} & \demph{$-$} & \demph{$-$} & \demph{$-$} & \demph{$-$} & \demph{53.7} & \demph{41.8} & \demph{24.7} & \demph{37.6} \\
\demph{BAM-DETR \cite{lee2024bamdetr}} & \demph{\ding{55}} & \demph{$-$} & \demph{FT} & \demph{72.9} & \demph{60.0} & \demph{39.4} & \demph{52.3} & \demph{$-$} & \demph{$-$} & \demph{$-$} & \demph{$-$} & \demph{56.7} & \demph{41.5} & \demph{26.8} & \demph{39.3} \\
\demph{UniVTG \cite{lin2023univtg}} & \demph{\ding{55}} & \demph{4.2M} & \demph{FT} & \demph{70.8} & \demph{58.0} & \demph{35.7} & \demph{50.1} & \demph{$-$} & \demph{$-$} & \demph{$-$} & \demph{$-$} & \demph{51.4} & \demph{35.0} & \demph{17.4} & \demph{33.6} \\

\demph{QD-DETR (SF+C) \cite{moon2023qddetr}} & \demph{\ding{55}} & \demph{$-$} & \demph{FT} & \demph{$-$} & \demph{57.3} & \demph{32.6} & \demph{$-$} & \demph{$-$} & \demph{$-$} & \demph{$-$} & \demph{$-$} & \demph{$-$} & \demph{$-$} & \demph{$-$} & \demph{$-$} \\

\demph{CG-DETR (SF+C) \cite{moon2023cgdetr}} & \demph{\ding{55}} & \demph{$-$} & \demph{FT} & \demph{70.4} & \demph{58.4} & \demph{36.3} & \demph{50.1} & \demph{$-$} & \demph{$-$} & \demph{$-$} & \demph{$-$} & \demph{54.4} & \demph{39.5} & \demph{23.4} & \demph{37.4} \\
\demph{TR-DETR (SF+C) \cite{sun2024trdetr}} & \demph{\ding{55}} & \demph{$-$} & \demph{FT} & \demph{$-$} & \demph{57.6} & \demph{33.5} & \demph{$-$} & \demph{$-$} & \demph{$-$} & \demph{$-$} & \demph{$-$} & \demph{$-$} & \demph{$-$} & \demph{$-$} & \demph{$-$} \\
\demph{GVL (C3D) \cite{wang2023gvl}} & \demph{\ding{55}} & \demph{$-$} & \demph{FT} & \demph{$-$} & \demph{$-$} & \demph{$-$} & \demph{$-$} & \demph{$-$} & \demph{48.9} & \demph{27.2} & \demph{46.4} & \demph{45.9} & \demph{34.6} & \demph{$-$} & \demph{32.5} \\
\demph{InternVideo2$^{\star}$ + CG-DETR \cite{wang2024internvideo2}} & \demph{\ding{55}} & \demph{2.1M} & \demph{FT} & \demph{79.7} & \demph{70.0} & \demph{48.9} & \demph{58.8} & \demph{$-$} & \demph{$-$} & \demph{$-$} & \demph{$-$} & \demph{$-$} & \demph{$-$} & \demph{$-$} & \demph{$-$} \\

\demph{SG-DETR \cite{gordeev2024sgdetr}} & \demph{\ding{55}} & \demph{$-$} & \demph{FT} & \demph{$-$} & \demph{71.1} & \demph{52.8} & \demph{60.7} & \demph{$-$} & \demph{$-$} & \demph{$-$} & \demph{$-$} & \demph{$-$} & \demph{46.4} & \demph{33.9} & \demph{42.4} \\

\demph{MGSL-Net \cite{liu2022mgslnet}} & \demph{\ding{55}} & \demph{150K} & \demph{FT} & \demph{$-$} & \demph{64.0} & \demph{41.0} & \demph{$-$} & \demph{$-$} & \demph{51.9} & \demph{31.4} & \demph{$-$} & \demph{42.5} & \demph{32.3} & \demph{$-$} & \demph{$-$} \\

\demph{EaTR \cite{jang2023eatr}} & \demph{\ding{55}} & \demph{150K} & \demph{FT} & \demph{$-$} & \demph{68.5} & \demph{44.9} & \demph{$-$} & \demph{$-$} & \demph{58.1} & \demph{37.6} & \demph{$-$} & \demph{$-$} & \demph{$-$} & \demph{$-$} & \demph{$-$} \\

\demph{EMB (ELA) \cite{huang2022emb}} & \demph{\ding{55}} & \demph{$-$} & \demph{FT} & \demph{79.7} & \demph{69.2} & \demph{51.4} & \demph{62.2} & \demph{73.7} & \demph{58.7} & \demph{40.7} & \demph{56.2} & \demph{63.3} & \demph{52.5} & \demph{37.0} & \demph{48.4} \\

\midrule

BLIP-2 (frames only) \cite{li2023blip2} & \ding{51} & 129M & FT & $-$ &  43.3 & \underline{32.6} & $-$ & $-$ & 25.8 & 9.7 & $-$ & $-$ & $-$ & $-$ & $-$ \\
VideoChat2 \cite{li2024videochat2} & \ding{51} & 2M & FT & $-$ & $-$ & $-$ & $-$ & 55.5 & \underline{34.7} & 17.7 & 38.9 & $-$ & $-$ & $-$ & $-$ \\
TimeChat \cite{ren2024timechat} & \ding{51} & 125K & FT & $-$ & 46.7 & 23.7 & $-$ & $-$ & $-$ & $-$ & $-$ & \underline{27.7} & \underline{15.1} & \underline{6.4} & \underline{18.4} \\
HawkEye \cite{wang2024hawkeye} & \ding{51} & 715K & FT & \underline{72.5} & \underline{58.3} & 28.8 & \underline{49.3} & \underline{55.9} & \underline{34.7} & \underline{17.9} & \underline{39.1} & $-$ & $-$ & $-$ & $-$ \\
VtimeLLM \cite{huang2024vtimellm} & \ding{51} & 170K & FT & $-$ & $-$ & $-$ & $-$ & $-$ & $-$ & $-$ & $-$ & 26.8 & 14.4 & 6.1 & 18.0 \\

\rowcolor{Light}
\model & \ding{51} & 136K & FT & \bf 78.2 & \bf 62.1 & \bf 35.0 & \bf 52.6 & \bf 67.6 & \bf 45.1 & \bf 22.7 & \bf 44.9 & \bf 46.0 & \bf 31.5 & \bf 15.8 & \bf 32.4 \\

\midrule

\bf \textcolor{blue}{$\Delta_{\text{Ours - HawkEye}}$} & $-$ & $-$ & FT & \textcolor{blue}{5.7} \textcolor{blue}{$\uparrow$} & \textcolor{blue}{3.8} \textcolor{blue}{$\uparrow$} & \textcolor{blue}{6.2} \textcolor{blue}{$\uparrow$} & \textcolor{blue}{3.3} \textcolor{blue}{$\uparrow$} & \textcolor{blue}{11.7} \textcolor{blue}{$\uparrow$} & \textcolor{blue}{10.4} \textcolor{blue}{$\uparrow$} & \textcolor{blue}{4.8} \textcolor{blue}{$\uparrow$} & \textcolor{blue}{5.8} \textcolor{blue}{$\uparrow$} &\textcolor{blue}{$-$} & \textcolor{blue}{$-$} & \textcolor{blue}{$-$}& \textcolor{blue}{$-$} \\
\bf \textcolor{blue}{$\Delta_{\text{Ours - VTimeLLM}}$} & $-$ & $-$ & FT & \textcolor{blue}{$-$} & \textcolor{blue}{$-$} & \textcolor{blue}{$-$} & \textcolor{blue}{$-$} & \textcolor{blue}{$-$} & \textcolor{blue}{$-$} & \textcolor{blue}{$-$}& \textcolor{blue}{$-$} & \textcolor{blue}{19.2} \textcolor{blue}{$\uparrow$} & \textcolor{blue}{17.1} \textcolor{blue}{$\uparrow$} & \textcolor{blue}{9.7} \textcolor{blue}{$\uparrow$} & \textcolor{blue}{14.4} \textcolor{blue}{$\uparrow$} \\

\bottomrule

\end{tabular}}
\vspace{-2mm}
\caption{\textbf{Extension of Table 3 in the main paper with a comprehensive list of task-specific specialist baselines.} \model\ beats many expert baselines, and significantly closes the gap between SOTA specialist models with MLLMs. $^{\dagger}$UMT uses video and audio as the input. $^{\star}$Though InterVideo2 is a generalist model, it fine-tunes CG-DETR \cite{moon2023cgdetr} head for grounding tasks, using the LLM only as a video feature extractor.}
\label{tab:stg_fine_tune_supplementary}
\end{table*}

\section{Pseudo-label Generation Pipeline}
\label{sec:pseudolabel_gen}

Since our proposed two-step cascaded grounding approach, Enrich and Detect, requires enriched queries as ground truths during training, we augment poorly worded or potentially incomplete input queries of all training benchmarks with additional context information using an open-source and broadly capable captioning model, LLaVA OneVision (OV) 72B \cite{li2024llavaonevision}. First, we crop the input videos between the annotated time intervals. Next, we input the original query and the cropped video to the OV model and ask it to enrich the description of the activities in the given segment while preserving the main focus of the original query. The prompt used in this step is shown in Figure \ref{fig:onevision_prompt_enrichment}. To partially tackle the hallucination issue of large LLMs during language generation, next we generate a few binary choice questions from each enriched query using a text-only LLaMA 3.1 8B model \cite{dubey2024llama3}, and filter the samples using a lower-sized OV 8B model, which is proficient at answering yes/no questions. If all descriptions in the enriched query are correct, we keep the sample; otherwise, we reiterate the process. Notably, even with our well-versed query augmentation pipeline, some enriched samples contain unimportant information for grounding, which we tackle with the proposed MIL training framework. During evaluation, we only feed the original queries as input to \model, and the model generates the enriched queries and perform grounding.


 \begin{figure}[!t]
 \centering
 \begin{minipage}{0.95\columnwidth} %
\begin{lstlisting}[
     basicstyle=\ttfamily\footnotesize,
     breaklines=true,
     tabsize=1, % <--- added this line (adjust the value as needed)
     breakindent=0em, % <--- added this line (adjust the value as needed)
     breakautoindent=true, % <--- added this line
     columns=fullflexible, % <--- added this line
     frame=single,
     backgroundcolor=\color{platinum},
 ]
You are given a cropped video segment. 
A brief description of the activity in this segment is: {{Input Query}}

This activity description is written by a human. Can you enrich the description of the activities happening in this segment? 

Make sure to preserve the meaning of the original annotation. Enrich the query with additional information. Moreover, keep the enriched description brief, preferably only one sentence.
\end{lstlisting}
 \end{minipage}
 \caption{\textbf{Prompt for query enrichment during the pseudo-label generation using a captioning model, LLaVA OneVision 72B \cite{li2024llavaonevision}}. We feed the cropped video between the annotated time interval along with the original query, and ask the model to enrich the query with additional information while maintaining the original focus of the query.}
 \label{fig:onevision_prompt_enrichment}
 \end{figure}


\section{Example Instructions for Different Tasks}
\label{sec:example_instructions}

High-quality language instructions are essential for effective instruction tuning of LLMs across various downstream tasks \cite{wang2024visionllm, li2023m3it, pramanick2024jack}. For each task, we manually write one high-quality instruction as starting and generate variations using GPT-4 \cite{achiam2023gpt4}. Eventually, we manually refine the LLM-generated instructions to obtain the final version. Based on insights from M$^3$IT \cite{li2023m3it} and TimeChat \cite{ren2024timechat}, we use six high-quality instructions per task. During training, we randomly pick one instruction for each sample. Table \ref{tab:instructions} shows one example instruction for each task.

\begin{table*}[!t]
\centering

\small
\setlength{\tabcolsep}{4pt}
\resizebox{\textwidth}{!}{\begin{tabular}{l | p{15cm}}

\toprule

\bf Task & \bf \quad \quad \quad \quad \quad \quad \quad \quad \quad \quad \quad \quad \quad \quad \quad \quad \quad \quad \quad \quad Example Instructions  \\

\midrule

\multirow{3}{*}{STG} & $\bullet$ Please look into the given video and localize the textual query: \( \langle \textit{Input Query} \rangle \). If the provided query is explicit, directly localize it. Otherwise, generate an enriched version which provides more information about the desired time window without changing the main focus, and then localize it. \\

\midrule

\multirow{4}{*}{VPG} & $\bullet$ Carefully review the video and textual queries provided. Your goal is to associate each query with a specific time interval in the video. If a query is clear-cut, directly localize it. For less explicit queries, develop an enhanced version that furnishes more details about the desired time window without changing the core focus, and then localize the enhanced query. Process the queries in the order they appear. The queries are:  \( \langle \textit{Input Queries} \rangle \). \\

\midrule

\multirow{5}{*}{QG} & $\bullet$ Analyze the provided video and the question: \( \langle \textit{Input Question} \rangle \) carefully. Your task is to identify the specific time interval in the video where the question can be accurately answered. If the question is straightforward and easily grounded, directly localize it in the video. However, if the question requires additional context or clarification, generate an enriched version that provides more information without altering its primary focus, and then determine the desired time interval. \\

\midrule

\multirow{5}{*}{AG} & $\bullet$ Carefully look into the given video and the textual queries. Your job is to localize the textual queries in the video. Some of the queries may not be groundable in the input video, in that case, mention it. If a query is groundable and explicit, directly localize it. Otherwise, if the query is groundable, but lacks information, output an enriched version of the query to provide more context about the desired time window without changing the main focus, and then localize the query. Process the queries in the same order as listed in this instruction. The queries are:  \( \langle \textit{Input Queries} \rangle \). \\

\bottomrule
\end{tabular}}
\caption{\textbf{Examples of instructions} for different tasks used by \model. Each instruction provides the model two options: $(i)$ to perform grounding directly when the query is simple and clear, and $(ii)$ to perform grounding in the enrich and detect paradigm, where the model first produces an enriched query with additional information about the desired time window, and then localize it.}
\label{tab:instructions}
\vspace{-2mm}
\end{table*}


\section{Error Analysis} \label{sec:error_analysis}

\begin{figure*}[!t]
\centering
\includegraphics[width=\textwidth]{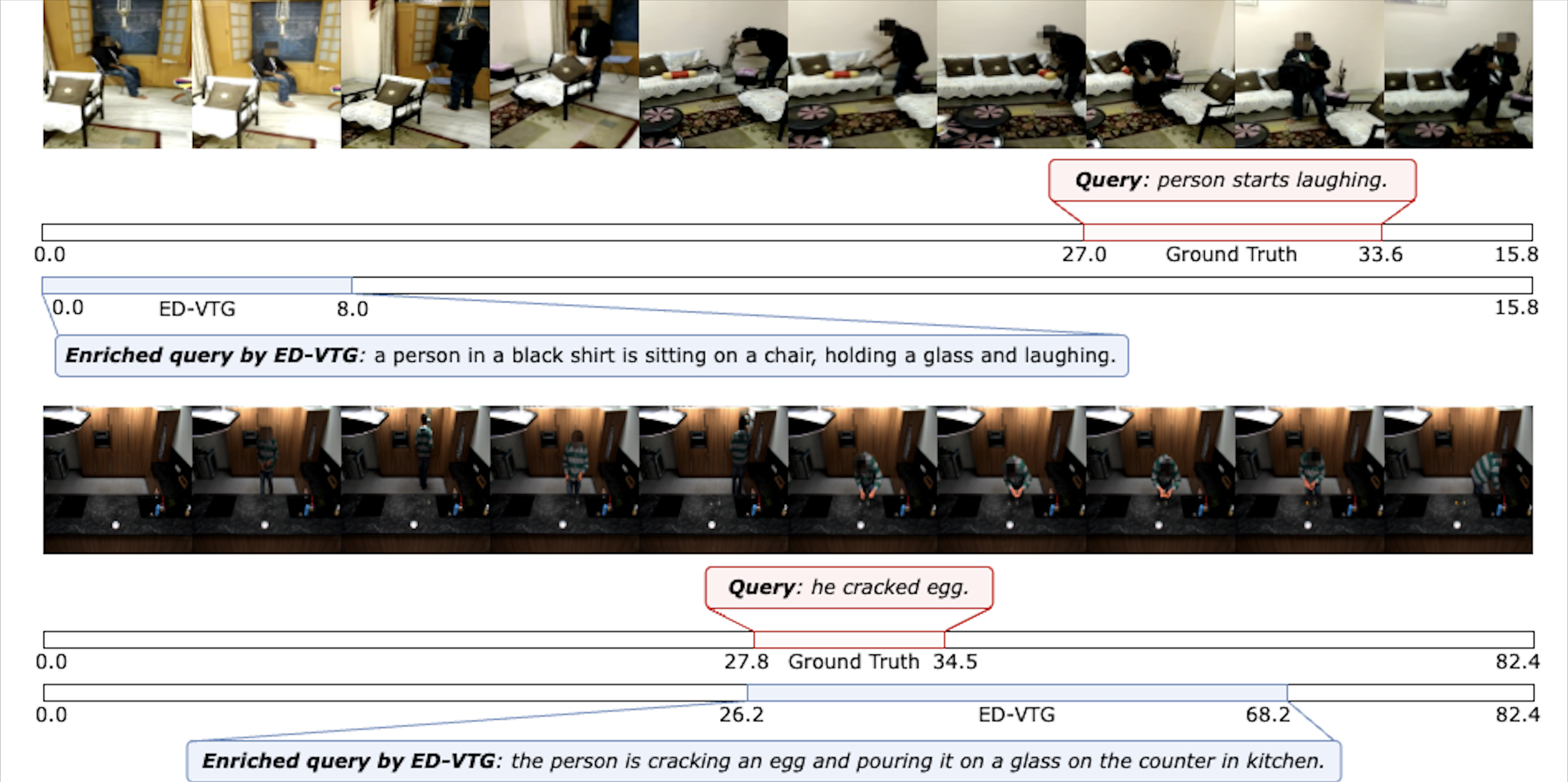}
\vspace{-3mm}
\caption{\textbf{Limitations of our method.} In this figure, we show two error cases where \model\ fails to accurately ground the input queries. The two samples are taken from Charades-STA \cite{gao2017charadessta} and TACoS \cite{regneri2013tacos}, respectively. In the first case, the model completely fails to recognize the correct interval. In the second case, \model\ produces an enriched query that contains an extra action compared to the original query (pouring the egg in a glass), which results in a longer temporal interval prediction which is incorrect.}
\vspace{-3mm}
\label{fig:error_analysis}
\end{figure*}

Although \model\ learns impressive video temporal grounding capability across many different benchmarks, there are still various cases where the model fails to correctly localize the input query, especially for small and obscured objects in long videos. Moreover, since \model\ does not use the audio modality, acoustic expressions are sometimes hard to localize. Figure \ref{fig:error_analysis} shows two such error cases. In the first example, \model\ fails to recognize where the person ``\textit{laughs}'', primarily due to minimal relevant activities before laughter happens. As the face of the person in this video is not fully visible throughout the video, the model fails to detect such sudden and unprecedented activity. However, with acoustic information, such activities would be easy to detect. In the second case, though the query asks to localize where the ``\textit{person cracks egg}'', \model\ produces an enriched query that contains an additional action (pouring the egg in the glass), and consequently grounds it to a longer interval.
This is an example where our enrich-and-detect paradigm fails, as although the enriched query is grounded properly, this behavior is undesired. However these cases are much less common than the ones where enrichment improves the grounding, providing overall - as we have demonstrated quantitively - net performance benefit.

\section{Hyper-parameter settings}
\label{sec:impl_details}
Our hyper-parameter settings during the pre-training and dataset-specific fine-tuning is provided in Tables \ref{tab:pretraining_hyperparams} and \ref{tab:finetuning_hyperparameters}, respectively. To find the most optimal hyper-parameter combinations for different tasks and datasets, we perform a grid search on batch size, learning rate and loss weights, and report the best configuration in Table \ref{tab:finetuning_hyperparameters}.

\begin{table}[!t]
\centering
  \resizebox{0.9\columnwidth}{!}{\begin{tabular}{@{} l|c|c @{}}
    \toprule
    \bf Hyper-parameters & \bf Notation & \bf Value\\
    \midrule
    \multicolumn{3}{c}{\textit{Vision Encoder}}\\
    \midrule
    Frame encoder & $-$ & EVA-CLIP \cite{sun2023evaclip} \\
    Image Q-Former num tokens & $-$ & 32 \\
    Image Q-Former hidden layers & $-$ & 2 \\
    Video Q-Former num tokens & $-$ & 32 \\
    Video Q-Former hidden layers & $-$ & 2 \\
    Video Q-Former window size & $-$ & 32 \\
    Video Q-Former stride & $-$ & 32 \\

    \midrule
    \multicolumn{3}{c}{\textit{Interval Decoder}}\\
    \midrule

    \# Transformer layers & $-$ & 2 \\
    Transformer layer num heads & $-$ & 12 \\
    Transformer layer hidden dim & $-$ & 768 \\
    MLP dim & $-$ & 768 - 256 - 128 - 2 \\

    \midrule
    \multicolumn{3}{c}{\textit{Pre-training}}\\
    \midrule
    Batch size & $-$ & 256\\
    Epochs & $-$ & 40 \\
    Number of frames & $-$ & 96 \\
    Frame resolution &  $-$ & 224 $\times$ 224 \\
    Max. length of text & $-$ & 2048 \\
    Loss weights & $\lambda_\mathrm{LM}$, $\lambda_\mathrm{L1}$, $\lambda_\mathrm{gIoU}$ & 2, 1, 1 \\

    Optimizer & $-$ & AdamW \cite{loshchilov2018adamw} \\
    LoRA rank & $-$ & 32 \\
    Peak LR & $-$ & 5$e$-5\\
    Warmup & $-$ & Linear (first 8 epochs) \\
    LR decay & $-$ & Cosine \cite{loshchilov2016sgdr} \\
    Start LR & $-$ & 1$e$-5 \\
    End LR & $-$ & 1$e$-6 \\
    Num workers & $-$ & 6 \\
    Betas in AdamW & $(\beta_{1}, \beta_{2})$ & (0.9, 0.98) \\
    Eps in AdamW & $-$ & 1$e$-8\\
    Weight decay & $-$ & 0.05 \\

    \bottomrule

  \end{tabular}}
  \caption{\textbf{Pre-training hyper-parameter details of \model.}}
  \label{tab:pretraining_hyperparams}
  \vspace{-2mm}
\end{table}

\begin{table*}[!t]
\centering

\small
\setlength{\tabcolsep}{4pt}
\resizebox{\textwidth}{!}{\begin{tabular}{l | l | c c c c c c c c c c}

\toprule

\multirow{2}{*}{\bf Task} & \multirow{2}{*}{\bf Dataset}  & \multicolumn{10}{c}{\bf \centering Fine-tuning Hyper-parameter Details} \\

& & Batch & Epochs & Warmup & \# Frames & $\lambda_\mathrm{LM}$ & $\lambda_\mathrm{L1}$ & $\lambda_\mathrm{gIoU}$ & Peak LR & Start LR & End LR \\

\midrule

\multirow{3}{*}{STG} & Charades-STA \cite{gao2017charadessta} & 32 & 120 & 24 & 96 & 2 & 1 & 1 & 3$e$-5 & 1$e$-5 & 1$e$-5 \\
& ActivityNet-Captions \cite{krishna2017dense} & 32 & 30 & 6 & 144 & 1 & 1 & 1 & 3$e$-5 & 1$e$-5 & 1$e$-5 \\
& TACoS \cite{regneri2013tacos} & 32 & 120 & 24 & 144 & 4 & 1 & 1 & 3$e$-5 & 1$e$-5 & 1$e$-5 \\

\midrule

\multirow{4}{*}{STG} & Charades-CD-OOD \cite{yuan2021charades-cd} & 32 & 120 & 24 & 96 & 2 & 1 & 1 & 3$e$-5 & 1$e$-5 & 1$e$-5 \\
& ActivityNet-Captions \cite{krishna2017dense} & 32 & 30 & 6 & 144 & 3 & 1 & 1 & 3$e$-5 & 1$e$-5 & 1$e$-5 \\
& TACoS \cite{regneri2013tacos} & 32 & 120 & 24 & 144 & 4 & 1 & 1 & 3$e$-5 & 1$e$-5 & 1$e$-5 \\
& YouCook2 \cite{zhou2018youcook2} & 32 & 120 & 24 & 144 & 1 & 1 & 1 & 3$e$-5 & 1$e$-5 & 1$e$-5 \\

\midrule

AG & HT-Step \cite{afouras2024htstep} & 32 & 120 & 24 & 144 & 2 & 1 & 1 & 3$e$-5 & 1$e$-5 & 1$e$-5 \\

\bottomrule

\end{tabular}}
\vspace{-2mm}
\caption{\textbf{Fine-tuning hyper-parameter details on different datasets.} LR denotes learning rate, $\lambda_\mathrm{LM}$, $\lambda_\mathrm{L1}$ and $\lambda_\mathrm{gIoU}$ denotes weights for LM, L1 and gIoU objectives, respectively. Since the NExT-GQA \cite{xiao2024nextgqa} dataset has no training split, no fine-tuning is performed on NExT-GQA, we report only zero-shot performance. All other hyper-parameters, which are not mentioned in this table, are kept the same as the pre-training setup as listed in Table \ref{tab:pretraining_hyperparams}.} \label{tab:finetuning_hyperparameters}
\end{table*}


\section{Dataset Details}
\label{sec:dataset_details}

This section provides additional details of our pre-training, fine-tuning and evaluation datasets with an in-depth description of our pseudo-label generation pipeline.

\subsection{Pre-training Datasets}
\label{sec:dataset_details_pretraining}

\vspace{1mm}
\noindent \textbf{DiDeMo:} DiDeMo\footnote{\url{https://github.com/LisaAnne/LocalizingMoments}} \cite{anne2017didemo} is a large-scale video temporal grounding dataset featuring 10,464 unique videos, annotated with natural language descriptions that highlight specific moments or events, including single-sentence summaries and shorter moment descriptions. The dataset is sourced from the Flickr Creative Commons dataset \cite{thomee2016yfcc100m} and encompasses a diverse array of topics such as outdoor activities, sports, food preparation, DIY projects, travel destinations, and animals. A notable limitation of DiDeMo is that its interval annotations are made in 5-second windows, which do not capture fine-grained activities. We utilize DiDeMo for pre-training in single-query temporal grounding (STG), where the model receives an input video along with a query and is expected to output a single time interval.

\vspace{1mm}
\noindent \textbf{QuerYD:} QuerYD\footnote{\url{https://www.robots.ox.ac.uk/~vgg/data/queryd/}} \cite{oncescu2021queryd}, sourced from the YouDescribe project \cite{youdescribe}, is a large-scale video grounding dataset designed for moment retrieval and event localization. A distinctive feature of QuerYD is that each video includes two audio tracks: the original audio and a high-quality spoken description of the visual content. We utilize the original audio to generate automatic speech recognition (ASR) transcripts, which are then used as input for the large language model (LLM) along with task instructions. We use this dataset in the STG task format. However, since some samples in QuerYD contain single timepoint annotations instead of time intervals, we introduce a \( \langle \textit{point} \rangle \)
 token to the LLM vocabulary. During pre-training, if a \( \langle \textit{point} \rangle \) token is present in the ground truth, we mask out the window logit in the decoder and set the generalized intersection over union (gIoU) loss to zero.

\vspace{1mm}
\noindent \textbf{COIN:} The COIN\footnote{\url{https://github.com/coin-dataset/annotations}} dataset \cite{tang2019coin} is a large-scale collection designed for comprehensive procedural activity recognition. It comprises over 11,800 videos covering 180 different tasks, which are organized into 12 distinct domains such as ``Sports", ``Leisure", ``Home Improvement", ``Food \& Drinks" etc. Each video is meticulously annotated with step-by-step instructions, providing a detailed breakdown of the procedural activities depicted. This structure allows for the analysis of both high-level task understanding and fine-grained action recognition. The dataset is notable for its diversity, featuring videos sourced from a wide range of environments and cultural contexts, which enhances its applicability to real-world scenarios. Most important to our application, COIN includes temporal annotations that specify the start and end times of each procedural step, facilitating precise temporal action localization. We utilize COIN in the video paragraph grounding (VPG) task format, where we input multiple step descriptions as queries, and ask the model to localize each input query.

\vspace{1mm}
\noindent \textbf{HiREST:} The Hierarchical Retrieval and Step-captioning (HiREST)\footnote{\url{https://github.com/j-min/HiREST}} dataset \cite{zala2023hirest} supports multiple related video-text tasks within an instructional video corpus, including (1) video retrieval, (2) moment retrieval, (3) moment segmentation, and (4) step captioning. HiREST contains 1.1K high-quality, human-annotated moment spans that are relevant to text queries, making it an excellent resource for video grounding. We employ HiREST in both the single-query temporal grounding (STG) and video paragraph grounding (VPG) task formats.

\vspace{1mm}
\noindent \textbf{VITT:} The Video Timeline Tags (VITT)\footnote{\url{https://github.com/google-research-datasets/Video-Timeline-Tags-ViTT}} \cite{huang2020vitt} dataset provides timestamped activity descriptions for a wide range of instructional videos, focusing on hands-on skills such as cooking, car maintenance, and home repairs. It comprises approximately 8,000 videos, each averaging 7.1 segments, with each segment accompanied by a concise free-text description. While VITT is primarily used for dense video captioning, we adapt the dataset to the video paragraph grounding (VPG) format, where segment descriptions are inputted, and the system is tasked with localizing them within the video. Similar to the QuerYD dataset, samples in VITT include single timepoint annotations, for which we employ a \( \langle \textit{point} \rangle \) token and back-propagate using only the L1 objective.

\vspace{1mm}
\noindent \textbf{YTTemporal:} YTTemporal-1B \cite{zellers2022yttemporal} comprises 18 million narrated videos sourced from YouTube, from which we utilize the same subset as TimeChat \cite{ren2024timechat}. In our approach, we employ YTTemporal in the video paragraph grounding (VPG) task setup, where the speech content from the narrations is inputted, and the model is tasked with predicting the start and end timestamps based on the video's visual signals. Due to the often poorly worded and incomplete nature of the narrations, this dataset serves as a weakly-supervised annotation source. The enriched queries significantly aid \model\ in achieving accurate grounding. Following the methodology of Vid2Seq \cite{yang2023vid2seq}, we use Whisper-timestamped \cite{whisper, radford2023robust} to automatically transcribe the speech, which is then used as input queries.


\vspace{1mm}
\noindent \textbf{CrossTask:} The CrossTask\footnote{\url{https://github.com/DmZhukov/CrossTask}} \cite{zhukov2019crosstask} dataset is a valuable resource for learning and evaluating models on cross-domain task understanding and procedural activity recognition. It consists of approximately 4,800 videos spanning 18 primary tasks and 65 related tasks, such as ``Make Pancakes", ``Change Car Tire" and ``Assemble Shelter" each sourced from diverse domains. We use a subset of CrossTask containing 2.7K videos for article grounding (AG). Since this dataset does not contain negative queries, we generate synthetic negatives using the LLaMA 3.1 8B \cite{dubey2024llama3} model. We provide the model with video descriptions (dense captions and ASR) and ask it to generate negative queries that resemble the video activities but do not actually occur in the video. Afterwards, we filter the generated negative queries using multimodal LLaVA OneVision 72B \cite{li2024llavaonevision}, and manually verify a small portion (5\%) of the filtered negative queries for quality assurance.

\vspace{1mm}
\noindent \textbf{VideoCC:} VideoCC\footnote{\url{https://github.com/google-research-datasets/videoCC-data}} \cite{nagrani2022videocc} is a large-scale dataset designed for video captioning and temporal video grounding, featuring 6.3 million video clips accompanied by 974,247 temporally-aligned captions. For our purposes, we utilize a smaller subset of 45,000 caption-interval pairs within the single-query temporal grounding (STG) task setup. The videos in this dataset span a wide array of categories, such as sports, cooking, travel, and more, offering a diverse range of scenarios for model training and evaluation. This diversity makes VideoCC an invaluable resource for developing models that can effectively understand and describe video content across various contexts. Notably, since we use only a subset of YTTemporal and VideoCC, we will easily be able to scale up our pre-training in future.

\subsection{Fine-tuning and Evaluation Datasets}
\label{sec:dataset_details_ft_and_eval}

\vspace{1mm}
\noindent \textbf{Charades-STA:} Charades-STA\footnote{\url{https://github.com/jiyanggao/TALL}} \cite{gao2017charadessta} is a specialized dataset designed for the task of temporal activity localization in videos, particularly focusing on the alignment of textual descriptions with specific video segments. Charades-STA contains 9,848 videos capturing daily indoor activities and 16,128 human-tagged query texts. Following previous works \cite{moon2023qddetr, lin2023univtg, sun2024trdetr, li2024momentdiff}, we use the train set containing 12,408 samples for fine-tuning while the test set with 3,720 samples for evaluation. We report the single-query temporal grounding (STG) results on Charades-STA.

\vspace{1mm}
\noindent \textbf{Charades-CD-OOD:} Charades-CD-OOD\footnote{\url{https://github.com/yytzsy/grounding_changing_distribution/tree/main/Charades-CD}} \cite{yuan2021charades-cd} is a reorganized version of the Charades-STA dataset, specifically designed to evaluate models on their ability to generalize to out-of-distribution (OOD) scenarios in the context of paragraph grounding, which involves testing models on novel combinations of actions and objects that were not seen during training, thereby assessing their ability to extrapolate learned knowledge to new contexts. The dataset is divided into train/val/test ood sets of 4,564/333/1,440 video-paragraph pairs, respectively. The average video duration in Charades-CD-OOD is 30.60 seconds, and the average paragraph length is 2.41 sentences. We report the video paragraph grounding (VPG) performance of \model\ on Charades-CD-OOD.

\vspace{1mm}
\noindent \textbf{ActivityNet-Captions:} ActivityNet-Captions\footnote{\url{http://activity-net.org/download.html}} \cite{krishna2017dense} dataset is a comprehensive resource designed for dense video captioning and temporal localization tasks, derived from the original ActivityNet \cite{krishna2017dense} dataset. ActivityNet-Captions features a diverse array of open-domain content, comprising 14,926 distinct videos and 19,811 localized video-paragraph pairs. On average, each video is approximately 117.63 seconds long, and each paragraph consists of about 3.63 sentences, providing detailed narrative descriptions of the video content. The dataset is structured into three subsets: training, val\_1, and val\_2, containing 10,009, 4,917, and 4,885 video-paragraph pairs, respectively. Consistent with prior research \cite{huang2024vtimellm, ren2024timechat, liu2021cbln, wang2022mmn, cao2021gtr, bao2021depnet, jiang2022svptr}, we use the val\_2 for evaluation. We report both STG and VPG performance of \model\ on ActivityNet-Captions.

\vspace{1mm}
\noindent \textbf{TACoS:} The TACoS\footnote{\url{https://www.mpi-inf.mpg.de/departments/computer-vision-and-machine-learning/research/vision-and-language/tacos-multi-level-corpus}} \cite{regneri2013tacos}
dataset is a specialized collection derived from the MPII Cooking Composite Activities video corpus \cite{rohrbach2012script}, focusing on cooking activities and kitchen scenarios. It comprises 127 videos, each accompanied by multiple paragraphs that describe the actions at varying levels of detail. Specifically, the dataset includes 1,107 video-paragraph pairs for training, 418 for validation, and 380 for testing. On average, the videos are 224.34 seconds long, and each paragraph contains approximately 8.75 sentences, providing rich and detailed descriptions of the cooking processes. The dataset's focus on cooking activities makes it an ideal benchmark for evaluating models that aim to comprehend and describe complex procedural tasks in a structured environment. We report the results on TACoS for the STG and VPG tasks.

\vspace{1mm}
\noindent \textbf{YouCook2:} The YouCook2\footnote{\url{http://youcook2.eecs.umich.edu/download}} \cite{zhou2018youcook2} dataset consists of 2,000 cooking videos sourced from YouTube, capturing a wide variety of cooking styles and cuisines from around the world. These videos are segmented into 15,400 clips, each annotated with detailed descriptions that provide step-by-step instructions for preparing various dishes. On average, each video is approximately 5.19 minutes long, and the dataset covers 89 different recipe types, offering a rich diversity of cooking scenarios. YouCook2 has 1095 and 415 ground truth video-paragraph pairs for train and evaluate, respectively. We report VPG performance of \model\ on YouCook2.

\vspace{1mm}
\noindent \textbf{NExT-GQA:} The NExT-GQA\footnote{\url{https://github.com/doc-doc/NExT-GQA}} \cite{xiao2024nextgqa} dataset is a manually annotated video question grounding dataset, where each question-answer pair is accompanied by a temporal segment annotation serving as evidence. Built upon the NExT-QA \cite{xiao2021nextqa} dataset, NExT-GQA was created by adding 10.5K temporal labels - specifying start and end timestamps - to the QA pairs in the validation and test sets. These labels were carefully annotated and verified as crucial for understanding the questions and identifying the correct answers. Since NExT-GQA does not contain a training split, we evaluate our model's performance on zero-shot question grounding (QG) using this dataset.

\vspace{1mm}
\noindent \textbf{HT-Step:} HT-Step\footnote{\url{https://github.com/facebookresearch/htstep}} \cite{afouras2024htstep} is a large-scale dataset containing temporal annotations of instructional article steps in cooking videos. It includes 116K segment-level annotations over 20K narrated videos (approximately 2.1k hours) of the HowTo100M \cite{miech2019howto100m} dataset. Each annotation provides a temporal interval and a categorical step label from a taxonomy of 4,958 unique steps automatically mined from wikiHow articles \cite{koupaee2018wikihow}, which include rich descriptions of each step. Since HTStep releases the negative queries, we report article grounding (AG) performance on this dataset.


\end{document}